\documentclass[10pt,twocolumn,letterpaper]{article}
\usepackage{cvpr}

\usepackage[dvipsnames]{xcolor}

\newcommand{\red}[1]{\textcolor{purple}{#1}} 
\newcommand{\gre}[1]{\textcolor{olive}{#1}} 
\newcommand{\blu}[1]{\textcolor{cyan}{#1}} 

\usepackage[accsupp]{axessibility} 
\usepackage{amsmath}
\usepackage{amssymb}
\DeclareMathOperator*{\argmax}{arg\,max}
\DeclareMathOperator*{\argmin}{arg\,min}

\usepackage{pifont}

\usepackage{multirow}
\usepackage{comment}
\usepackage{bm}
\usepackage[fixed]{fontawesome5} 
\usepackage{dashundergaps} 

\newcommand{\Cow}[1][]{\includegraphics[width=10pt,trim={6cm 9cm 5cm 6cm},clip]{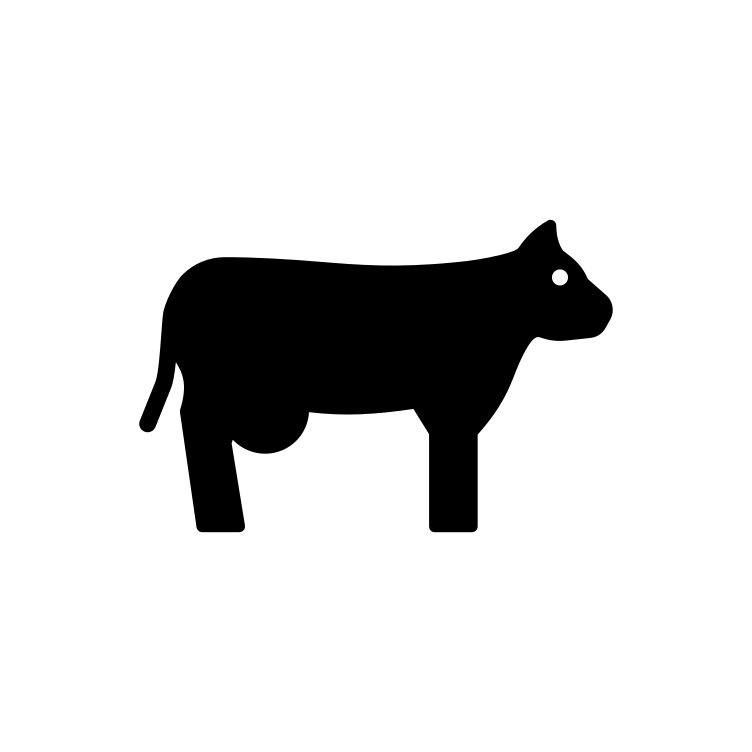}}
\newcommand{\Plant}[1][]{\includegraphics[width=10pt,trim={7cm 6cm 5cm 2cm},clip]{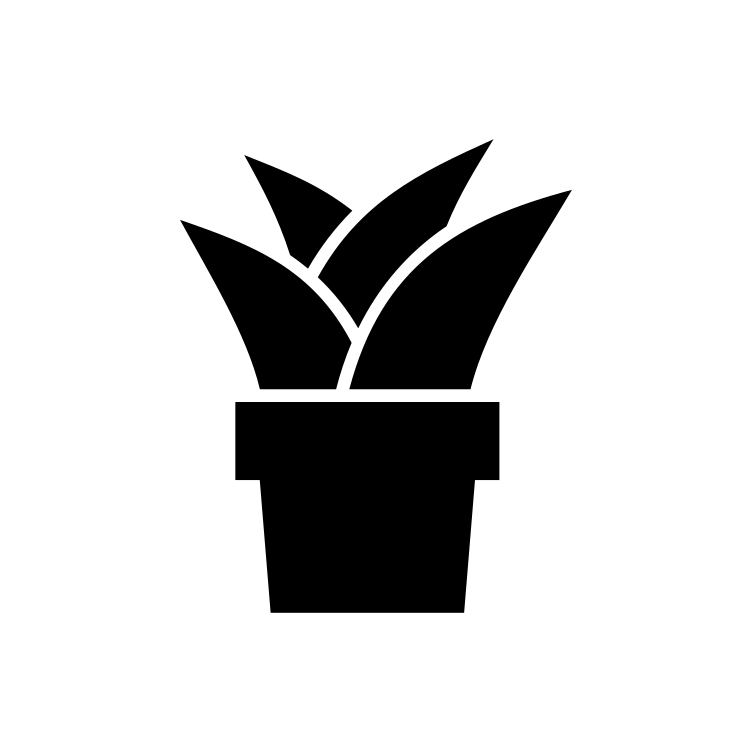}}
\newcommand{\Sheep}[1][]{\includegraphics[width=10pt,trim={6cm 7cm 5cm 6cm},clip]{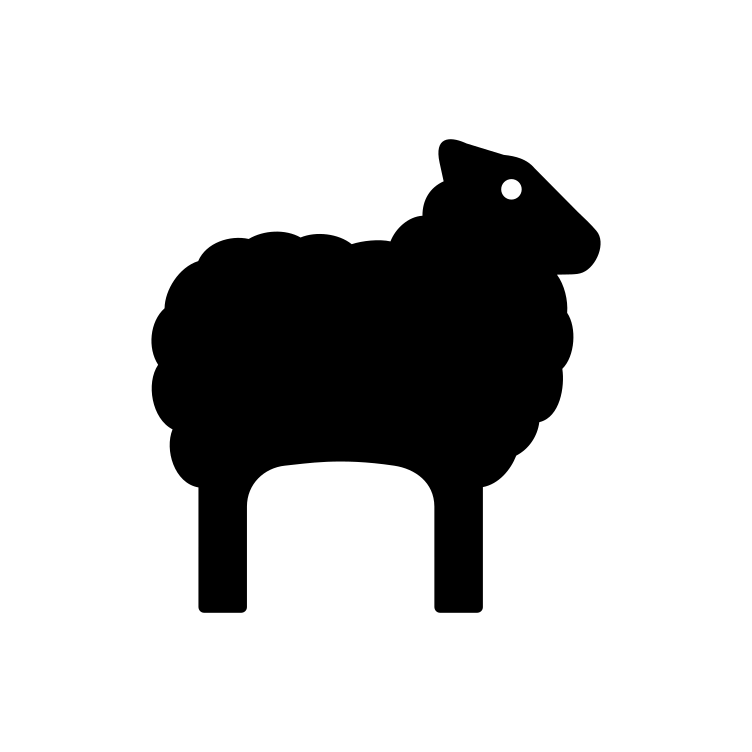}}

\definecolor{cvprblue}{rgb}{0.21,0.49,0.74}
\usepackage[pagebackref,breaklinks,colorlinks,citecolor=cvprblue]{hyperref}

\title{Improving Semantic Correspondence with Viewpoint-Guided Spherical Maps}

\author{Octave Mariotti
\quad\quad Oisin Mac Aodha  
\quad\quad Hakan Bilen \\ [0.3em]
University of Edinburgh \\
\href{https://groups.inf.ed.ac.uk/vico/research/SphericalMaps}{groups.inf.ed.ac.uk/vico/research/sphericalmaps}
}

\begin{document}
\maketitle

\begin{abstract}
Recent self-supervised models produce visual features that are not only effective at encoding image-level, but also pixel-level, semantics. 
They have been reported to obtain impressive results for dense visual semantic correspondence estimation, even outperforming fully-supervised methods. 
Nevertheless, these models still fail in the presence of challenging image characteristics such as symmetries and repeated parts. 
To address these limitations, we propose a new semantic correspondence estimation method that supplements state-of-the-art self-supervised features with 3D understanding via a weak geometric spherical prior. 
Compared to more involved 3D pipelines, our model provides a simple and effective way of injecting informative geometric priors into the learned representation while requiring only weak viewpoint information.
We also propose a new evaluation metric that better accounts for repeated part and symmetry-induced mistakes. 
We show that our method succeeds in distinguishing between symmetric views and repeated parts across many object categories in the challenging SPair-71k dataset and also in generalizing to previously unseen classes in the AwA dataset.
\end{abstract}

\section{Introduction}
\label{sec:intro}

\begin{figure}[t]
    \centering
    \includegraphics[width=\linewidth]{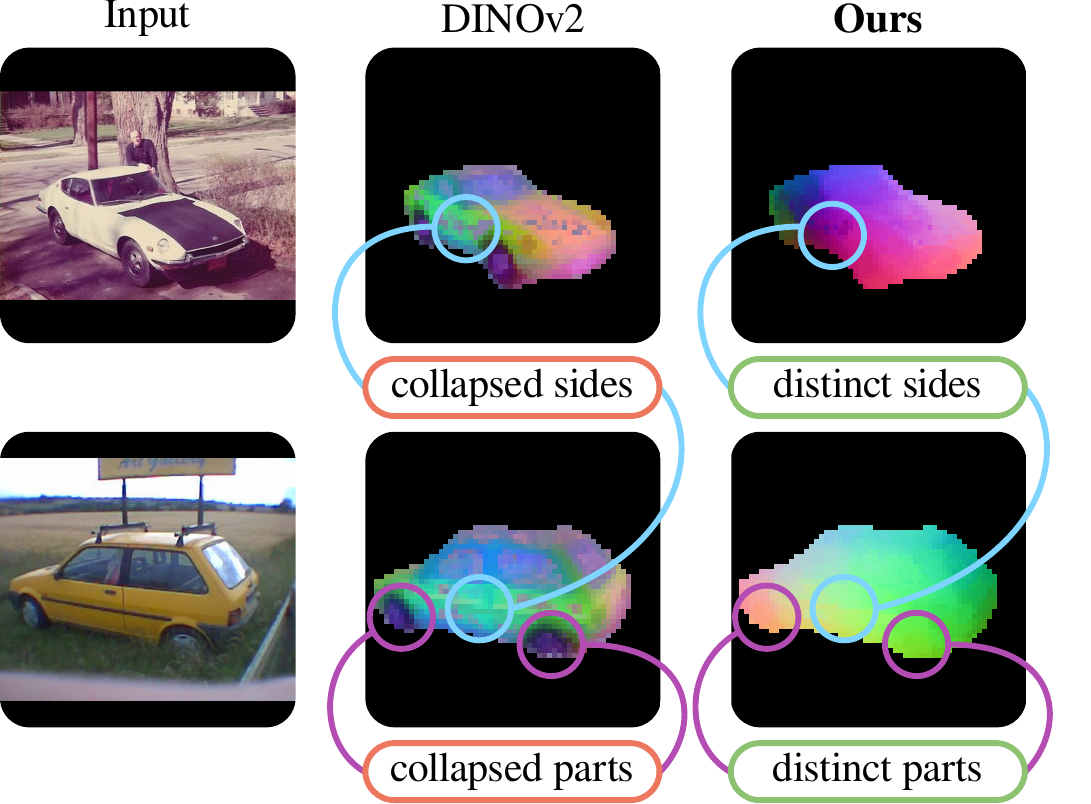} 
    \vspace{-20pt}
    \caption{Features from self-supervised methods such as DINOv2~\cite{oquab2023dinov2} have been used to discover parts and regions of objects. 
    However, these features fail to correctly distinguish (i) object \emph{symmetries}, \eg the left and right side of the car have the same features and (ii) \emph{individual parts}, \eg the wheels are represented by the same features irrespective of their location on the car. 
    Our approach through use of a weak geometric spherical prior addresses these issues. 
    Note, we use the 3D PCA projection of the features for DINOv2 and our learned spherical mapping for our method.
    }
    \label{fig:eyecatcher}
    \vspace{-15pt}
\end{figure}

Semantic correspondence (SC) estimation aims to find local regions that correspond to the same semantic entities across a collection of images, where each image contains a different instance of the same object category~\cite{liu2008sift}. 
SC has been studied in both supervised~\cite{cho2021cats, zhao2021multi, huang2022learning} and unsupervised~\cite{thewlis2017unsupervised, aygun2022demystifying} settings.
Thanks to the progress in self-supervised learning (SSL)~\cite{caron2021emerging, zhou2021ibot, oquab2023dinov2}, recent SSL-based approaches have been shown to obtain strong performance on multiple benchmarks~\cite{amir2021deep, walmer2023teaching, aygun2022demystifying}. 
However, they often act as simple part detectors, without having any 3D understanding of object extent, or relative location of object parts. 
As these SSL methods are typically trained using only 2D-based objectives, they are not able to learn 3D-aware representations, and often converge to similar features for object parts that share appearance but not fine-grained semantics. 

The two main failure modes of current methods are depicted in \cref{fig:eyecatcher}. 
First, they are often confused by the symmetries that objects can exhibit (\eg left \vs right side of a car), producing reflected versions of the feature maps when presented with the two different sides of a symmetric object.
Second, they predict similar features for similar looking parts (\eg the front left \vs back left wheel of a car), and fail to discriminate between part instances. 
Such symmetries and repeated structures are ubiquitous in both human-made and naturally occurring object categories. 

We posit that these shortcomings can be addressed by enforcing an explicit 3D structure into correspondence learning, allowing the model to disambiguate multiple repeated parts, and accounting for the fact that some object parts might not be visible in all images. 
Indeed, scene-based correspondence pipelines use geometric verification in order to guarantee robustness in their matches~\cite{schonberger2016structure}, while other works have shown that detailed 3D meshes can be used to discover category-based correspondences through a render-and-compare pipeline~\cite{zhou2016learning, kulkarni2019canonical}.
However, these approaches typically require 3D mesh supervision which is difficult and costly to obtain at scale. 

Motivated by the limitations, and the promise of the recent SSL representations~\cite{oquab2023dinov2}, we propose a novel method that incorporates the missing 3D geometry into object representations.
We make use of a weak 3D prior by representing object features on the surface of a sphere which encodes both geometry and appearance information. 
Our method consistently maps features corresponding to the same semantic parts (\eg left headlight of a car) in images of different object instances to the same spherical coordinates, hence enabling a 3D aware and accurate correspondence. 
However, learning such a mapping without the ground-truth 3D supervision is nontrivial because of repeated parts and object symmetries. 
To this end, we propose to use much weaker supervision via coarse camera viewpoint annotations to help separate these confusing features during training. 
We couple this with geometric constraints to prevent collapsed representations for repeated parts.
As a result, our method learns to separate object parts based on their position and spatial configuration in an image.

Furthermore, we point out that the standard evaluation metric for SC, Percentage of Correct Keypoints (PCK), is not able to properly evaluate symmetry-related failures as it only considers keypoints that are \emph{simultaneously visible} in both the source and target image. For instance, a system predicting high similarity between a front left wheel in the source and front right wheel in the target is unlikely to be penalized for it, as the target image is unlikely to contain both a front left \emph{and} front right wheel due to self-occlusions. In such cases PCK will simply not evaluate what the nearest neighbor of the source wheel is.
Rather than simple accuracy-based metric, we argue that average precision is a more robust scoring mechanism and propose a new metric, Keypoint Average Precision (KAP), to account for potential spurious matches.

Our contributions are: i) a novel SC approach that learns to separate visually similar parts by mapping them to different location on a sphere, ii) a set of simple geometric constraints that force separation of visually similar parts during training, and iii) a new evaluation protocol that better accounts for the failure cases of current methods.

\vspace{-5pt}
\section{Related work}
\label{sec:relworks}
\vspace{-5pt}

\noindent{\bf Image matching.}
Visual correspondences were first explored as exact matches between different views of the same scene with applications to stereo vision~\cite{nishihara1984prism, hirschmuller2005accurate}, object detection~\cite{lowe2004distinctive, ferrari2006simultaneous}, structure from motion~\cite{torr1999feature, hartley2003multiple, schonberger2016structure}, and SLAM~\cite{harris19883d, davison2007monoslam}. 
Correspondences can also be computed across pixels in the frames of a video,  which is referred to as optical flow~\cite{beauchemin1995computation, chen2013large}. 
Traditionally they are obtained by computing hand-crafted visual descriptors ~\cite{lowe1999object,dalal2005histograms,rublee2011orb}.

\noindent{\bf Semantic correspondence (SC).} 
In SC estimation~\cite{liu2008sift}, the goal is not only to find matches across different views of the exact same object instance, but also across different instances of the same object category.  
While earlier work~\cite{liu2008sift} match two images through densely sampled SIFT~\cite{lowe1999object} features, later works extended pairwise correspondence learning to a global matching problem~\cite{zhou2015flowweb} and established region correspondences using object proposals and their geometric relations over dense descriptors~\cite{ham2017proposal}. 
Subsequently, CNNs trained with only image-level supervision have been shown to discover correspondences~\cite{long2014convnets}, paving the way for many deep learning correspondence systems~\cite{choy2016universal, han2017scnet, zhao2021multi, huang2022learning}.
Typical correspondence-specific CNN architectures rely on aggregating features along the depth of a network~\cite{ufer2017deep, min2020learning, zhao2021multi} or use 4D convolutions spanning two images simultaneously~\cite{rocco2018neighbourhood}. 
Recently, transformers have been shown to be particularly suited to correspondence discovery thanks to their attention mechanism~\cite{cho2021cats}.

While the early works in unsupervised correspondence discovery~\cite{thewlis2017unsupervised, jakab2018unsupervised, aygun2022demystifying} use on the convolutional representations, recent  methods~\cite{amir2021deep, walmer2023teaching, taleof2feats, shtedritski2023learning} build on the success of transformer-based self-supervised representations~\cite{caron2021emerging, oquab2023dinov2, zhou2021ibot} to discover salient and robust features, which subsequent works have applied to detect semantic correspondences successfully, often surpassing supervised methods.
The features from generative image models have also been used in SC~\cite{hedlin2023unsupervised,taleof2feats, tang2023emergent}. 
Two closely related works~\cite{ofri2023neural, gupta2023asic} propose to create category-wide representations by aligning self-supervised feature maps. 
Unlike our approach, these works are limited to 2D, and find correspondences between images without knowing the underlying 3D structure of the world which renders them incapable of distinguishing different sides of objects (\eg left \vs right side of a car).
A concurrent work~\cite{zhang2023telling} also identifies geometry-related issues in SSL-based correspondence but targets a supervised training regime where our approach does not necessitate keypoint supervision.

\noindent{\bf 3D priors for correspondence.}
In literature, 3D priors have been used for SC, mainly in the form of annotated meshes~\cite{zhou2016learning, kulkarni2019canonical, neverova2020continuous, shapovalov2021densepose, cheng2021learning}, where annotations can either be from image to mesh, or between meshes. \cite{kulkarni2019canonical, neverova2020continuous} in particular use a dense parametrization of the mesh surface similar to our sphere. When no mesh is available, precise camera pose can be used to recover sparse correspondences through constrained geometric optimization~\cite{suwajanakorn2018discovery} or by training a neural field~\cite{yen2022nerf}. 
In comparison, our approach requires no meshes and only very coarse viewpoint supervision.

Alternatively, it is possible to estimate 3D models of object categories from image supervision alone~\cite{henderson2018learning, goel2020shape, hu2021self, monnier2022share, wu2023magicpony, aygun2023saor}, from which correspondences are extracted. 
However, this is very challenging by itself, requiring complex pipelines capable of modeling accurate mesh deformations, pose, and rendering, which are all failure-prone. 
\section{Method}

\begin{figure*}[t]
    \centering
    \includegraphics[width=1\linewidth]{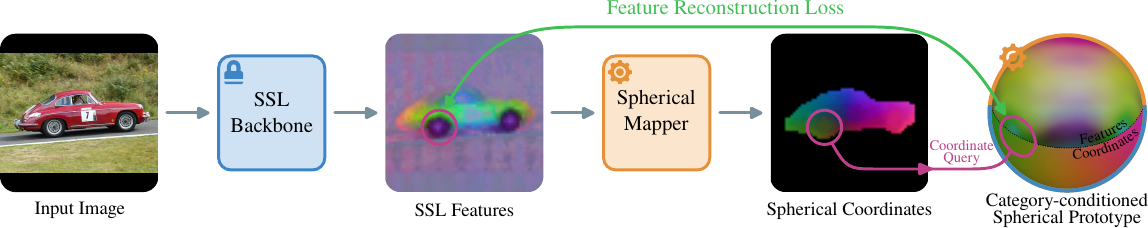}
    \vspace{-17pt}
    \caption{Overview of our semantic correspondence estimation approach. We begin by extracting features from a frozen self-supervised backbone, and further use them to predict spherical coordinates via a learned module. Each predicted point is used to query a jointly learned prototype, providing the supervision signal~(\cref{sec:corresp_learning}). The sphere is used to enforce weak geometric priors~(\cref{sec:geom_priors}). During inference, SSL features are combined with spherical coordinates~(\cref{sec:alpha_mix}). A blue outline indicates a fixed module, while orange indicate learned parameters.}
    \label{fig:arch}
    \vspace{-12pt}
\end{figure*}

\subsection{Problem statement} 
\paragraph{General formulation.}
Given an unordered monocular image collection, we wish to learn a mapping $f(\bm{I},x)=\bm{z}$ that takes in an RGB image $\bm{I}:\Lambda\rightarrow \mathbb{R}^3$, $\Lambda \subset \mathbb{R}^2$ and a pixel location $x \in \Lambda$ to an $n$-dimensional embedding $\bm{z} \in \mathcal{Z} \subset \mathbb{R}^n$.
In particular, we would like $f$ to map \emph{semantic} concepts from different  instances of the same category (\eg center of front-right wheel of a car) to the same point in the embedding space $\mathcal{Z}$.
Formally, $f(\bm{I},x)=f(\bm{I}',x')$,  when the pixels $x$ in $\bm{I}$, and $x'$ in $\bm{I}'$ correspond to the same semantic part and instance.
In practice, the corresponding location for $(\bm{I},x)$ in $\bm{I}'$ can be obtained by finding the nearest neighbor of $f(\bm{I},x)$ in the embedding space, \ie $\argmin_{x'}d_{\mathcal{Z}}(f(\bm{I},x), f(\bm{I}',x'))$, where $d_{\mathcal{Z}}$ is a distance function over $\mathcal{Z}$. 
For simplicity, we will use $f(\bm{I})$ to refer to the feature map obtained by evaluating $f$ on all pixels. 

An ideal embedding function should be general enough as to not only match related semantic parts across different object instances (\eg match the front-left wheel across two different car models) but also specific enough to distinguish between different part instances within the same object instance (\eg front-left wheel \vs front-right wheel). 
Learning this from image-based self-supervision alone is a challenging task due to the visual similarity of repeated parts and self-occlusion. 
As a result, it is quite challenging for correspondence estimators to learn that cars have four distinct wheels from single images without explicit supervision.

\paragraph{Spherical mapping.}
Instead, it is beneficial to endow a correspondence estimator with priors about the underlying 3D structure of objects. 
To this end, previous methods~\cite{zhou2016learning, kulkarni2019canonical} use detailed 3D meshes via a render-and-compare approach.
However, they rely on the presence of groundtruth 3D meshes and the precise viewpoint information.
Instead, following~\cite{thewlis2017unsupervised,thewlis2019unsupervised}, we use a simplified 3D structure, specifically a sphere $S^2 \subset \mathbb{R}^3$ (denoted as $S$ for simplicity) based on the  assumption that the object categories of interest are approximately homeomorphic to spheres (\ie objects without holes).
While this may at first appear particularly ill-suited for some categories, \eg bicycles, the geometric simplicity of $S$ provides a convenient parameterization of an object's surface by allowing easy computation of geometric constraints to separate visual and semantic similarity.
In particular, $S$ provides a simple geodesic distance through the dot product, making it straightforward to push different semantic parts apart, which we use to distinguish repeated object features.
Formally, we are interested in building a \emph{spherical mapping} $f_S(\bm{I}, x)$ that maps pixels in $\bm{I}$ to the surface of $S$.
Unlike an arbitrary $n$ dimensional space, $S$ forms an object-centric specific coordinate system where each region ideally represents a distinct semantic part \emph{and} instance on the surface of the object.

\subsection{Learning formulation}
\label{sec:corresp_learning}
In the absence of ground truth correspondences, ensuring the semantic consistency of our spherical mapping function is very challenging. 
\cite{thewlis2017unsupervised} enforces correspondence between image pairs through synthetically augmenting training images via  thinplate splines. 
Unfortunately, such augmentations do not produce realistic 3D transformations such as 3D rotations which can result in the self-occlusion of parts.
Instead, similar to~\cite{ofri2023neural, gupta2023asic}, we formulate the learning objective as one of aligning features from a pretrained SSL model $\phi: \Lambda\rightarrow\mathcal{Z}_\phi$ between different instances of an object category. For simplicity, we will refer to the SSL embedding space $\mathcal{Z}_\phi$ simply as $\mathcal{Z}$ in the rest of the text.

While those approaches use flat atlases $\mathbb{R}^2 \rightarrow \mathcal{Z}$, we build a \emph{spherical prototype} $S_\mathcal{Z}: S \rightarrow \mathcal{Z}$ that maps a point on the surface of the sphere $s \in S^2$ to a feature vector $\mathbf{z} \in \mathcal{Z}$.
This formulation has two important benefits. 
First, the input domain of $S_\mathcal{Z}$ is simply $S$, meaning that it does not depend on $I$, and it is shared across all images, constituting a category-based map. 
Second, and in contrast to~\cite{ofri2023neural, gupta2023asic}, because we chose the output of $f_S$ to lie on the surface of a sphere, we can use it to enforce 3D priors during training~(\cref{sec:geom_priors}). 

Intuitively, for a given category, the spherical prototype $S_\mathcal{Z}$ should be able to reconstruct any self-supervised feature map $\phi(\bm{I})$ when queried on the coordinates predicted by $f_S$, \ie $S_\mathcal{Z}(f_S(\bm{I},x))$ must closely approximate $\phi(\bm{I},x)$ for any $\bm{I},x$. 
Even if $S_\mathcal{Z}$ encodes a perfect category prototype, it will be unable to encode information about the image background. 
Hence we use instance masks $M(\bm{I})$ at training time to restrict the loss computation to only the pixels on an object's surface. 
In practice, these can be obtained using pretrained segmentation models~\cite{he2017mask}. 
Thus, given a training image $\bm{I}$ and a pixel location $x$, our primary training objective to learn the parameters of $f_S$ and $S_\mathcal{Z}$ can be formulated as the minimization of 
\begin{equation}
    \mathcal{L}_{rec} = \frac{1}{|\Lambda|}\sum_{x\in\Lambda} M(\bm{I},x) \times \Gamma \left( \phi(\bm{I},x), S_\mathcal{Z}(f_S(\bm{I},x))\right),
\end{equation}
where $\Gamma$ is the cosine distance between the two inputs. 
Our spherical prototypes are implemented via a neural network that maps points in $\mathbb{R}^3$ to $\mathcal{Z}$. 
Importantly, each point $s \in S^2 \subset \mathbb{R}^3$ must be mapped independently, meaning that it should correspond to a unique point $\mathbf{z} \in \mathcal{Z}$, irrespective of which instance it comes from or where in the image this part is located. 
To model multiple categories, and to encourage information sharing between closely related ones, we use a single network conditioned on a category embedding. 
In practice, $S_\mathcal{Z}$ is a visual transformer~\cite{dosovitskiy2020image} \emph{without} self-attention, but with cross-attention between image tokens and a single category token $\mathbf{c}$ which is a one hot embedding of the category.
As learning $f_S$ from scratch can be challenging, we further leverage the SSL backbone by simply adding a prediction head $f_S'$ to it to produce spherical maps, \ie $f_S = f_S' \circ \phi$. An overview of our architecture is presented in \cref{fig:arch}.

\subsection{Enforcing geometric priors}
\label{sec:geom_priors}
The above spherical reconstruction formulation alone offers little benefit over matching the self-supervised features from $\phi$ directly, especially considering the low dimensionality of the intermediate spherical space, which is not guaranteed to be well-behaved. 
However, this spherical structure enables the computation of simple geometric priors that can be enforced during training. 

\paragraph{Viewpoint regularization.}
Many object categories exhibit some form of symmetry which can make it particularly challenging for image-based models to distinguish between different sides of an object, creating spurious correlations between distinct parts (\eg see cars in \cref{fig:eyecatcher}). 
One potential way of addressing this challenge, without using dense groundtruth correspondence, is to exploit viewpoint information of the object instance.

An interesting property of using a spherical prototype to represent an object category is that it can easily be used to infer an object's viewpoint with respect to the camera.
We posit that the average coordinate of a spherical map of an image, which we present as $\mu(f_S(\bm{I})) = {1}/{|\Lambda|}\sum_{x\in\Lambda}f_S(\bm{I},x)$, can be viewed as a coarse approximation of the camera viewpoint under which the object is seen, \eg right side views should be approximately mapped to the right side of the sphere, and left side views to the left. 
Intuitively, thinking of $f_S$ as a mapping from image pixels to the object's surface, the visible parts of an object should roughly be those closest to the camera, while the others should be hidden by self-occlusion.

By using the groundtruth camera viewpoint $v_{\bm{I}}$, we enforce the average direction of the spherical map $\mu(f_S(\bm{I}))$ to align with $v_{\bm{I}}$. 
When this is satisfied, symmetric views must be mapped to different parts of the sphere.
While precise viewpoint information is hard to obtain, we observe it is sufficient to use \emph{coarse} relative viewpoint supervision between images. 
In practice, viewpoints are discretized into a small number of bins, and $v_{\bm{I}}$ is represented by the median value of the corresponding bin. Then, we simply enforce that the dot product between viewpoints from two images equals that of their average spherical maps,
\begin{equation}
    \mathcal{L}_{vp} = \sum_{\bm{I},\bm{I'}}|| v_{\bm{I}}\cdot v_{\bm{I'}} -  \mu(f_S(\bm{I}))\cdot \mu(f_S(\bm{I}'))||^2.
    \label{eq:vp_loss}
\end{equation}

\paragraph{Relative distance loss.}

As noted earlier, current self-supervised extracted features can confuse repeated object repeated parts.
Depending on the target goal, this can be desirable, if the parts do indeed belong to the same part category. 
However, this is detrimental for dense correspondence estimation as the parts may actually be different part instances, \eg left \vs right leg. 
It can be hard to distinguish between these based on their appearance alone and additional geometric context is required. 

One trivial way of encoding this context is by making the assumption that similar parts that appear in different locations on an image (\ie not beside each other) are in fact different instances of the same semantic part.
Therefore, we would like to enforce the property on our predicted spherical maps that \emph{distant pixels in the image should be mapped far apart on the sphere}. 
Due to scale variation, distance on the image plane is hard to link to distances on the surface of the sphere. 
However, we can reformulate this property as a relative distance comparison using a  triplet. 
Given three points $a$, $b$, and $c$ on an image $\bm{I}$, and their corresponding positions $s_a=f_S(\bm{I},a), s_b$, and $s_c$ on the sphere, the ordering between distances should be preserved, that is: 
\begin{equation}
||a-b|| \leq ||a-c|| \iff \Gamma(s_a,s_b) \leq \Gamma(s_a,s_c).
\end{equation}
In reality, this does not hold for distances along an object's surface, which the sphere is supposed to represent, partly due to perspective. 
For example, in the case of a long train seen from a near frontal view, parts at the back of the train might appear very close to the front in the image, even though they are very far in 3D space. 
Nevertheless, we find that this is still an effective prior for discouraging repeated part instances from being mapped to the same location.

\begin{figure}[t]
    \centering
    \includegraphics[width=\linewidth]{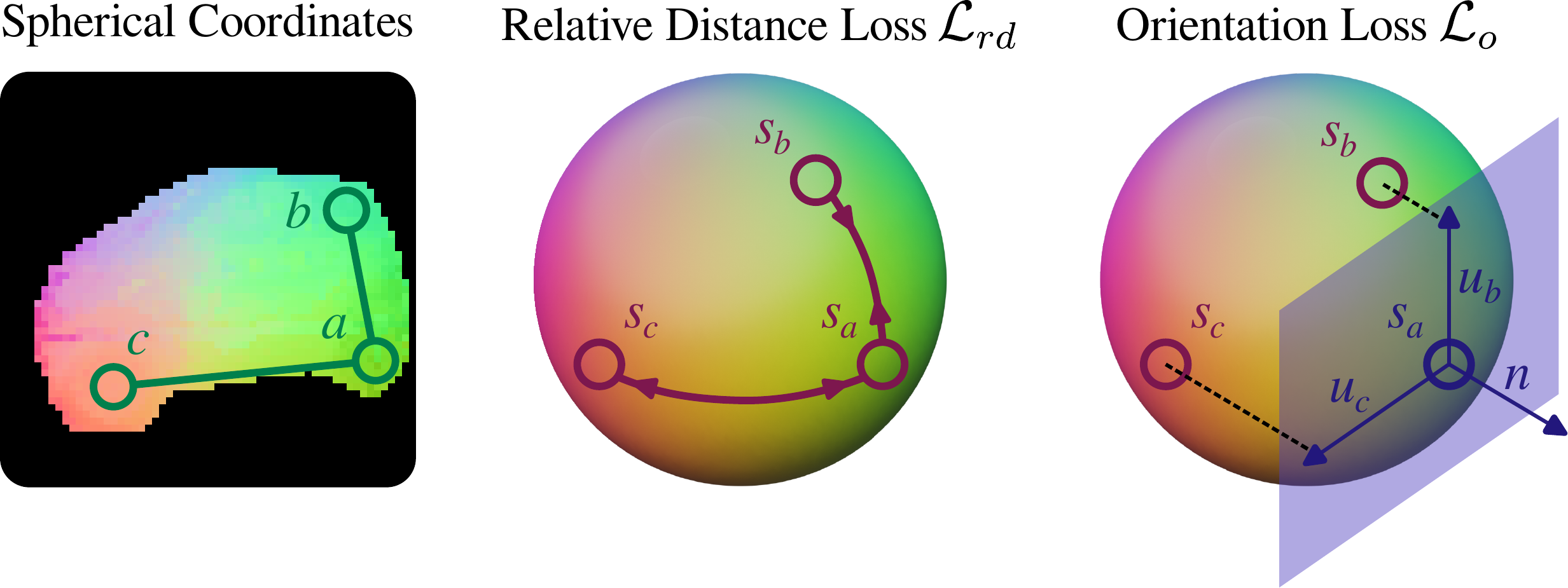}
    \vspace{-10pt}
    \caption{Illustration of our geometry losses $\mathcal{L}_{rd}$ and $\mathcal{L}_{o}$. The left image shows a spherical map from which a triplet of points is sampled. $\mathcal{L}_{rd}$: as the anchor patch $a$ is closer to the positive $b$ on the image compared with the negative $c$, its corresponding position $s_{a}$ on the sphere must also be closer to $s_b$ than $s_c$. $\mathcal{L}_{o}$: after projecting $s_{b}$ and $s_{c}$ to the plane tangent to the sphere at $s_{a}$, we ensure orientation is preserved by enforcing positive colinearity between $u_b \times u_c$ and the normal vector $n$.
    }
    \label{fig:rel_dist_loss}
    \vspace{-8pt}
\end{figure}

During training, point triplets $(a,b,c)$ are randomly sampled among pixels that belong to the object. 
Among each such triplet, we chose an anchor location $anc = a$, a positive location $pos = \argmin_{x\in\{b,c\}}||a-x||$, and a negative location $neg = \argmax_{x\in\{b,c\}}||a-x||$ (see \cref{fig:rel_dist_loss}). 
We can then define a triplet margin loss as:
\begin{equation}
    \begin{split}
        \mathcal{L}_{rd} = \max(&\Gamma(f_S(\bm{I}, anc), f_S(\bm{I}, pos))\\
                               -&\Gamma(f_S(\bm{I}, anc), f_S(\bm{I}, neg)) + \delta, 0), 
    \end{split}
\end{equation}
where $\delta$ is a hyperparameter which specifies the margin.
A useful side product of this formulation is that it encourages the spherical maps to be smooth, as strong discontinuities would produce a large anchor to positive distance.

\paragraph{Orientation loss.}
A final advantage of using $S$ as a latent space is that it is an orientable 2D space, just like images. Therefore, enforcing that $f_S$ preserves orientation prevents spurious matches between symmetric views. Intuitively, if a point $a$ appears on the right side of $b$ in the image, then $s_a$ must appear in the same side of $s_b$. Given a point triplet as defined earlier, this can be enforced by making sure that the determinant of the image triplet and sphere triplet positively correlate. Formally, defining $P_a$ as the orthogonal projection to the plane tangent to $S$ at $s_a$ and $u_b$ as $P_a(s_b) - s_a$, we have $d_I = \text{det}(b-a, c-a)$ and $d_S = \text{det}(u_b, u_c)$, and want $sign(d_I) = sign(d_S)$~(see \cref{fig:rel_dist_loss}).

As objects can have complex geometry that our model does not have access to, there exists no exact relation linking the two determinants together, as for instance perspective distortions might alter angles. However, we can assume that image triplets of large determinants should also have large determinant on the sphere, as a change of sign in determinant indicates the relative orientation on the sphere has been inverted.
In practice, we randomly sample image triplets $(a,b,c)$ and select those whose image determinant $d_I$ is higher than a threshold $d_\tau = 0.7$, while swapping $b$ and $c$ if $d_I$ is negative. Then, we enforce the determinant of the corresponding sphere triplet to be at least $d_\tau$, 
\begin{equation}
    \begin{split}
        \mathcal{L}_{o} = \begin{cases}
            0 & \mbox{if } d_I < d_\tau \\
            max(d_\tau - d_S, 0) & \mbox{if } d_I \geq d_\tau .
        \end{cases} 
    \end{split}
\end{equation}

\begin{table*}[t]
    \centering
    \renewcommand{\b}{\bfseries}
    \renewcommand{\u}{\rlap{\underline{\phantom{00.0}}}}
    \renewcommand{\d}{\rlap{\dashuline{\phantom{00.0}}}}
    \resizebox{\textwidth}{!}{
    \begin{tabular}{cl|rrrrrrrrrrrrrrrrrr|r}
            && \faIcon{plane} & \faIcon{bicycle} & \faIcon{crow} & \faIcon{ship} & \faIcon{wine-bottle} & \faIcon{bus} & \faIcon{car} & \faIcon{cat} & \faIcon{chair} & \Cow & \faIcon{dog} & \faIcon{horse} & \faIcon{motorcycle} & \faIcon{walking} & \Plant & \Sheep & \faIcon{train} & \faIcon{tv} & avg\\ 
        \midrule
        \multirow{3}{*}{\rotatebox{90}{\parbox[t]{35pt}{\hspace*{\fill}Custom\hspace*{\fill}}}} &
         CATS~\cite{cho2021cats}
            & 52.0 & 34.7 & 72.2 & 34.3 & 49.9 & 57.5 & 43.6 & 66.5 & 24.4 & 63.2 & 56.5 & 52.0 & 42.6 & 41.7 & 43.0 & 33.6 & 72.6 & 58.0 & 49.9\\
         & MMNet+FCN~\cite{zhao2021multi}
            & 55.9 & 37.0 & 65.0 & 35.4 & 50.0 & 63.9 & 45.7 & 62.8 & 28.7 & 65.0 & 54.7 & 51.6 & 38.5 & 34.6 & 41.7 & 36.3 &\d 77.7 & 62.5 & 50.4\\
         & SCorrSan~\cite{huang2022learning}
            & 57.1 & 40.3 & 78.3 & 38.1 & 51.8 & 57.8 & 47.1 & 67.9 & 25.2 & 71.3 & 63.9 & 49.3 & 45.3 & 49.8 & 48.8 & 40.3 &\d 77.7 &\b 69.7 & 54.4\\
        \midrule
        \multirow{3}{*}{\rotatebox{90}{\parbox[t]{35pt}{\hspace*{\fill}DINOv1\hspace*{\fill}}}} &
         DINOv1~\cite{caron2021emerging}
            & 44.3 & 26.8 & 57.6 & 22.0 & 29.3 & 32.8 & 19.7 & 54.0 & 14.9 & 40.1 & 39.3 & 29.3 & 29.0 & 37.0 & 20.0 & 28.2 & 40.6 & 21.1 & 32.6\\
         & ASIC~\cite{gupta2023asic}
            & 57.9 & 25.2 & 68.1 & 24.7 & 35.4 & 28.4 & 30.9 & 54.8 & 21.6 & 45.0 & 47.2 & 39.9 & 26.2 & 48.8 & 14.5 & 24.5 & 49.0 & 24.6 & 37.0\\
         & Ours
            & 47.1 & 26.0 & 70.9 & 21.8 & 37.5 & 34.9 & 32.4 & 60.0 & 23.2 & 53.6 & 48.5 & 42.5 & 28.3 & 42.7 & 21.1 & 41.9 & 39.7 & 41.7 & 39.7\\
        \midrule
        \multirow{3}{*}{\rotatebox{90}{\parbox[t]{0pt}{\hspace*{\fill}SD\hspace*{\fill}}}} &
            DIFT~\cite{tang2023emergent}
            & 63.5 & 54.5 & 80.8 & 34.5 & 46.2 & 52.7 & 48.3 & 77.7 & 39.0 & 76.0 & 54.9 & 61.3 & 53.3 & 46.0 &\u 57.8 & 57.1 & 71.1 &\u 63.4 & 57.7\\
         & SD~\cite{taleof2feats} 
            & 63.1 & 55.6 & 80.2 & 33.8 & 44.9 & 49.3 & 47.8 & 74.4 & 38.4 & 70.8 & 53.7 & 61.1 & 54.4 & 55.0 & 54.8 & 53.5 & 65.0 & 53.3 & 56.1\\
            \midrule
        \multirow{3}{*}{\rotatebox{90}{\parbox[t]{55pt}{\hspace*{\fill}DINOv2\hspace*{\fill}}}} &
         DINOv2~\cite{oquab2023dinov2}
            & 72.7 & 62.0 & 85.2 & 41.3 & 40.4 & 52.3 & 51.5 & 71.1 & 36.2 & 67.1 & 64.6 & 67.6 & 61.0 &\u 68.2 & 30.7 & 62.0 & 54.3 & 24.2 & 56.2 \\
         & DINOv2 + SD~\cite{taleof2feats} 
             & 73.0 &\u 64.1 &\u 86.4 & 40.7 &\b 52.9 & 55.0 & 53.8 & 78.6 & 45.5 & 77.3 & 64.7 & 69.7 & 63.3 &\b 69.2 &\b 58.4 & 67.6 & 66.2 & 53.5 & 63.3\\
         &  Ours (sphere only)
            & 46.7 & 28.8 & 66.3 & 33.0 & 36.5 & 66.6 & 59.1 & 74.9 & 25.4 & 65.7 & 50.1 & 52.7 & 27.1 & 13.7 & 15.8 & 46.6 & 73.5 & 36.7 & 45.5\\
         &  Ours
            &\b 76.9 & 61.2 & 85.9 &\u 42.1 & 48.4 &\u 73.3 &\u 67.2 &\u 80.0 &\u 46.3 &\u 80.2 &\u 66.7 &\u 71.2 &\u 66.0 & 63.9 & 36.2 &\u 68.6 & 67.8 & 42.2 &\u 63.6\\
         & Ours + SD
            &\u 74.8 &\b 64.5 &\b 87.1 &\b 45.6 &\u 52.7 &\b 77.8 &\b 71.4 &\b 82.4 &\b 47.7 &\b 82.0 &\b 67.3 &\b 73.9 &\b 67.6 & 60.0 & 49.9 &\b 69.8 &\b 78.5 & 59.1 &\b 67.3\\
    \end{tabular}
    }
    \vspace{-5pt}
    \caption{Keypoint matching scores on SPair-71k evaluated using PCK@0.1 with \emph{macro}-averaging for the summary scores. 
    We present our approach using DINOv1~\cite{caron2021emerging} features (middle rows) and DINOv2~\cite{oquab2023dinov2} features (bottom rows). 
    In both cases, we improve over the DINO only baselines and are superior to fully supervised methods (top rows). \textbf{Bold} entries are best per category and \underline{underlined} are second-best. 
    }
    \vspace{-5pt}
    \label{tab:PCK_Spair}
\end{table*}

\subsection{Correspondence via combined representations}
\label{sec:alpha_mix}

During inference, it is possible to directly query the spherical maps $f_S$ of two images to obtain correspondences using the cosine distance. 
That is, for an image pair $\bm{I}, \bm{I}'$ and a query location $q$ on $\bm{I}$, its location in $\bm{I}'$ can be computed as
\begin{equation}
\label{eq:match}
    p^* = \argmin_p \Gamma(f_S(\bm{I}, q), f_S(\bm{I}', p)). 
\end{equation}
While this protocol is sound, it comes with two important drawbacks. 
First, $f_S$ produces spherical maps for the whole image, including the background, meaning segmentation masks would be required at inference time to prevent the emergence of spurious matches. 
Second, the spherical map is designed to be a smooth parameterization of the object surface, making it susceptible to missing fine details. 
A slightly incorrect mapping $f_S(\bm{I}, q) + \epsilon$ has a high probability of finding a higher match in a nearby region of the correct match $f_S(\bm{I}', p^*)$ if both maps are smooth. 
In comparison, SSL feature maps can exhibit strict feature separation, \eg wheel features are very different from other nearby non-wheel features, making it unlikely that a wheel pixel finds a nearest neighbor among nearby non-wheel pixels. 

To mitigate this issue, we make use of the feature maps that are already computed by the self-supervised backbone network $\phi$, in a similar way to supervised correspondence approaches that aggregate features from a network~\cite{ufer2017deep, min2020learning, zhao2021multi}.
Existing literature has shown that DINO features are particularly effective at differentiating between the foreground and background~\cite{amir2021deep}, removing the need for inference-time segmentation.
To leverage these DINO features, we reformulate \cref{eq:match} as the combination of self-supervised features and spherical locations:
\begin{equation}
    \begin{split}
        p^* = {\argmin}_p (1 - \alpha)\;&\Gamma(\phi(\bm{I}, q), \phi(\bm{I}', p)) \\ + \;\alpha\; &\Gamma(f_S(\bm{I}, q), f_S(\bm{I}', p)),  
    \end{split}
    \label{eqn:comb_match}
\end{equation}
where $\alpha$ is a hyperparameter that balances each term.

\section{Experiments}

\noindent{\bf Implementation details.}
We build our models on DINOv1-B/8~\cite{caron2021emerging} and DINOv2-B/14~\cite{oquab2023dinov2} backbones. 
Training masks are obtained using a pretrained Mask R-CNN model~\cite{he2017mask}.
Our spherical mapper consists of a linear dimension reduction layer halving the number of features, a visual transformer block with self-attention, and a final linear layer with an output dimension of three. The resulting feature map is then normalized pixel-wise so that its output values lies on the surface of $S^2$.
The dimensionality reduction layer, while not strictly necessary, is helpful considering the small number of images used during training, while the transformer block helps in disentangling sides and repeated parts by allowing global reasoning between image patches.

The final loss term is computed as $\mathcal{L} = \mathcal{L}_{rec} + \lambda_{rd} \mathcal{L}_{rd} + \lambda_{o} \mathcal{L}_{o} + \lambda_{vp} \mathcal{L}_{vp}$. We set $\lambda_{rd} = \lambda_{o} = 0.3$, while $\lambda_{vp} = 0.1$ is set slightly lower as it can have the detrimental effect of pulling all predictions towards the average over image pixels - \cf~\cref{eq:vp_loss}. The relative distance margin is set to $\delta=0.5$ to encourage feature separation. 
To perform matching in \cref{eqn:comb_match}, $\alpha$ is set to $0.2$ so that we leverage the highly discriminative DINO features, while using the sphere as a symmetry and repetition-breaking mechanism. 
Our model is trained for a total of 200 epochs using Adam~\cite{kingma2014adam}.

\noindent{\bf Datasets.}
\label{sec:dataset}
Following existing work~\cite{huang2022learning,gupta2023asic,taleof2feats}, we evaluate our method primarily on the Spair-71k dataset~\cite{min2019spair}, which contains images from 18 different object categories. 
The evaluation set contains between 600 and 900 image pairs, each of which are annotated with the keypoints that are visible in both images.
We train our model on the Spair-71k training split on all categories simultaneously, without using any keypoint annotations. 
Additionally, we evaluate on the AwA-pose dataset~\cite{banik2021AwApose}, which contains images from 35 quadruped categories, annotated with keypoints.
We also present results by training on the Freiburg cars dataset~\cite{sedaghat2015unsupervised}. 
This dataset was collected by densely sampling approximately 100 images each around 48 different car instances. 
The higher image count, coupled with 360$^o$ dense sampling, enables us to learn a more detailed spherical prototype.

\begin{figure*}[t]
    \setlength{\tabcolsep}{3pt}
    \resizebox{\textwidth}{!}{
    \begin{tabular}{ccccccccccccccc}
        \rotatebox{90}{\parbox[t]{3cm}{\hspace*{\fill}\Large{Image}\hspace*{\fill}}}\hspace*{5pt}
         & \includegraphics[width=.2\linewidth, height=.2\linewidth]{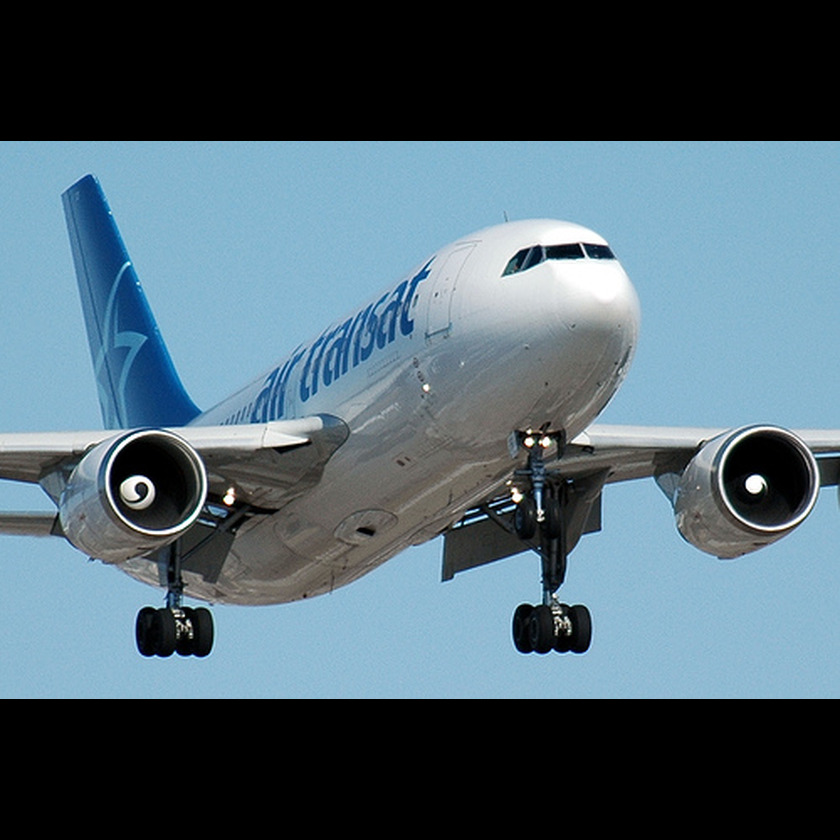}
         & \includegraphics[width=.2\linewidth, height=.2\linewidth]{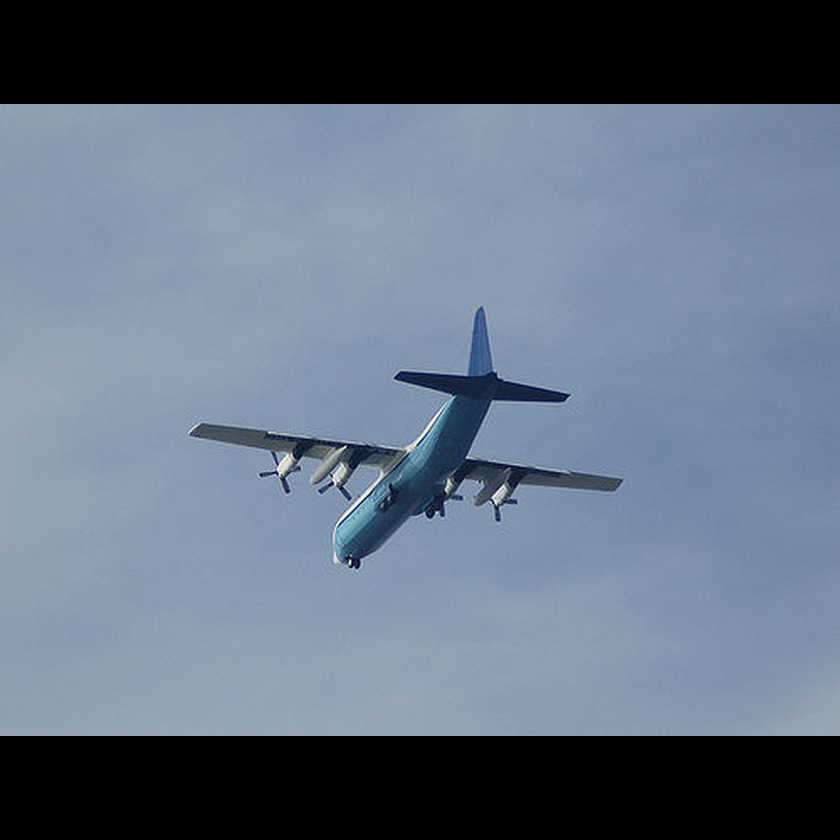}
         & \includegraphics[width=.2\linewidth, height=.2\linewidth]{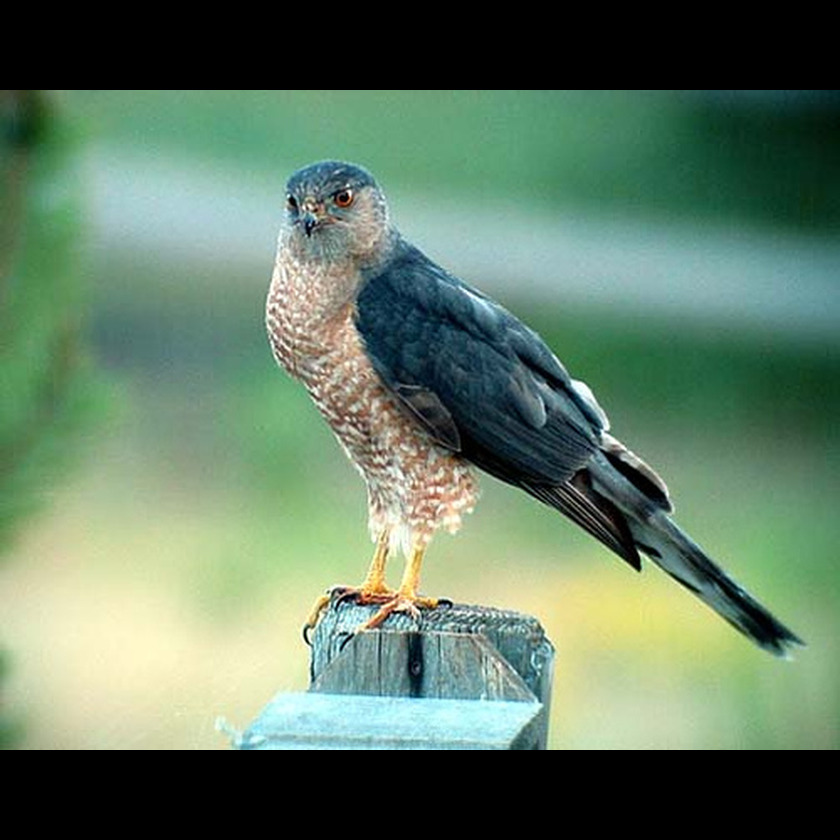}
         & \includegraphics[width=.2\linewidth, height=.2\linewidth]{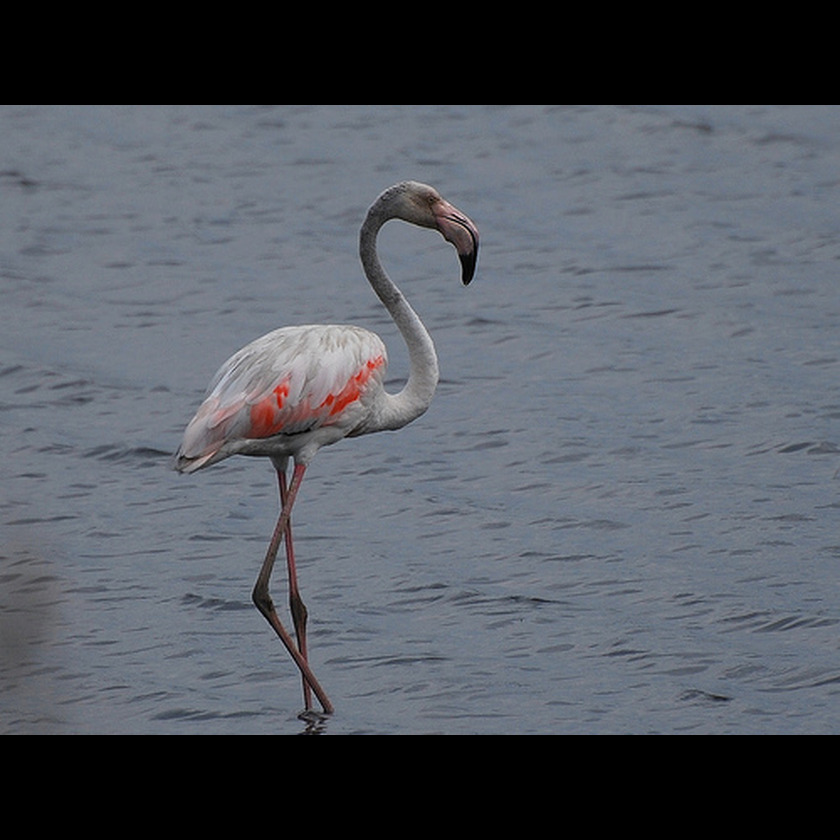}
         & \includegraphics[width=.2\linewidth, height=.2\linewidth]{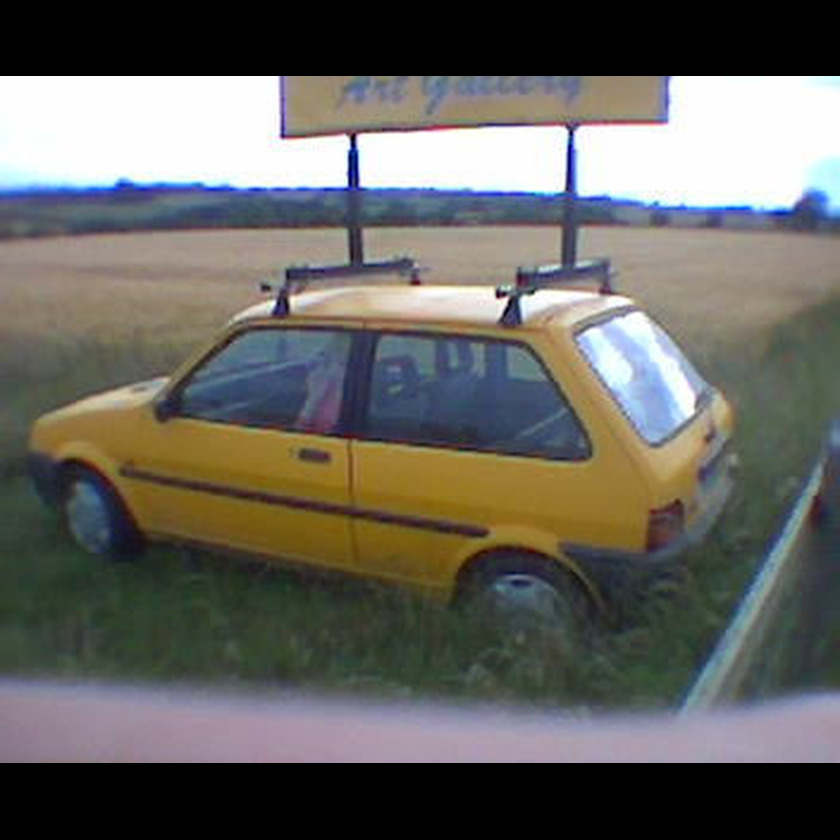}
         & \includegraphics[width=.2\linewidth, height=.2\linewidth]{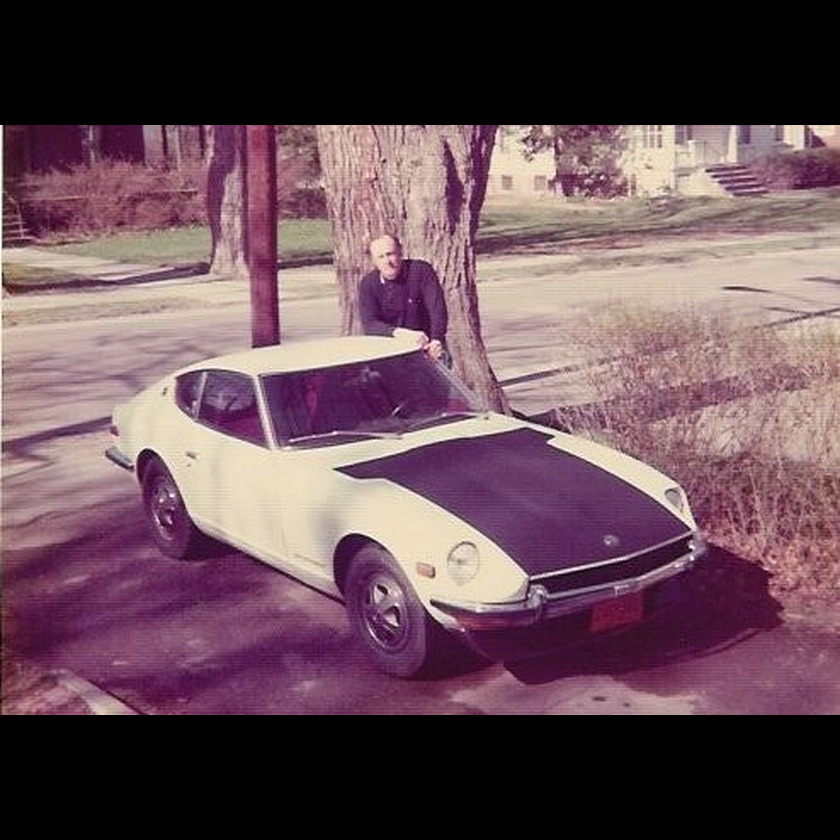}
         & \includegraphics[width=.2\linewidth, height=.2\linewidth]{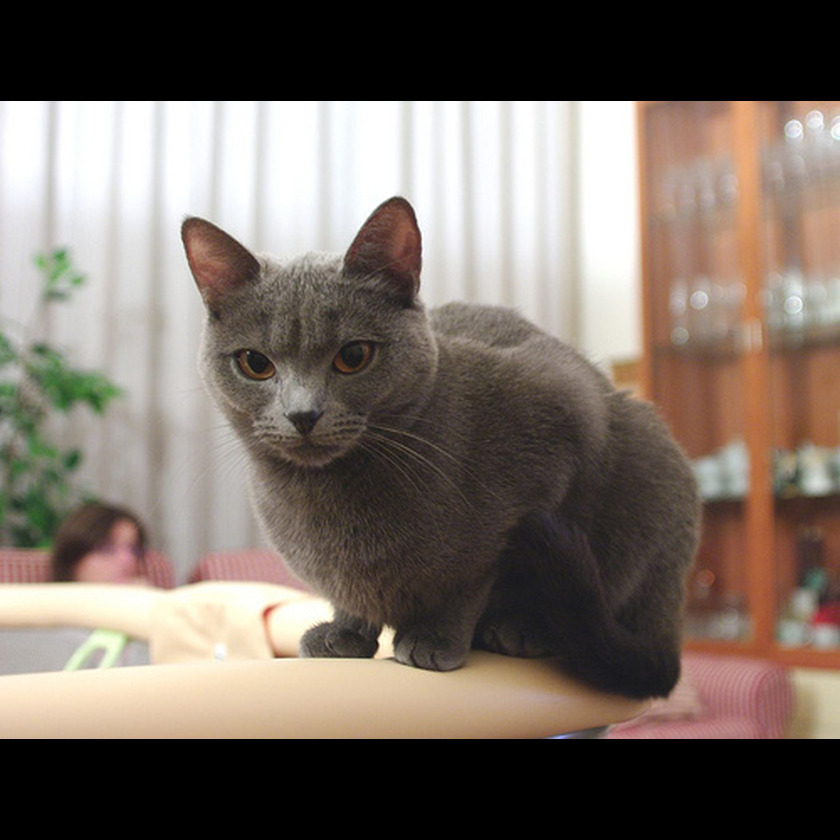}
         & \includegraphics[width=.2\linewidth, height=.2\linewidth]{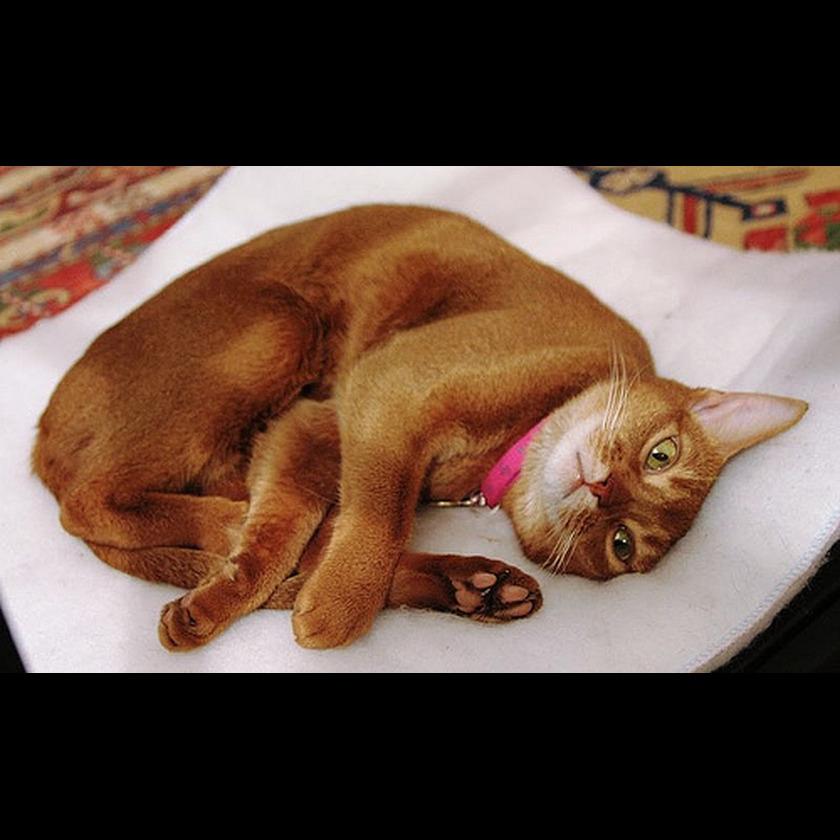}
         & \includegraphics[width=.2\linewidth, height=.2\linewidth]{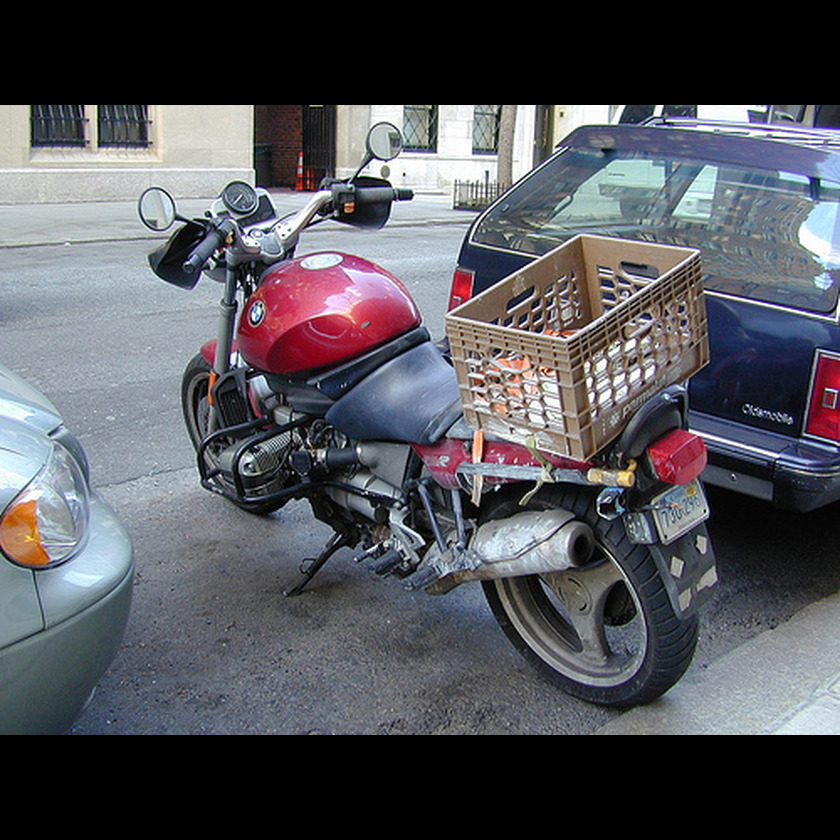}
         & \includegraphics[width=.2\linewidth, height=.2\linewidth]{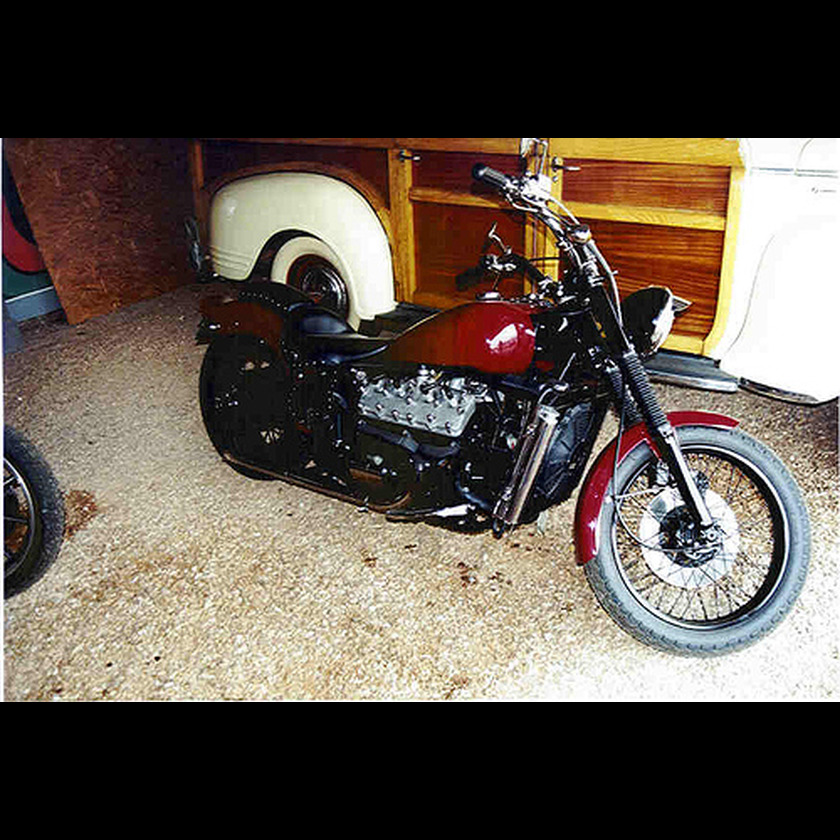}
         & \includegraphics[width=.2\linewidth, height=.2\linewidth]{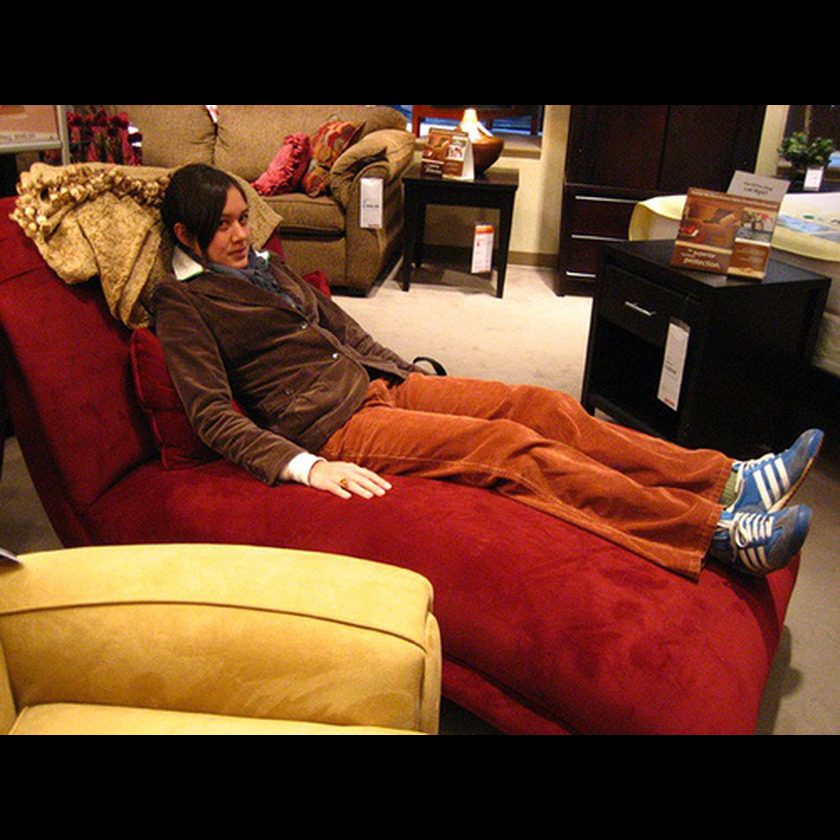}
         & \includegraphics[width=.2\linewidth, height=.2\linewidth]{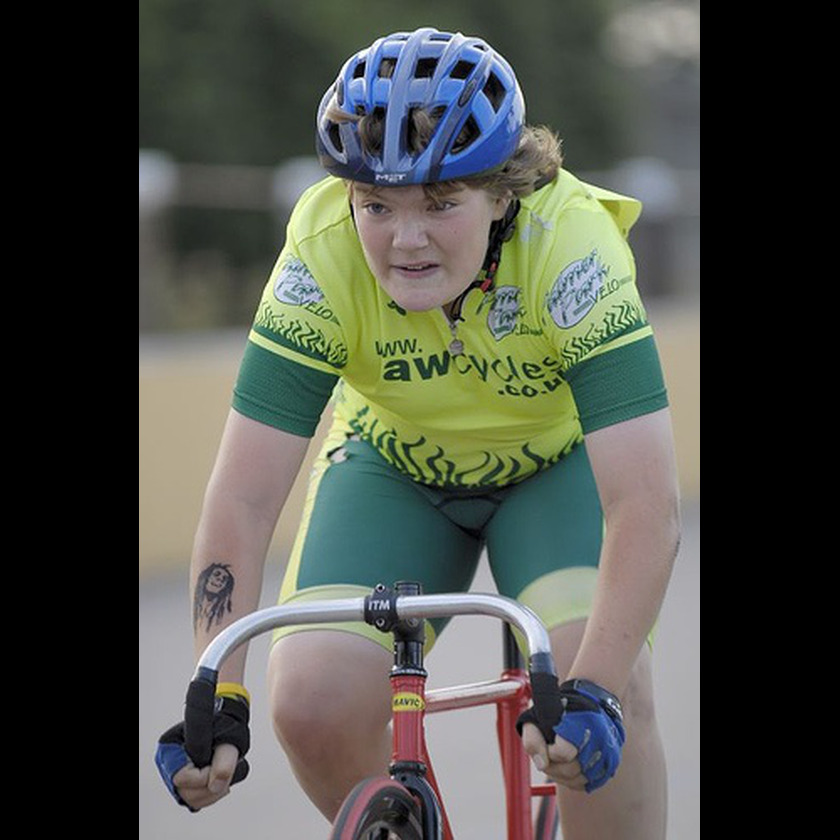}
         \\
        \rotatebox{90}{\parbox[t]{3cm}{\hspace*{\fill}\Large{DINOv2}\hspace*{\fill}}}\hspace*{5pt}
         & \includegraphics[width=.2\linewidth, height=.2\linewidth]{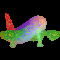}
         & \includegraphics[width=.2\linewidth, height=.2\linewidth]{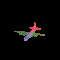}
         & \includegraphics[width=.2\linewidth, height=.2\linewidth]{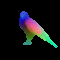}
         & \includegraphics[width=.2\linewidth, height=.2\linewidth]{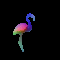}
         & \includegraphics[width=.2\linewidth, height=.2\linewidth]{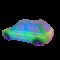}
         & \includegraphics[width=.2\linewidth, height=.2\linewidth]{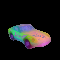}
         & \includegraphics[width=.2\linewidth, height=.2\linewidth]{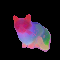}
         & \includegraphics[width=.2\linewidth, height=.2\linewidth]{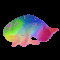}
         & \includegraphics[width=.2\linewidth, height=.2\linewidth]{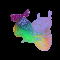}
         & \includegraphics[width=.2\linewidth, height=.2\linewidth]{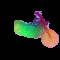}
         & \includegraphics[width=.2\linewidth, height=.2\linewidth]{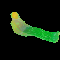}
         & \includegraphics[width=.2\linewidth, height=.2\linewidth]{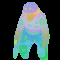}
         \\
        \rotatebox{90}{\parbox[t]{3cm}{\hspace*{\fill}\Large{SD}\hspace*{\fill}}}\hspace*{5pt}
         & \includegraphics[width=.2\linewidth, height=.2\linewidth]{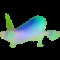}
         & \includegraphics[width=.2\linewidth, height=.2\linewidth]{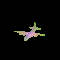}
         & \includegraphics[width=.2\linewidth, height=.2\linewidth]{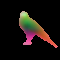}
         & \includegraphics[width=.2\linewidth, height=.2\linewidth]{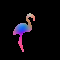}
         & \includegraphics[width=.2\linewidth, height=.2\linewidth]{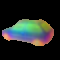}
         & \includegraphics[width=.2\linewidth, height=.2\linewidth]{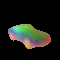}
         & \includegraphics[width=.2\linewidth, height=.2\linewidth]{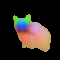}
         & \includegraphics[width=.2\linewidth, height=.2\linewidth]{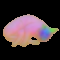}
         & \includegraphics[width=.2\linewidth, height=.2\linewidth]{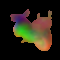}
         & \includegraphics[width=.2\linewidth, height=.2\linewidth]{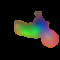}
         & \includegraphics[width=.2\linewidth, height=.2\linewidth]{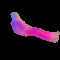}
         & \includegraphics[width=.2\linewidth, height=.2\linewidth]{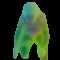}
         \\
        \rotatebox{90}{\parbox[t]{3cm}{\hspace*{\fill}\Large{DINOv2+SD}\hspace*{\fill}}}\hspace*{5pt}
         & \includegraphics[width=.2\linewidth, height=.2\linewidth]{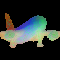}
         & \includegraphics[width=.2\linewidth, height=.2\linewidth]{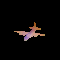}
         & \includegraphics[width=.2\linewidth, height=.2\linewidth]{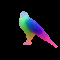}
         & \includegraphics[width=.2\linewidth, height=.2\linewidth]{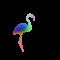}
         & \includegraphics[width=.2\linewidth, height=.2\linewidth]{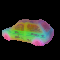}
         & \includegraphics[width=.2\linewidth, height=.2\linewidth]{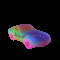}
         & \includegraphics[width=.2\linewidth, height=.2\linewidth]{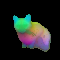}
         & \includegraphics[width=.2\linewidth, height=.2\linewidth]{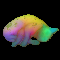}
         & \includegraphics[width=.2\linewidth, height=.2\linewidth]{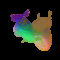}
         & \includegraphics[width=.2\linewidth, height=.2\linewidth]{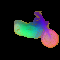}
         & \includegraphics[width=.2\linewidth, height=.2\linewidth]{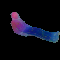}
         & \includegraphics[width=.2\linewidth, height=.2\linewidth]{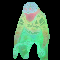}
         \\
        \rotatebox{90}{\parbox[t]{3cm}{\hspace*{\fill}\Large{Sphere}\hspace*{\fill}}}\hspace*{5pt}
         & \includegraphics[width=.2\linewidth, height=.2\linewidth]{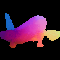}
         & \includegraphics[width=.2\linewidth, height=.2\linewidth]{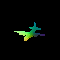}
         & \includegraphics[width=.2\linewidth, height=.2\linewidth]{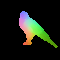}
         & \includegraphics[width=.2\linewidth, height=.2\linewidth]{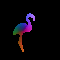}
         & \includegraphics[width=.2\linewidth, height=.2\linewidth]{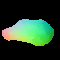}
         & \includegraphics[width=.2\linewidth, height=.2\linewidth]{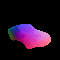}
         & \includegraphics[width=.2\linewidth, height=.2\linewidth]{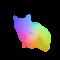}
         & \includegraphics[width=.2\linewidth, height=.2\linewidth]{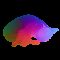}
         & \includegraphics[width=.2\linewidth, height=.2\linewidth]{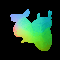}
         & \includegraphics[width=.2\linewidth, height=.2\linewidth]{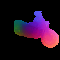}
         & \includegraphics[width=.2\linewidth, height=.2\linewidth]{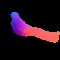}
         & \includegraphics[width=.2\linewidth, height=.2\linewidth]{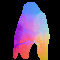}
         \\
    \end{tabular}
    }
    \vspace{-5pt}
    \caption{Qualitative comparison of dense correspondence maps. 
    For DINOv2, SD, and DINO+SD features we perform PCA on the segmented object features independently for each category, then visualize the three main components. Note that the SD and DINO+SD features are not completely equivalent to the ones used to compute matches, but are provided here for illustration.
    Spherical maps from $f_S$ (Sphere) for our approach are visualized directly. 
    Our spherical maps correctly identify the different sides of objects, whereas other features fail to capture these differences. 
    }
    \label{fig:Spair_maps}
    \vspace{-12pt}
\end{figure*}

\subsection{Keypoint average precision} 
While issues with symmetries are clear in \cref{fig:Spair_maps}, we argue that the PCK metric fails to account for them as it is only computed between keypoints that appear in \emph{both} the source and target image. 
Hence, models that predict high similarity between two visually similar but semantically different keypoints might not be penalized for it. 
PCK can partly account for repetition-related mistakes if the parts appear simultaneously in both images, but it does not penalize symmetry-related issues, like mistaking the left and right side of a car, as the corresponding keypoints do not appear simultaneously. PCK was initially proposed for simpler cases where keypoints were assumed to be always be visible, \eg faces or different instances in the same pose.
However, as benchmarks become more challenging, a more robust protocol should be used.
To illustrate how much these failure cases impact SSL models, we propose an alternative evaluation metric, Keypoint Average Precision - KAP@$\kappa$.
It is similar to PCK@$\kappa$, but accounts for these errors by evaluating average precision instead of accuracy and penalizes methods that predict matches when none exist. 

When performing evaluation on a pair of images, instead of restricting it to keypoints that simultaneously appear in both images, we consider \emph{all} keypoints that appear in the source, irrespective of whether they are also visible in the target image. 
For each such keypoint, if it appears in the target image, we extract the highest similarity value within radius $\kappa$ of the ground-truth and label it as positive. 
We also extract the highest similarity value \emph{outside} the radius and label it as negative, penalizing overly high predictions on wrong, possibly repeated parts. 
If the source keypoint does not appear on the target image, we simply extract the highest similarity prediction across the whole target image and label it as negative to penalize incorrectly high similarity when none should exist.
This effectively reformulates the task as a binary classification problem, where the embedding of each source keypoint has to be close to, and only to, the corresponding target keypoint. 
Finally, we compute the average precision over all the extracted pairs, and report mean average precision per-category.

\subsection{SPair-71k results}
We first evaluate on the task of keypoint transfer between pairs of images using Spair-71k~\cite{min2019spair}. 
We present quantitative per-category results in \cref{tab:PCK_Spair}, using the standard PCK metric along with \emph{macro}-averaged scores as the number of  per-category pairs varies greatly. 
We split the models according to their backbone: custom supervised models, DINOv1-based, SD-based, and DINOv2-based. 
A striking preliminary observation is the effectiveness of DINOv2, surpassing supervised approaches on most categories. 
While its success can partially be attributed to evaluation biases~\cite{aygun2022demystifying}, this  demonstrates the effectiveness of large scale learned self-supervised features. 

Using a DINOv1 backbone, our model improves over its backbone on most categories, and over ASIC~\cite{gupta2023asic} on average.
When combined with the DINOv2 backbone, our model provides improvements on all but two categories. 
A clear pattern can be identified, with strongest improvements being observed over blob-like symmetric objects with repeated parts (+21.0 on bus and +15.7 on car), while spherical mapping hinders performance on highly deformable categories (-4.3 on person), or high-genus objects for which a sphere is a poor surface prior (-0.8 on bicycle).
Our model performs slightly better compared to DINO+SD~\cite{taleof2feats}, while requiring a fraction of the computational cost at inference time.
Finally, following~\cite{taleof2feats} we evaluate adding Stable Diffusion (SD)~\cite{rombach2022high} features to our method at inference time, which yields further improvements. 
This illustrates that spherical maps and SD features capture different facets of the correspondence problem. 

\begin{table*}[t]
    \centering
    \renewcommand{\b}{\bfseries}
    \renewcommand{\u}{\rlap{\underline{\phantom{00.0}}}}
    \resizebox{\textwidth}{!}{
    \begin{tabular}{l|rrrrrrrrrrrrrrrrrr|r}
            & \faIcon{plane} & \faIcon{bicycle} & \faIcon{crow} & \faIcon{ship} & \faIcon{wine-bottle} & \faIcon{bus} & \faIcon{car} & \faIcon{cat} & \faIcon{chair} & \Cow & \faIcon{dog} & \faIcon{horse} & \faIcon{motorcycle} & \faIcon{walking} & \Plant & \Sheep & \faIcon{train} & \faIcon{tv} & avg\\ 
        \midrule
        
         DINOv2~\cite{oquab2023dinov2}
            & 53.5 & 54.0 & 60.2 & 35.5 & 44.4 & 36.3 & 31.7 & 61.3 & 37.4 & 54.7 & 52.5 & 51.5 & 48.8 & 48.2 & 37.8 & 44.1 & 47.4 & 38.2 & 46.5\\
         SD~\cite{taleof2feats} 
            & 44.4 & 48.5 & 54.5 & 31.5 & 45.2 & 32.7 & 30.0 & 68.4 & 35.8 & 55.2 & 47.9 & 48.1 & 44.8 & 42.3 &\u 44.5 & 39.2 & 52.7 &\u 51.2 & 45.4\\
         DINOv2 + SD~\cite{taleof2feats} 
            & 52.0 &\b 55.9 & 59.2 & 34.7 &\b 49.0 & 36.0 & 32.5 & 70.3 & 39.8 & 59.8 & 53.1 & 52.4 & 50.6 &\b 50.4 &\b 47.8 & 46.2 & 53.3 & 49.8 & 49.6\\    
          Ours (sphere only)
            & 38.4 & 34.2 & 53.9 & 33.0 & 37.9 & 49.7 & 43.4 &\u 71.7 & 29.8 & 57.1 & 45.8 & 42.5 & 32.4 & 27.0 & 29.5 & 37.1 &\u 57.4 & 36.0 & 42.1\\
          Ours
            &\b 60.7 & 51.2 &\b 63.1 &\u 38.4 & 45.0 &\b 55.9 &\u 45.7 & 69.7 &\u 40.4 &\u 63.2 &\u 54.8 &\u 54.3 &\u 51.2 &\u 48.7 & 38.8 &\u 47.9 & 55.5 & 42.2 &\u 51.5\\
         Ours + SD
            &\u 58.9 &\u 54.2 &\u 62.2 &\b 39.6 &\u 46.6 &\u 54.5 &\b 47.1 &\b 76.2 &\b 40.9 &\b 65.3 &\b 57.3 &\b 56.1 &\b 54.2 & 47.4 & 43.7 &\b 49.4 &\b 62.4 &\b 52.0 &\b 53.8\\
    \end{tabular}
    }
    \vspace{-5pt}
    \caption{Keypoint matching scores on SPair-71k evaluated using KAP@0.1 with macro-averaging for the summary scores.  
    }
    \vspace{-5pt}
    \label{tab:KAP_Spair}
\end{table*}

We show the results of evaluating correspondences with KAP in \cref{tab:KAP_Spair}. 
We observe particularly large discrepancies between PCK and KAP in highly symmetric objects with repeated parts, \eg buses and cars, and the ranking between our model and DINOv2 is reversed on humans, possibly due to DINOv2's symmetric predictions on left and right limbs. All models across all categories obtain lower results when using KAP highlight the symmetry issues they are facing.
A limitation of KAP is that it might still count correct repeated parts mistakes if their distance in the image is less than $\kappa$. 
Nonetheless, we argue that KAP is more informative than PCK and is just as efficient to evaluate.

The qualitative visualization shown in \cref{fig:Spair_maps} illustrates these improved results. 
Our model correctly maps the two opposite sides of the cars to different sphere regions, while other feature maps look extremely similar, even when looking at two opposite sides of an object, with the notable exception of SD on birds. 
The fact that such similarities exist between opposite views suggest that the utility of baselines such as DINOv2 would be quite limited in real scenarios.

\subsection{Additional evaluation}

\noindent{\bf Animals with Attributes (AwA).}
To further evaluate our model, we also report results on the AwA-pose dataset from~\cite{banik2021AwApose}, which contains images from 35 different categories of quadruped. 
While our spherical prototype needs category information during training to reconstruct category-specific features, only the sphere mapper is used during inference, meaning it can be readily applied to unseen categories. 
Five quadruped classes (cat, cow, dog, horse, and sheep) appear in both SPair and AwA-pose, and are therefore seen during training, but the rest are completely new for our sphere mapper. 
Nonetheless, results shown in \cref{tab:AwA} show that the sphere mapper generalizes well to unseen categories. 
We attribute this to the assistance from the strong category-agnostic features from DINOv2. 
Once again, the improvement of using our spherical method is much more apparent using KAP, \ie the PCK improvement over DINOv2 alone is 2.2 when adding SD and 2.6 when adding spheres, while these become 1.8 and 3.9 respectively with KAP.

\begin{table}[htb]
    \centering
    \vspace{-5pt}
    \resizebox{.9\columnwidth}{!}{
    \begin{tabular}{r|ccccc}
                 & Dv2  & SD   & Dv2+SD & Ours & Ours+SD\\ \hline
        PCK@0.1 & 65.9 & 56.0 & 68.1 & 68.7 & 69.8 \\
        KAP@0.1 & 55.0 & 50.7 & 56.8 & 58.9 & 60.6 \\

    \end{tabular}
    }
    \vspace{-10pt}   
    \caption{Average scores when evaluating on AwA-pose using a random subset of 200 pairs per category. 
    `Ours' denotes our model trained on SPair-71k with a DINOv2 backbone.
    }
    \label{tab:AwA}
\end{table}

\noindent{\bf Model ablation.} We also analyze the impact of removing each individual loss term when training our model in an ablation study. 
The results in \cref{tab:ablations} indicate that all variants except the no $\mathcal{L}_{rd}$ one do not significantly improve over the DINOv2 baseline as they also fail to identify opposite sides of objects and repeated parts.

\begin{table}[htb]
    \centering
    \vspace{-5pt}
    \resizebox{0.9\columnwidth}{!}{
    \begin{tabular}{ccccc}
        DINOv2 & no $\mathcal{L}_{vp}$ & no $\mathcal{L}_{rd}$ & no $\mathcal{L}_{o}$ & full model\\ \hline
        56.2 & 58.6 & 61.2 & 56.0 & 63.6\\

    \end{tabular}
    }
    \vspace{-10pt}   
    \caption{Average PCK@0.1 scores when training and testing on SPair-71k using different ablated versions of our approach. 
    }
    \label{tab:ablations}
\end{table}

\noindent{\bf Inference speed.}  
The results in \cref{tab:PCK_Spair} demonstrate how effective the addition of Stable Diffusion features are. 
However, computing correspondences through diffusion requires adding noise to images, running iterative diffusion, and performing joint decomposition between the source and target features. 
As a result, its inference speed is much slower compared to using DINOv2 features. 
In comparison, our approach only requires a few extra layers on top of DINOv2 and has negligible additional overhead. 
Timing results in \cref{tab:throughput} illustrate that adding diffusion-extracted features reduces  throughput by an order of magnitude. 

\begin{table}[htb]
    \centering
    \vspace{-5pt}
    \resizebox{.9\columnwidth}{!}{
    \begin{tabular}{ccccc}
         DINOv2 & SD   & DINOv2+SD & Ours & Ours+SD\\ \hline
         3.4 & 0.38 & 0.35 & 3.3 & 0.34\\
    \end{tabular}
    }
    \vspace{-10pt}   
    \caption{Descriptor computation throughput in pairs/second at inference time on a single A5000 GPU, where higher is better. 
    }
    \label{tab:throughput}
\end{table}

\subsection{Limitations} 
While our approach results in significant improvements over the DINOv2 baseline for many categories (see \cref{tab:PCK_Spair}), there are still some limitations. 
For example, we do not perform as well on a small number of categories that cannot be well approximated by a sphere (\eg humans). 
Furthermore, the very low-dimensional spherical encoding on its own is not as effective at discriminating parts and sub-parts and thus needs to be combined with self-supervised features. 
However, as we already densely compute these features at inference time for each test image, there is no additional cost in combining them with our spherical mapping when evaluating correspondence. 
Finally, our method relies on additional supervision at training time in the form of automatically computed segmentation masks and weak camera pose supervision. 
However, results in the supplementary material suggest that coarse camera viewpoints are enough to obtain good performance. 
\section{Conclusion}
We presented a new approach for semantic correspondence estimation that is robust to issues resulting from object symmetries and repeated part instances. 
In addition, we proposed a new evaluation protocol that is more sensitive to these issues. 
Our approach leverages recent advances in the self-supervised learning of discriminative image features and combines them with a weak geometric spherical prior. 
By using weak pose supervision to represent objects categories on the surface of spheres, we are able to enforce simple geometric constraints during training that result in more geometric and semantically consistent representations. 
Results on the task of semantic keypoint matching between images from different object instances, a fundamental task in 3D computer vision, demonstrate that our approach alone is superior to existing strong baselines and can be combined with existing methods to further boost performance.

\vspace{4pt}
{
\noindent{\bf Acknowledgments.} This project was supported by the EPSRC Visual AI grant EP/T028572/1. 
}

{
   \small
   \bibliographystyle{ieeenat_fullname}
   \bibliography{main}
}

\clearpage
\appendix
\setcounter{table}{0}
\renewcommand{\thetable}{A\arabic{table}}
\setcounter{figure}{0}
\renewcommand{\thefigure}{A\arabic{figure}}

\noindent{\LARGE Appendix}
\vspace{5pt}

\noindent In this supplementary document, we provide results on three additional datasets: PF-PASCAL~\cite{ham2017proposal}, TSS~\cite{taniai2016TSS}, and Freiburg cars~\cite{sedaghat2015unsupervised}. 
We also visualize feature maps on SPair-71k~\cite{min2019spair} and show the results of keypoint matches. 

\section{Implementation details}
\begin{figure*}
    \centering
    \includegraphics[width=\linewidth]{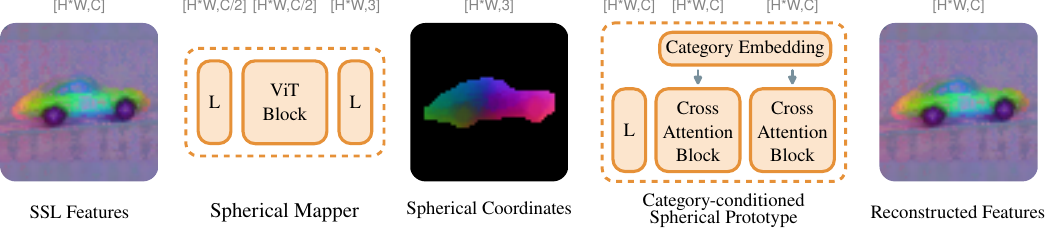}
    \caption{Architecture details for our sphere mapper and spherical prototype. C denotes the dimension of the SSL embedding, and block marked with L are simple linear layers used to change dimensionality.}
    \label{fig:arch_detail}
\end{figure*}

For the results labeled as `Ours+SD' (\eg in Tab.~\textcolor{red}{1} in the main paper), this model use our spherical mapper but at inference time also combines the DINOv2 and Stable Diffusion (SD) features, applying Eq.~(7) with the fused DINO+SD features in place of $\phi$.  We also show architeture details in \cref{fig:arch_detail}

\section{Continuous surface embeddings}

\begin{table}[h]
    \centering
    \renewcommand{\b}{\bfseries}
    \renewcommand{\u}{\rlap{\underline{\phantom{00.0}}}}
    \resizebox{.8\linewidth}{!}{
    \begin{tabular}{cl|rrrrrr}
            && \faIcon{cat} & \Cow & \faIcon{dog} & \faIcon{horse} & \faIcon{walking} & \Sheep \\ 
        \midrule
        \multirow{3}{*}{\rotatebox{90}{\parbox[t]{35pt}{\hspace*{\fill}PCK\hspace*{\fill}}}}
        & CSE
            &   28.1 &   23.4 &   24.1 &   25.5 &   86.4 &   12.1\\
        & Ours
            &   80.0 &   80.2 &   66.7 &   71.2 &   63.9 &   68.6\\
        & Ours + SD
            &   82.4 &   82.0 &   67.3 &   73.9 &   60.0 &   69.8\\
        \midrule
        \multirow{3}{*}{\rotatebox{90}{\parbox[t]{35pt}{\hspace*{\fill}KAP\hspace*{\fill}}}}
        & CSE
            &   36.7 &   30.2 &   32.8 &   32.8 &   72.2 &   27.1\\
        & Ours
            &   69.7 &   63.2 &   54.8 &   54.3 &   48.7 &   47.9\\
        & Ours + SD
            &   76.2 &   65.3 &   57.3 &   56.1 &   47.4 &   49.4\\
    \end{tabular}
    }
    \caption{Keypoint matching scores on SPair-71k evaluated using PCK@0.1 and KAP@0.1 for CSE~\cite{neverova2020continuous} and our models.
    }
    \label{tab:CSE_Spair}
\end{table}

Although not directly comparable, the spherical embedding learned by our model is similar to Continuous Surface Embeddings (CSE)~\cite{neverova2020continuous}, in that it aims to densely represent points on an object's surface in a smooth way. However, CSE is learned in a fully supervised way, using images densely annotated with correspondences to 3D meshes of object categories. In comparison, we only rely on viewpoint and weak geometric priors. A consequence of this is that our sphere mappings do not necessarily converge to a unique solution, for instance, applying a random rotation to it would still satisfy the geometric constraints that we use during training. This makes it challenging to evaluate our spheres under CSE protocol, as recovering the transformation between them and ground truth meshes is non trivial. Still, CSE can easily be evaluated by keypoint metrics on SPair-71k for the categories on which it was trained by using the publicly released implementation. The PCK and KAP@0.1 are shown in~\cref{tab:CSE_Spair}. Despite being supervised, CSE only outperforms our approach on humans, most likely because it is the only category with enough high-quality annotations. Another interesting results is that CSE suffers a much smaller drop in performance when evaluated with KAP instead of PCK, as its fully supervised regime makes it more robust to object symmetries and repetitions, constituting a strong argument in favor of using KAP when evaluating correspondences. 

\section{Additional ablation results}
We investigate the balance between the different geometric losses to assess the sensitivity of our model. In \cref{tab:geo_losses_weights}, we show PCK@0.1 scores when training and testing on SPair-71k where we set the losses to the same weight, or globally increase or decrease the weight using the following color-coding: \red{$\lambda_{rd}$}, \gre{$\lambda_{o}$}, and \blu{$\lambda_{vp}$}. We observe that altering the losses balance slightly decreases performances, but all models consistently beat the DINOv2 backbone by a large margin.

\begin{table}[t]
    \centering
    \resizebox{\columnwidth}{!}{
    \begin{tabular}{cccccc}
        DINOv2 & \red{0.1}, \gre{0.1}, \blu{0.1} & \red{0.3}, \gre{0.3}, \blu{0.3} & \red{1}, \gre{1}, \blu{0.3} & \red{0.1}, \gre{0.1}, \blu{0.03} & \red{0.3}, \gre{0.3}, \blu{0.1}\\ \hline
        56.2 & 62.2 & 62.5 & 61.4 & 62.0 & 63.6\\

    \end{tabular}
    }
    \caption{Average PCK@0.1 on SPair-71k for different values of gemetric losses.}
    \label{tab:geo_losses_weights}
\end{table}

\section{Additional datasets}

\noindent{\bf Freiburg cars.}
Spair-71k training data is relatively limited, \ie it contains approximately only 50 training images per category. 
In order to explore the behavior of our model when more data is available, we trained it using images from the Freiburg cars dataset~\cite{sedaghat2015unsupervised}. 
Freiburg cars contains 46 scenes each centered around a single car, and there is an average of 120 images sampled from 360$^\circ$ around each car.
As it comes with precise viewpoint annotations, we can use it to study the sensitivity of our model to the granularity of the viewpoint supervision. 
We discretize the camera viewpoint supervision into different numbers of discrete bins (\eg four bins would correspond to the camera viewing the car from the front, back, and two sides) and evaluate these models on Spair-71k car test pairs. 

Our model trained and tested on SPair-71k from Tab.~\textcolor{red}{1} in the main paper obtains at PCK@0.1 of 67.2 on cars. 
The results in \cref{tab:fc_vp} show that there is no significant benefit from having even finer-grained viewpoint supervision beyond a certain number of bins. 
The best performing model trained on Freiburg cars improves PCK@0.1 by 4.6 points compared to SPair-71k training. 
This illustrates the potential of adding additional training data even when the  viewpoint supervision is coarse.   

As Freiburg cars scenes are densely sampled, we can also use them to qualitatively assess the consistency of feature maps under viewpoint changes. Images in \cref{fig:Fcar} show strong consistency of the maps across the whole viewpoint range, while maintaining semantic consistence between visually different instances for our sphere mapper. Note, our results in \cref{fig:Fcar} are for our model trained on SPair-71k. 

\begin{table}[t]
    \centering
    \resizebox{0.95\columnwidth}{!}{
    \begin{tabular}{rrrrrrrr}
        \# bins  &    4 &    8 &   16 &   32 &   64 &  128 & 360\\ \hline
        PCK@0.1  & 60.1 & 71.8 & 71.1 & 71.2 & 69.2 & 68.1 & 71.0\\
    \end{tabular}
    }
    \vspace{-5pt}
    \caption{Impact of viewpoint supervision granularity. 
    Here we train with coarse-to-finer discretized poses from Freiburg cars and evaluate on the car category in SPair-71k. 
    Only when using very few bins (\ie four) does the performance significantly drop.
    This indicates that our approach is capable of training on relatively weak pose supervision. 
    For context, for the results in the main paper, we use the eight viewpoint bins provided by the SPair-71k annotations.
    }
    \label{tab:fc_vp}
    \vspace{-5pt}
\end{table}

\begin{table}[t]
    \centering
    \renewcommand{\b}{\bfseries}
    \renewcommand{\u}{\rlap{\underline{\phantom{00.0}}}}
    \resizebox{\columnwidth}{!}{
    \begin{tabular}{l|ccc|cccc}
    & \multicolumn{3}{c|}{PF-PASCAL, PCK@$\kappa$} & \multicolumn{4}{c}{TSS, PCK@0.05} \\ 
    & 0.15 & 0.10 & 0.05 & FG3DCar & JODS & Pascal & avg\\
        \midrule
        
         DINOv2~\cite{oquab2023dinov2}    & 61.1 & 77.3 & 83.3 & 82.8 & 73.9 & 53.9 & 72.0\\
         SD~\cite{taleof2feats}           & 61.0 & 83.3 & 86.3 & 93.9 & 69.4 & 57.7 & 77.7\\
         DINOv2 + SD~\cite{taleof2feats}  &\u 73.0 &\u 86.1 &\u 91.1 &\u 94.3 & 73.2 &\u 60.9 &\u 79.7\\    
         Ours                             & 66.2 & 83.9 & 90.2 & 83.1 &\u 74.1 & 54.4 & 75.5\\
         Ours + SD                        &\b 74.0 &\b 88.4 &\b 92.6 &\b 95.3 &\b 78.7 &\b 64.2 &\b 82.3\\
    \end{tabular}
    }
    \vspace{-5pt}
    \caption{Scores for PF-PASCAL and TSS.}
    \label{tab:PF-TSS}
    \vspace{-10pt}
\end{table}

\noindent{\bf PF-PASCAL \& TSS.} We also evaluate our model on PF-PASCAL~\cite{ham2017proposal} and TSS~\cite{taniai2016TSS}. 
As these sets exhibit less challenging pose variations compared with SPair-71k, the benefit of using spherical maps is more limited, as it can only help separate repeated parts that appear in the  same image and not issues due to large pose variation (as they are not present). 
Nonetheless the results in \cref{tab:PF-TSS} indicate that our spherical maps yield consistent improvements.

\vspace{-5pt}
\section{Additional qualitative results and failure cases}
\label{sec:kpts_match}
\begin{figure*}[t]
    \centering
    \resizebox{\textwidth}{!}{
    \begin{tabular}{ccccc}
        \rotatebox{90}{\parbox[t]{3cm}{\hspace*{\fill}\Large{DINOv2}\hspace*{\fill}}}\hspace*{1pt}
         & \includegraphics[width=.4\linewidth, height=.2\linewidth,trim={2cm 1cm 2cm 1cm}, clip]{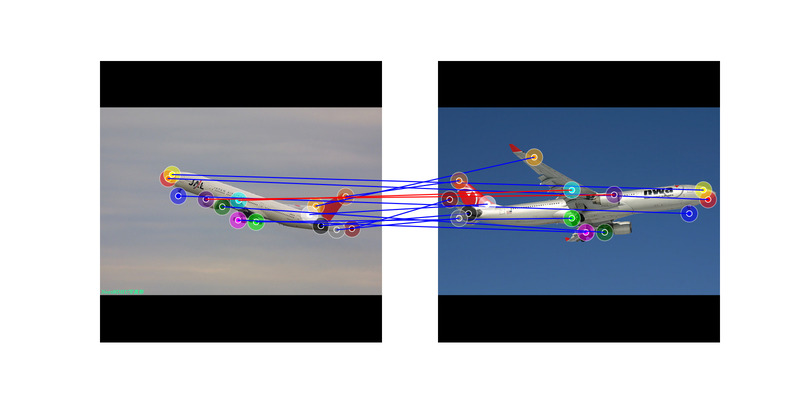}
         & \includegraphics[width=.4\linewidth, height=.2\linewidth,trim={2cm 1cm 2cm 1cm}, clip]{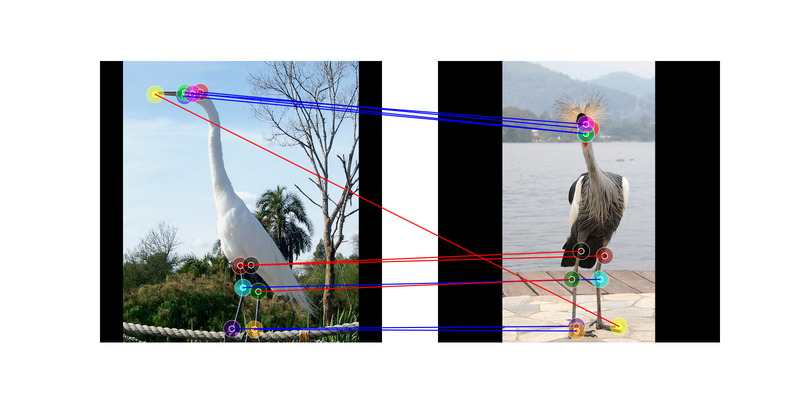}
         & \includegraphics[width=.4\linewidth, height=.2\linewidth,trim={2cm 1cm 2cm 1cm}, clip]{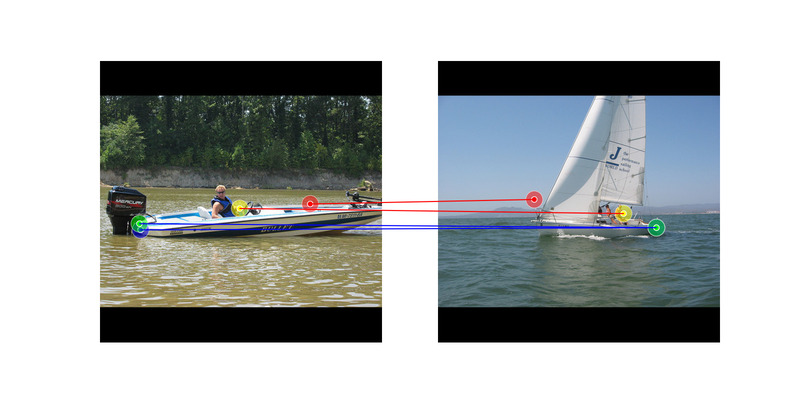}
         & \includegraphics[width=.4\linewidth, height=.2\linewidth,trim={2cm 1cm 2cm 1cm}, clip]{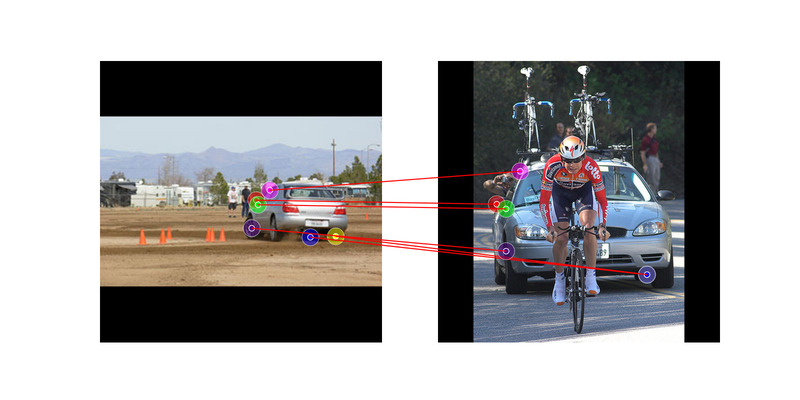}\\
        \rotatebox{90}{\parbox[t]{3cm}{\hspace*{\fill}\Large{SD}\hspace*{\fill}}}\hspace*{1pt}
         & \includegraphics[width=.4\linewidth, height=.2\linewidth,trim={2cm 1cm 2cm 1cm}, clip]{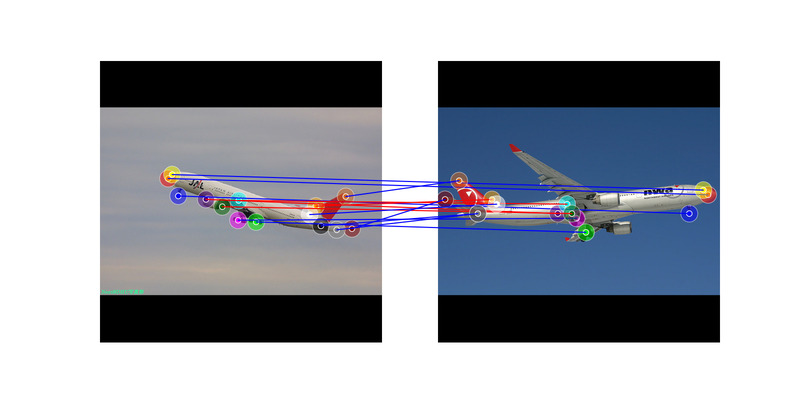}
         & \includegraphics[width=.4\linewidth, height=.2\linewidth,trim={2cm 1cm 2cm 1cm}, clip]{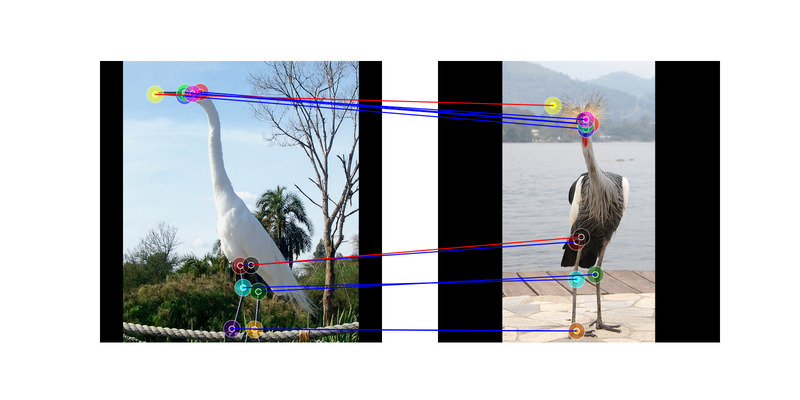}
         & \includegraphics[width=.4\linewidth, height=.2\linewidth,trim={2cm 1cm 2cm 1cm}, clip]{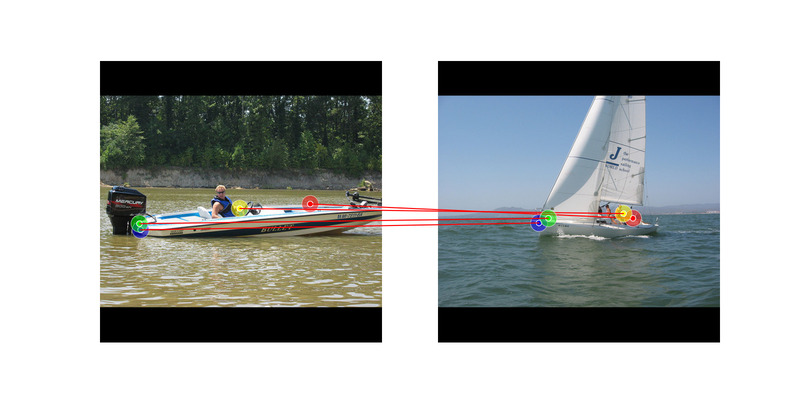}
         & \includegraphics[width=.4\linewidth, height=.2\linewidth,trim={2cm 1cm 2cm 1cm}, clip]{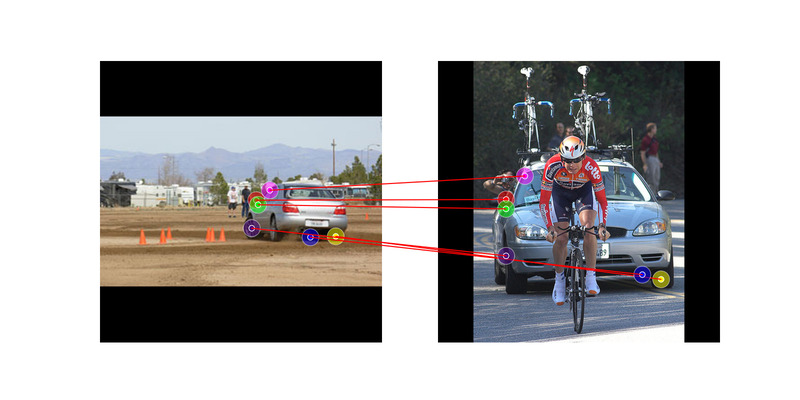}\\
        \rotatebox{90}{\parbox[t]{3cm}{\hspace*{\fill}\Large{DINOv2+SD}\hspace*{\fill}}}\hspace*{1pt}
         & \includegraphics[width=.4\linewidth, height=.2\linewidth,trim={2cm 1cm 2cm 1cm}, clip]{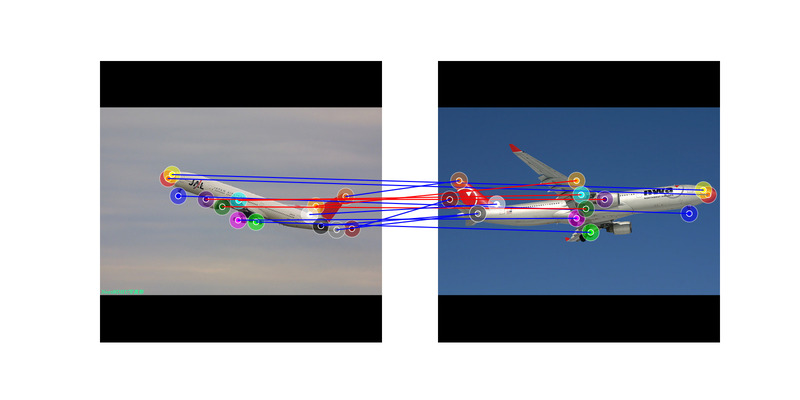}
         & \includegraphics[width=.4\linewidth, height=.2\linewidth,trim={2cm 1cm 2cm 1cm}, clip]{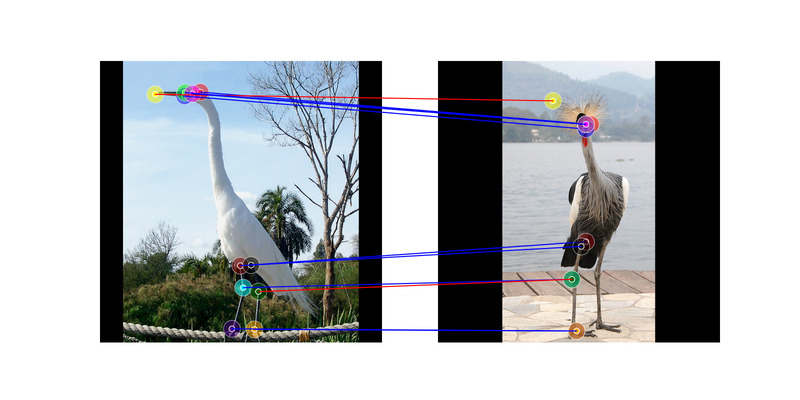}
         & \includegraphics[width=.4\linewidth, height=.2\linewidth,trim={2cm 1cm 2cm 1cm}, clip]{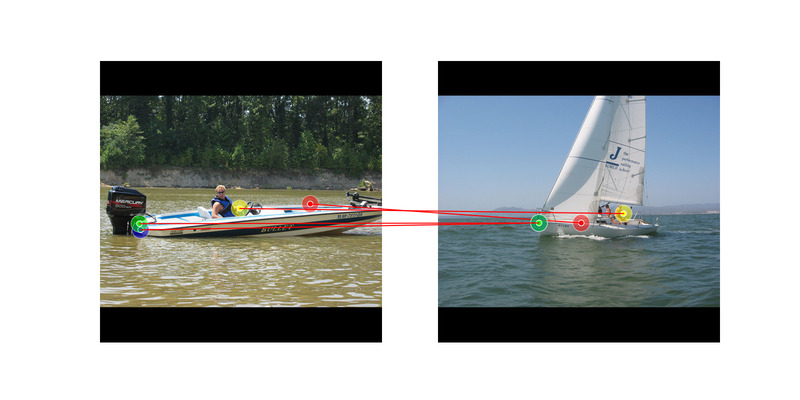}
         & \includegraphics[width=.4\linewidth, height=.2\linewidth,trim={2cm 1cm 2cm 1cm}, clip]{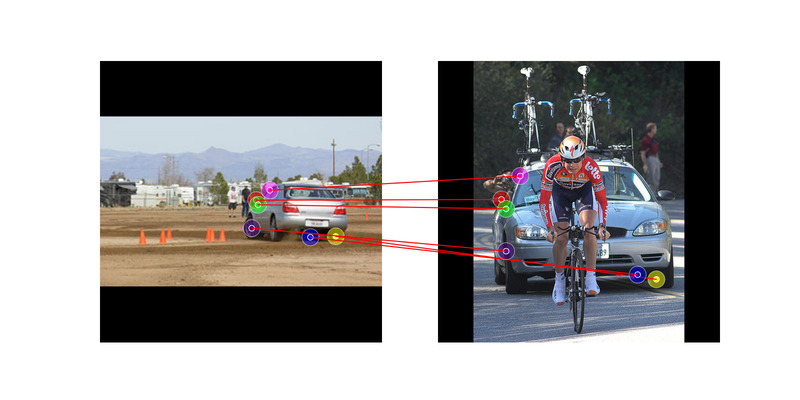}\\
        \rotatebox{90}{\parbox[t]{3cm}{\hspace*{\fill}\Large{Ours}\hspace*{\fill}}}\hspace*{1pt}
         & \includegraphics[width=.4\linewidth, height=.2\linewidth,trim={2cm 1cm 2cm 1cm}, clip]{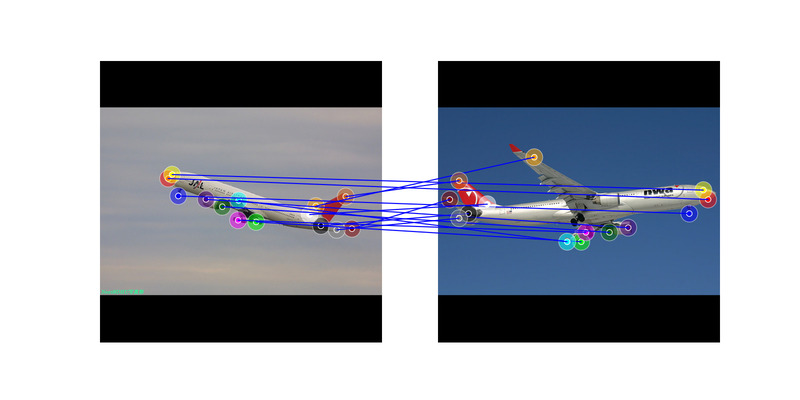}
         & \includegraphics[width=.4\linewidth, height=.2\linewidth,trim={2cm 1cm 2cm 1cm}, clip]{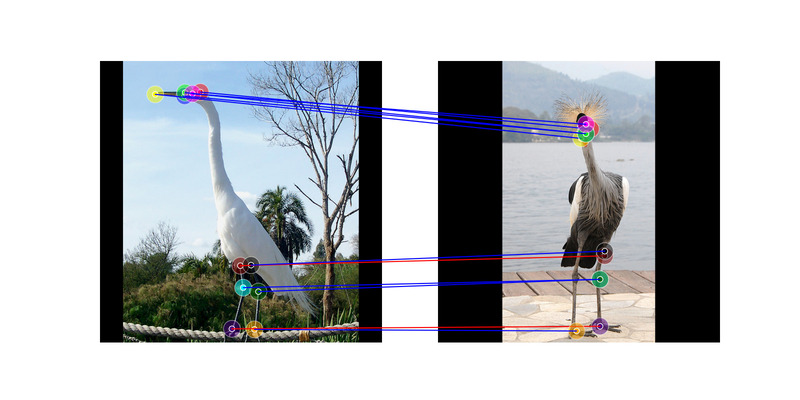}
         & \includegraphics[width=.4\linewidth, height=.2\linewidth,trim={2cm 1cm 2cm 1cm}, clip]{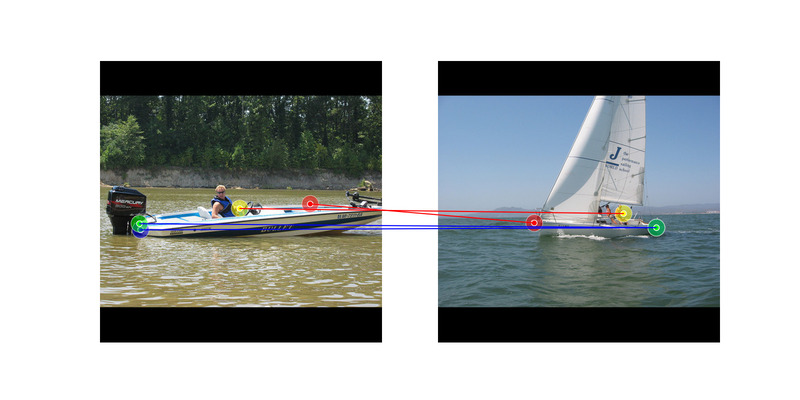}
         & \includegraphics[width=.4\linewidth, height=.2\linewidth,trim={2cm 1cm 2cm 1cm}, clip]{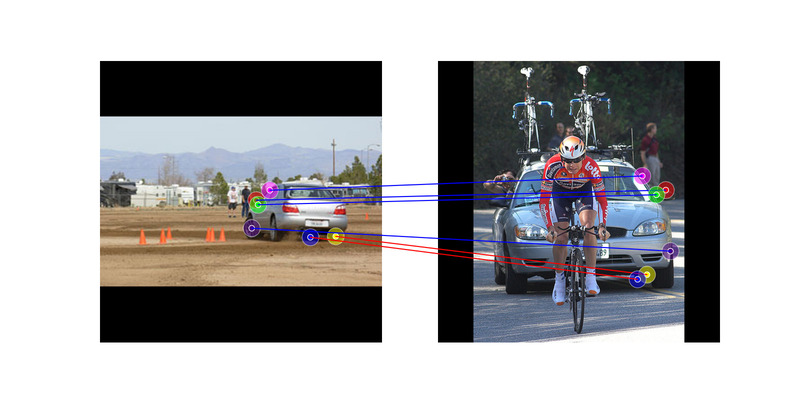}\\
        \rotatebox{90}{\parbox[t]{3cm}{\hspace*{\fill}\Large{Ours+SD}\hspace*{\fill}}}\hspace*{1pt}
         & \includegraphics[width=.4\linewidth, height=.2\linewidth,trim={2cm 1cm 2cm 1cm}, clip]{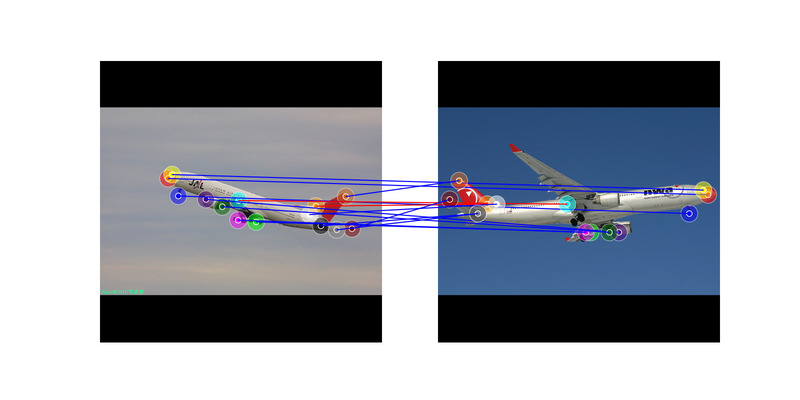}
         & \includegraphics[width=.4\linewidth, height=.2\linewidth,trim={2cm 1cm 2cm 1cm}, clip]{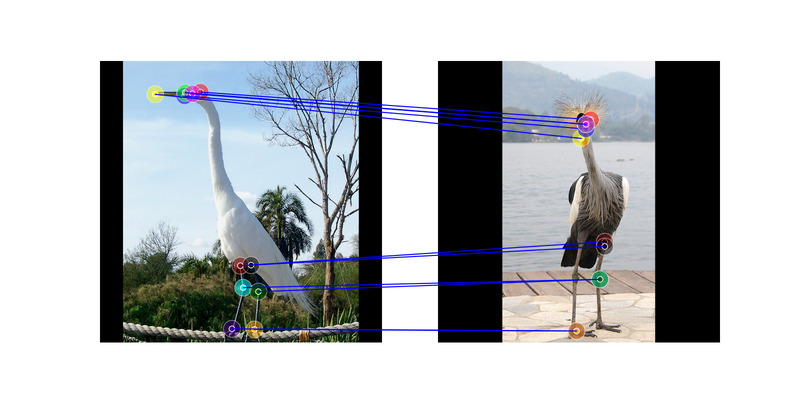}
         & \includegraphics[width=.4\linewidth, height=.2\linewidth,trim={2cm 1cm 2cm 1cm}, clip]{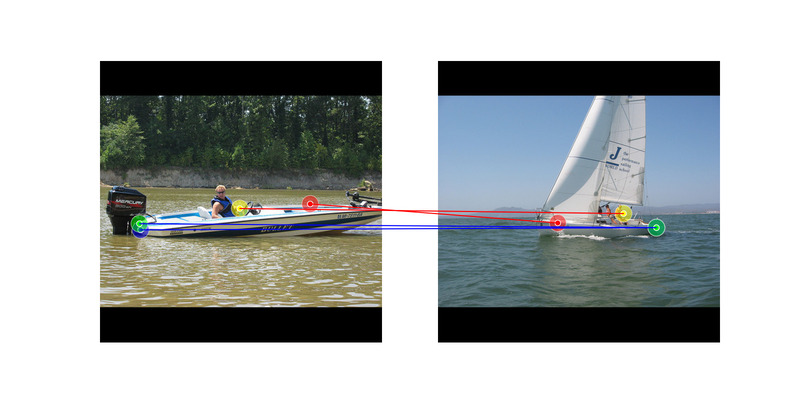}
         & \includegraphics[width=.4\linewidth, height=.2\linewidth,trim={2cm 1cm 2cm 1cm}, clip]{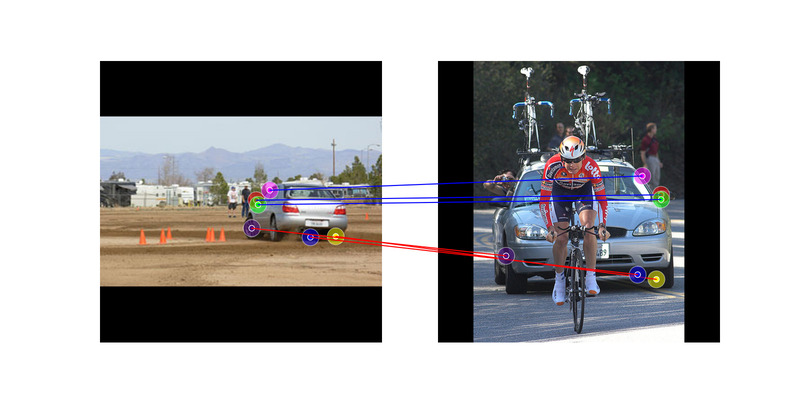}\\
        \rotatebox{90}{\parbox[t]{3cm}{\hspace*{\fill}\Large{DINOv2}\hspace*{\fill}}}\hspace*{1pt}
         & \includegraphics[width=.4\linewidth, height=.2\linewidth,trim={2cm 1cm 2cm 1cm}, clip]{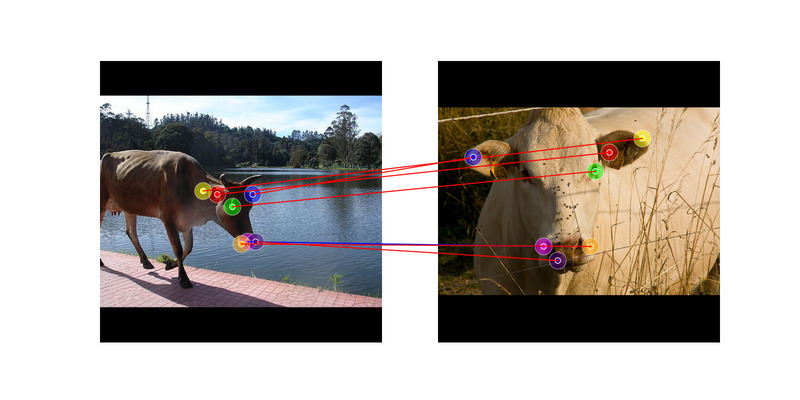}
         & \includegraphics[width=.4\linewidth, height=.2\linewidth,trim={2cm 1cm 2cm 1cm}, clip]{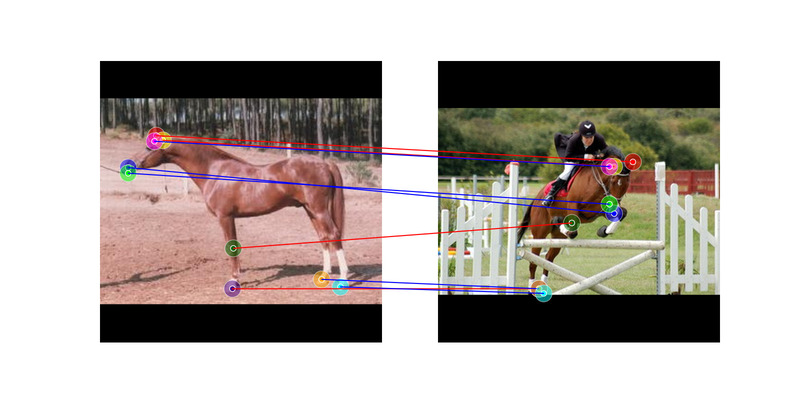}
         & \includegraphics[width=.4\linewidth, height=.2\linewidth,trim={2cm 1cm 2cm 1cm}, clip]{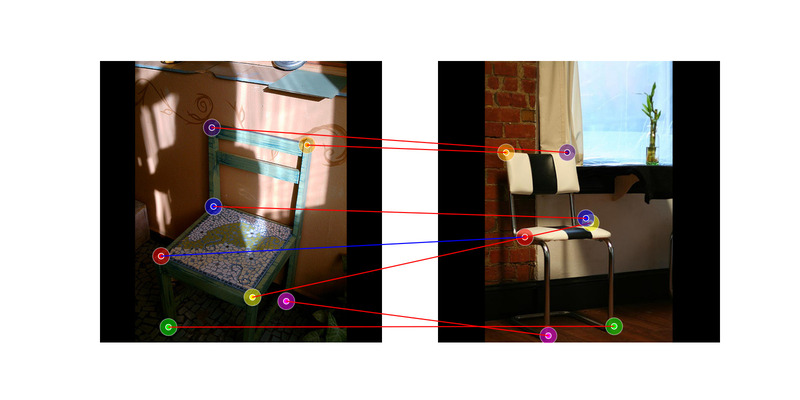}
         & \includegraphics[width=.4\linewidth, height=.2\linewidth,trim={2cm 1cm 2cm 1cm}, clip]{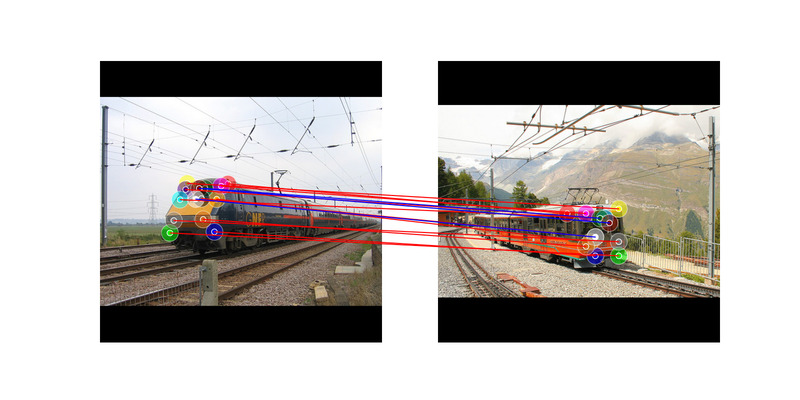}\\
        \rotatebox{90}{\parbox[t]{3cm}{\hspace*{\fill}\Large{SD}\hspace*{\fill}}}\hspace*{1pt}
         & \includegraphics[width=.4\linewidth, height=.2\linewidth,trim={2cm 1cm 2cm 1cm}, clip]{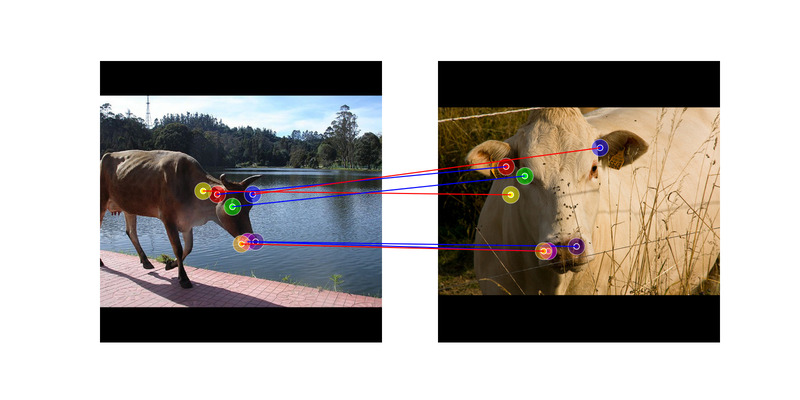}
         & \includegraphics[width=.4\linewidth, height=.2\linewidth,trim={2cm 1cm 2cm 1cm}, clip]{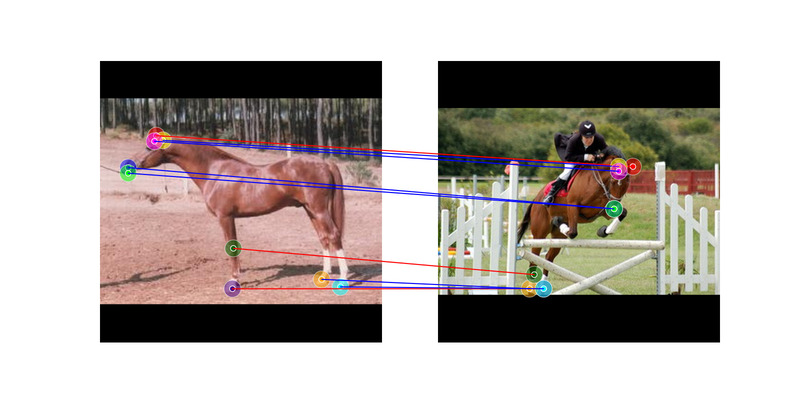}
         & \includegraphics[width=.4\linewidth, height=.2\linewidth,trim={2cm 1cm 2cm 1cm}, clip]{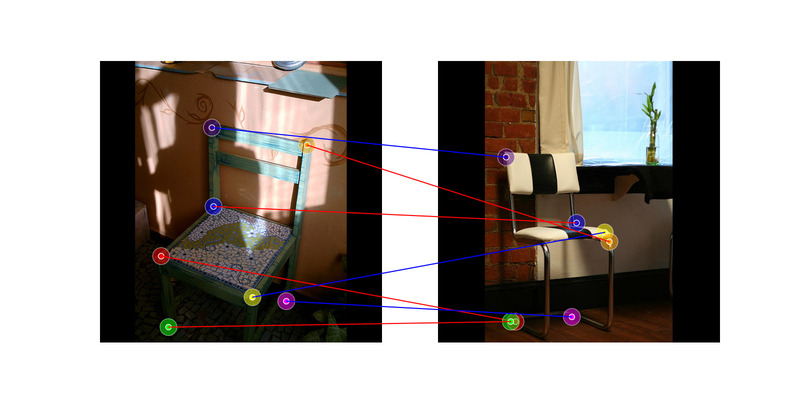}
         & \includegraphics[width=.4\linewidth, height=.2\linewidth,trim={2cm 1cm 2cm 1cm}, clip]{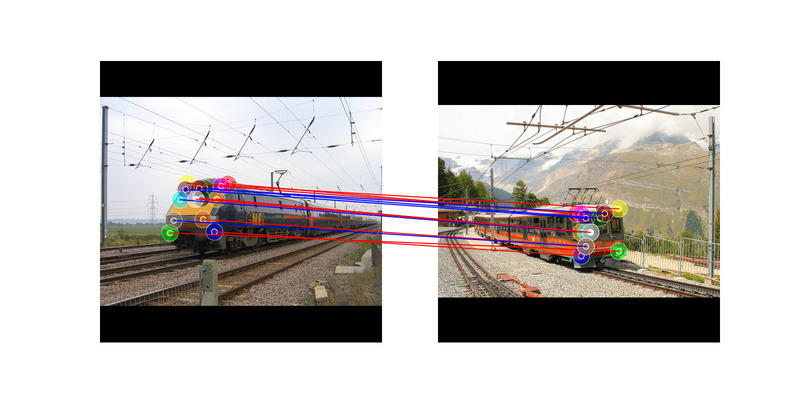}\\
        \rotatebox{90}{\parbox[t]{3cm}{\hspace*{\fill}\Large{DINOv2+SD}\hspace*{\fill}}}\hspace*{1pt}
         & \includegraphics[width=.4\linewidth, height=.2\linewidth,trim={2cm 1cm 2cm 1cm}, clip]{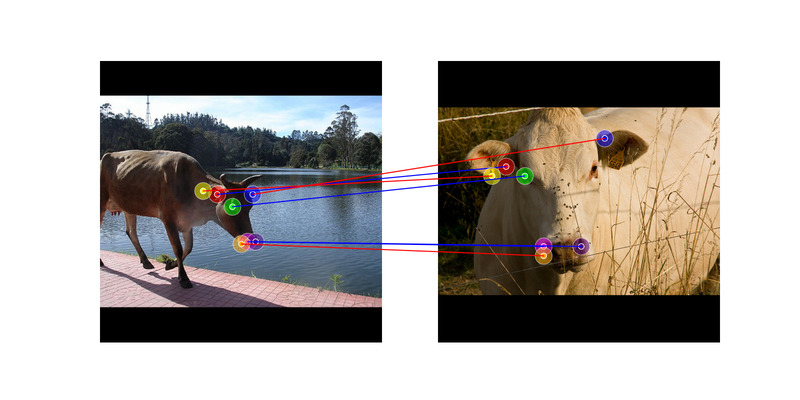}
         & \includegraphics[width=.4\linewidth, height=.2\linewidth,trim={2cm 1cm 2cm 1cm}, clip]{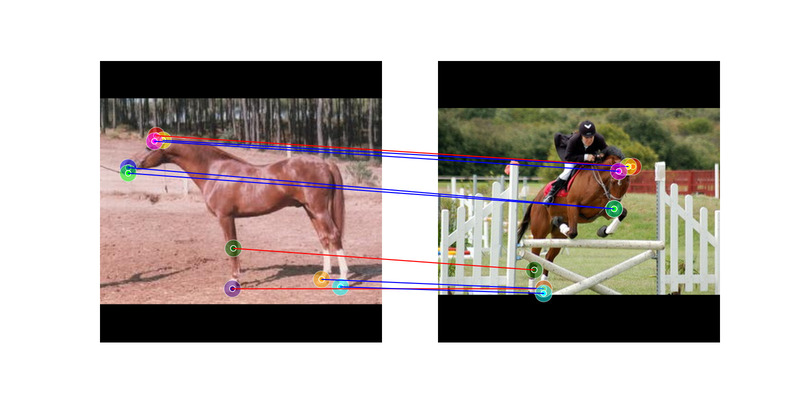}
         & \includegraphics[width=.4\linewidth, height=.2\linewidth,trim={2cm 1cm 2cm 1cm}, clip]{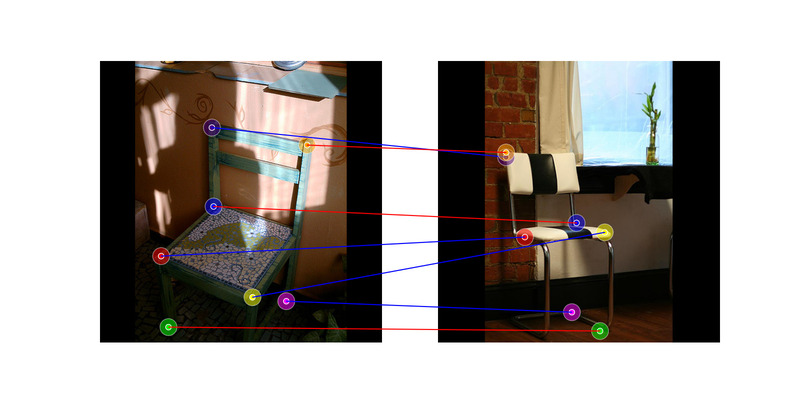}
         & \includegraphics[width=.4\linewidth, height=.2\linewidth,trim={2cm 1cm 2cm 1cm}, clip]{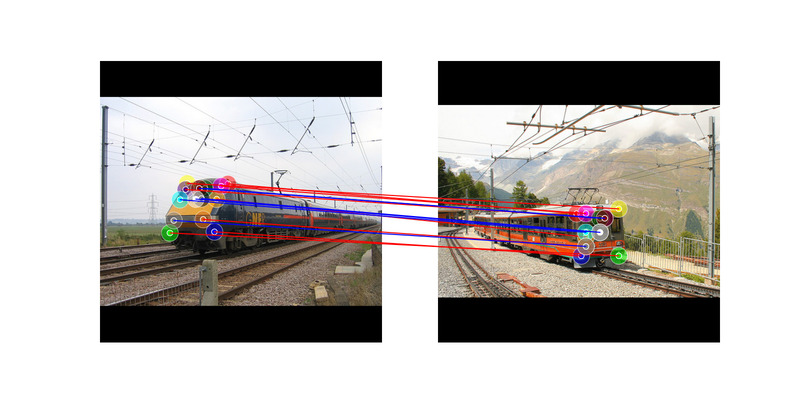}\\
        \rotatebox{90}{\parbox[t]{3cm}{\hspace*{\fill}\Large{Ours}\hspace*{\fill}}}\hspace*{1pt}
         & \includegraphics[width=.4\linewidth, height=.2\linewidth,trim={2cm 1cm 2cm 1cm}, clip]{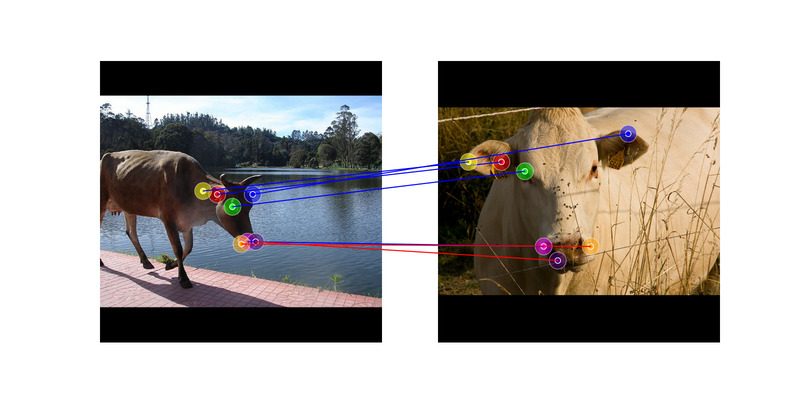}
         & \includegraphics[width=.4\linewidth, height=.2\linewidth,trim={2cm 1cm 2cm 1cm}, clip]{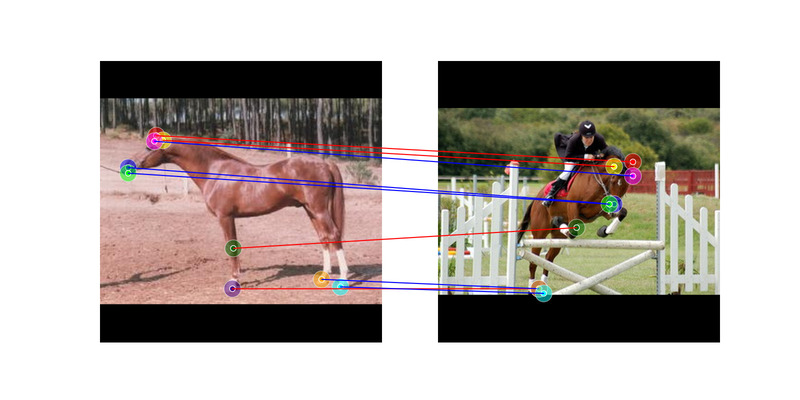}
         & \includegraphics[width=.4\linewidth, height=.2\linewidth,trim={2cm 1cm 2cm 1cm}, clip]{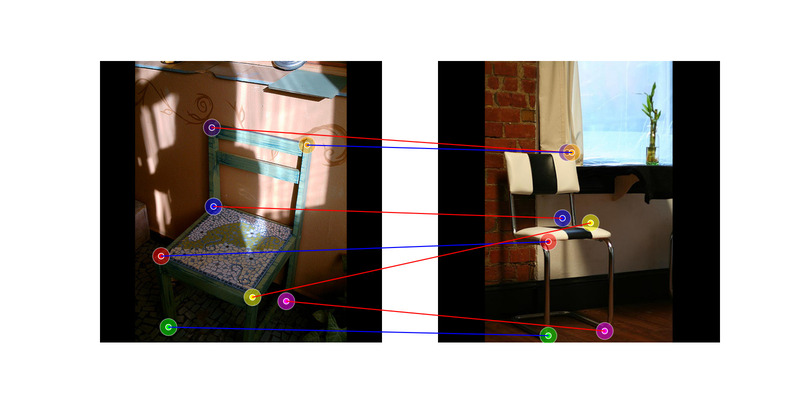}
         & \includegraphics[width=.4\linewidth, height=.2\linewidth,trim={2cm 1cm 2cm 1cm}, clip]{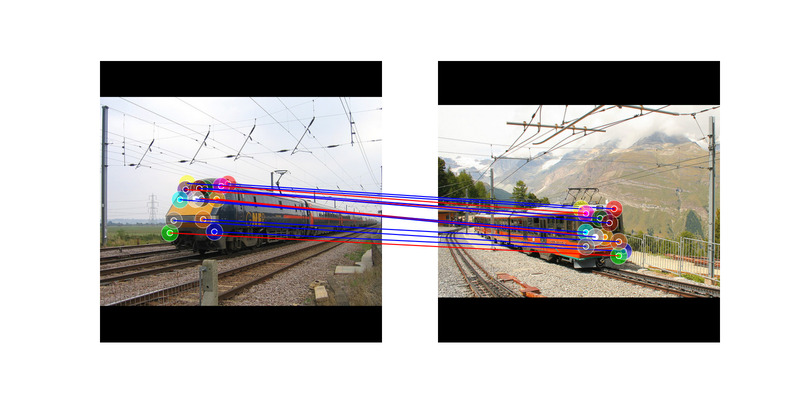}\\
        \rotatebox{90}{\parbox[t]{3cm}{\hspace*{\fill}\Large{Ours+SD}\hspace*{\fill}}}\hspace*{1pt}
         & \includegraphics[width=.4\linewidth, height=.2\linewidth,trim={2cm 1cm 2cm 1cm}, clip]{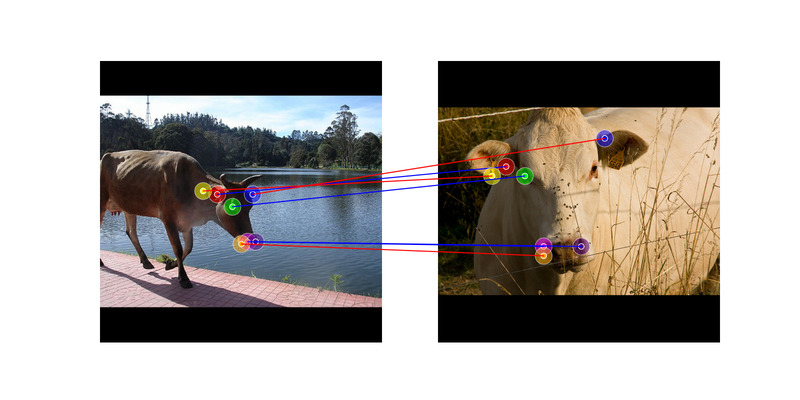}
         & \includegraphics[width=.4\linewidth, height=.2\linewidth,trim={2cm 1cm 2cm 1cm}, clip]{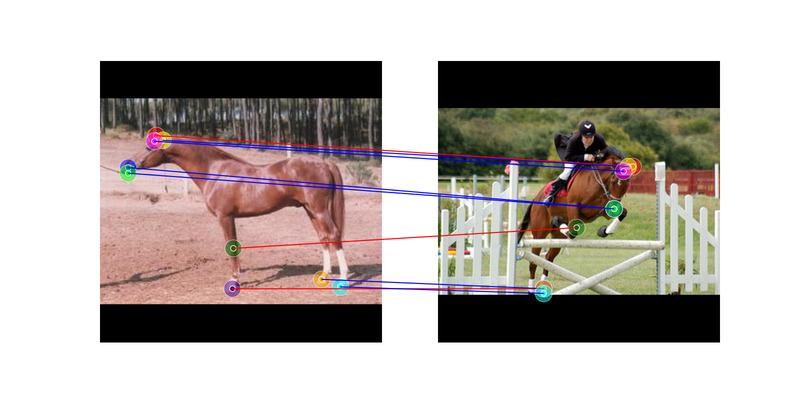}
         & \includegraphics[width=.4\linewidth, height=.2\linewidth,trim={2cm 1cm 2cm 1cm}, clip]{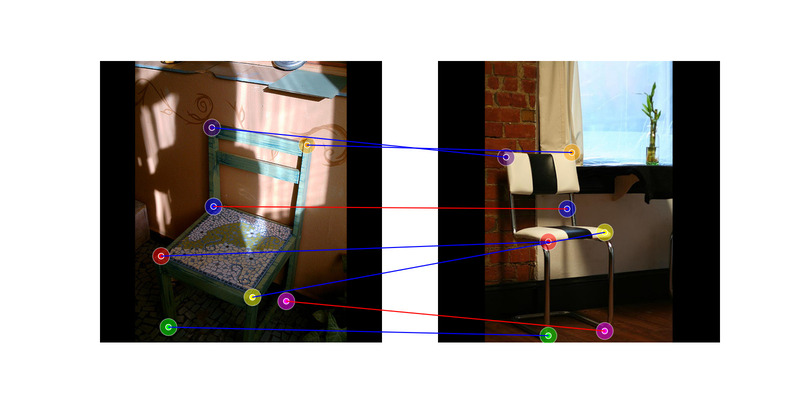}
         & \includegraphics[width=.4\linewidth, height=.2\linewidth,trim={2cm 1cm 2cm 1cm}, clip]{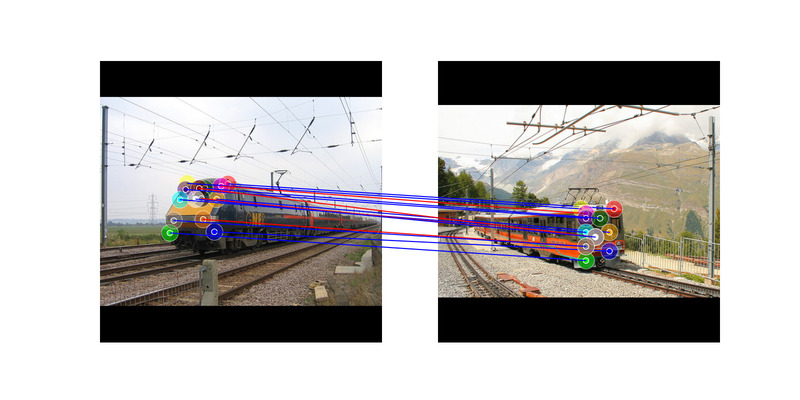}\\
    \end{tabular}
    }
    \vspace{-7pt}
    \caption{Keypoint matching on SPair-71k~\cite{min2019spair}. In each pair, the left image acts as the source, from which keypoint features are sampled, then the nearest neighbor of each feature is computed on the target to its right. Blue lines indicate correct matches, \ie within a $\tau = 0.1$ threshold of the ground truth, while red lines indicate incorrect matches.}
    \label{fig:kpts_match}
\end{figure*}

In \cref{fig:kpts_match} we present qualitative results illustrating keypoint matching on some particularly hard SPair-71k evaluation pairs that exhibit large camera viewpoint differences. 
For each keypoint in a source image, we show where its matched nearest neighbor lies in the target image.
These results show that our spherical maps make fewer mistakes on repeated parts, and are more likely to predict points on the correct side of objects in instances where there is visual ambiguity. It is particularly visible on the car example, where all models but ours map the left side of the source car to the right side of the target car, as they both appear on the same side of the image.

Our model still makes mistakes, though these are also present in other models.  
In particular, it struggles in the presence of large object scale variation (cow), confuses quadruped legs (horse), and deals poorly with large intra-class shape variations (chair)
A limitation of our model is the confusion it makes between legs of quadrupeds. 
However, these these mistakes are also present in other models.

\section{Supplementary video}
\label{sec:video}
\begin{figure*}[t]
    \resizebox{\textwidth}{!}{
    \begin{tabular}{ccccccccccccccccc}
        \rotatebox{90}{\parbox[t]{3cm}{\hspace*{\fill}\Large{Image}\hspace*{\fill}}}\hspace*{5pt}
         & \includegraphics[width=.2\linewidth, height=.2\linewidth]{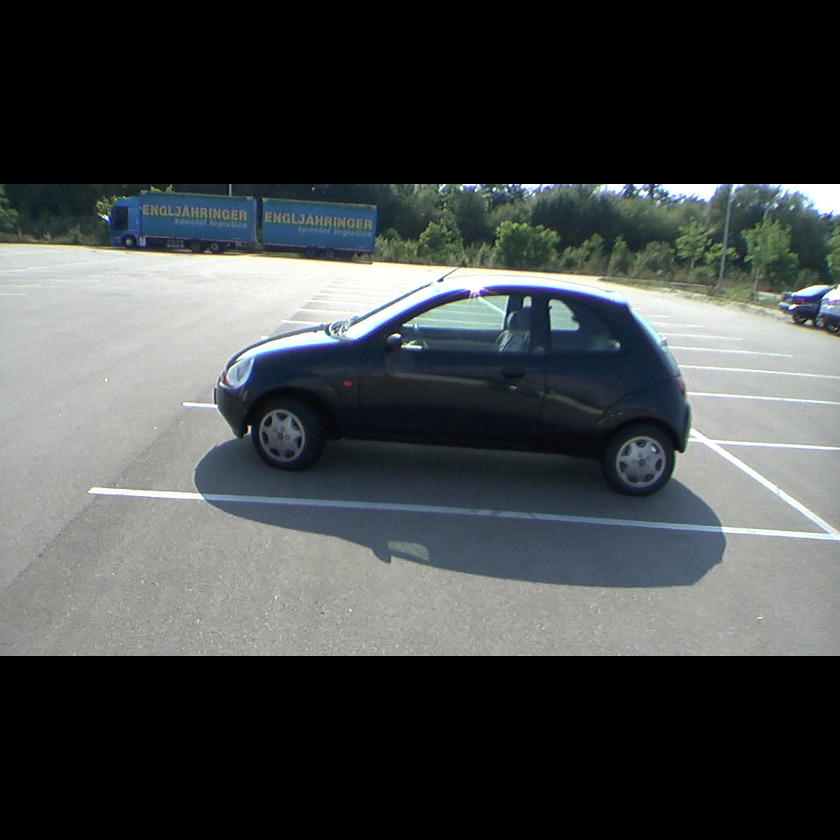}
         & \includegraphics[width=.2\linewidth, height=.2\linewidth]{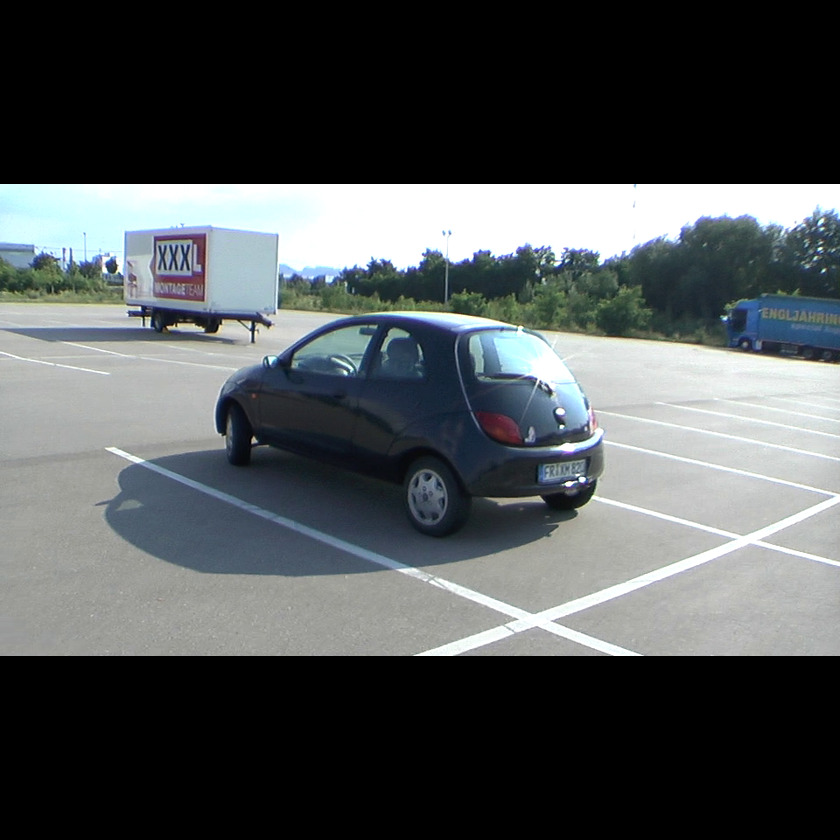}
         & \includegraphics[width=.2\linewidth, height=.2\linewidth]{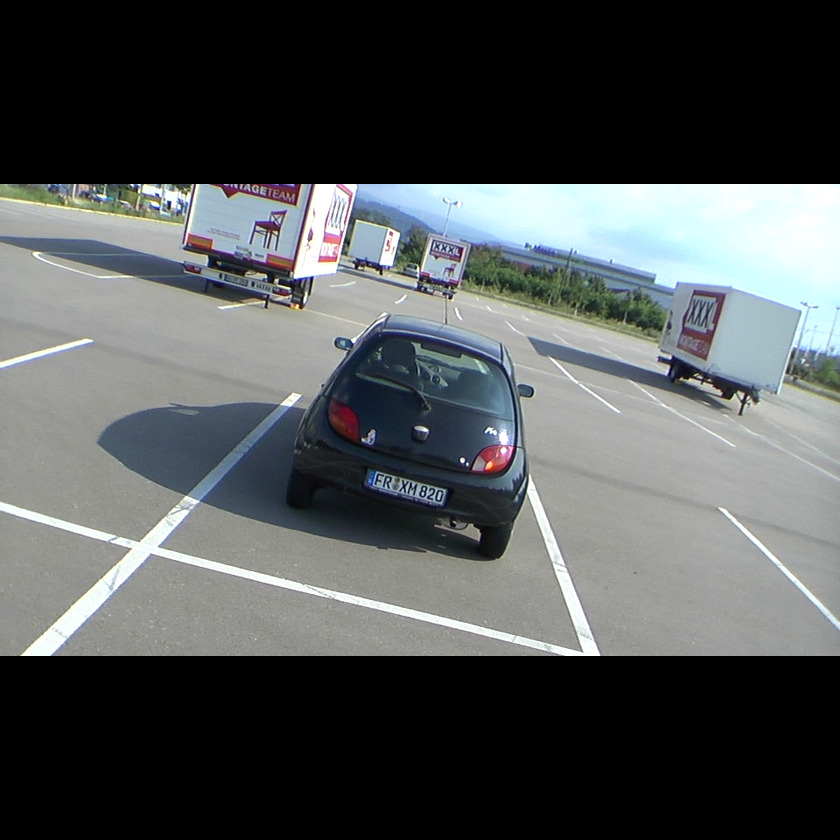}
         & \includegraphics[width=.2\linewidth, height=.2\linewidth]{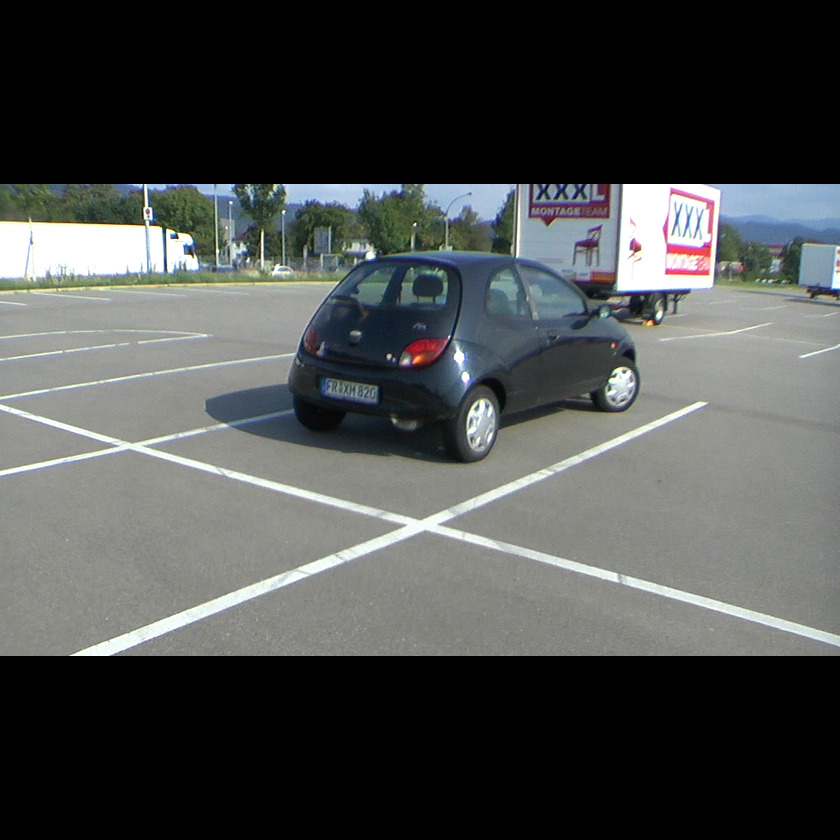}
         & \includegraphics[width=.2\linewidth, height=.2\linewidth]{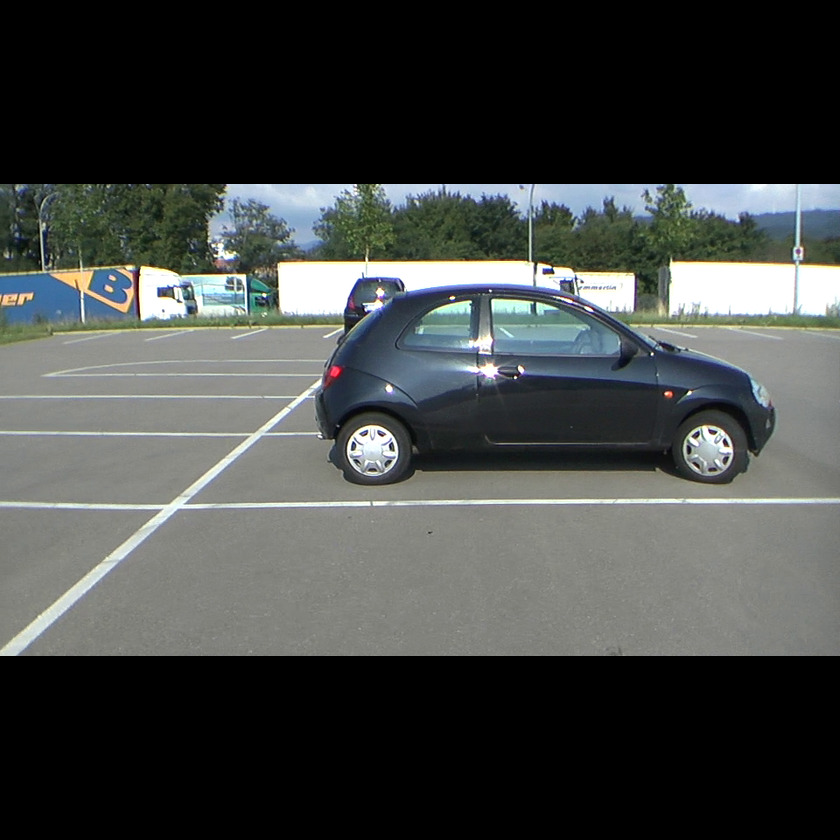}
         & \includegraphics[width=.2\linewidth, height=.2\linewidth]{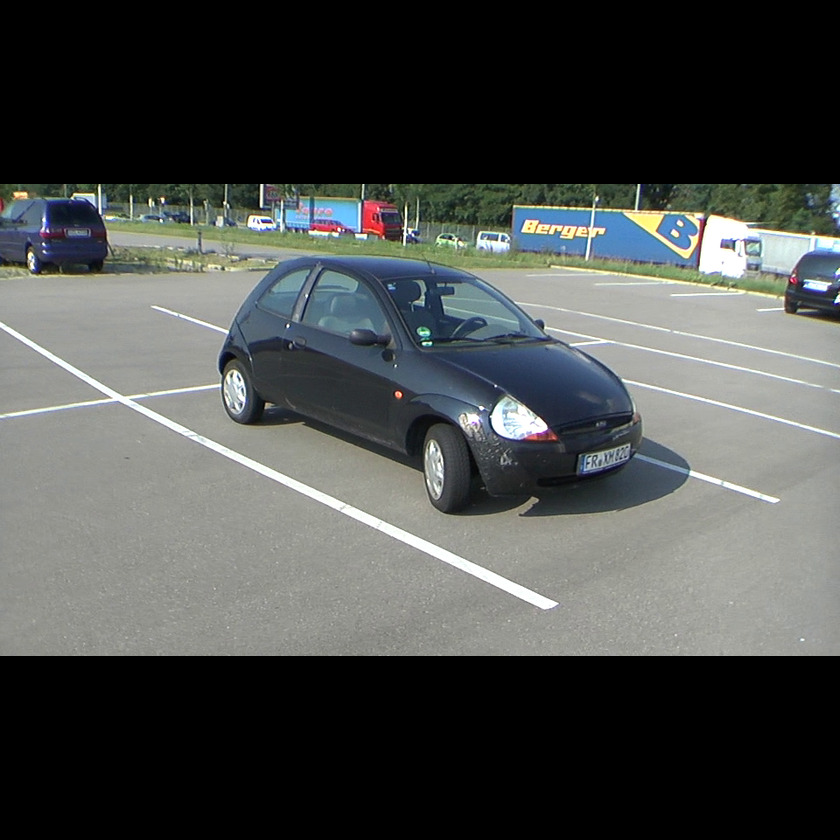}
         & \includegraphics[width=.2\linewidth, height=.2\linewidth]{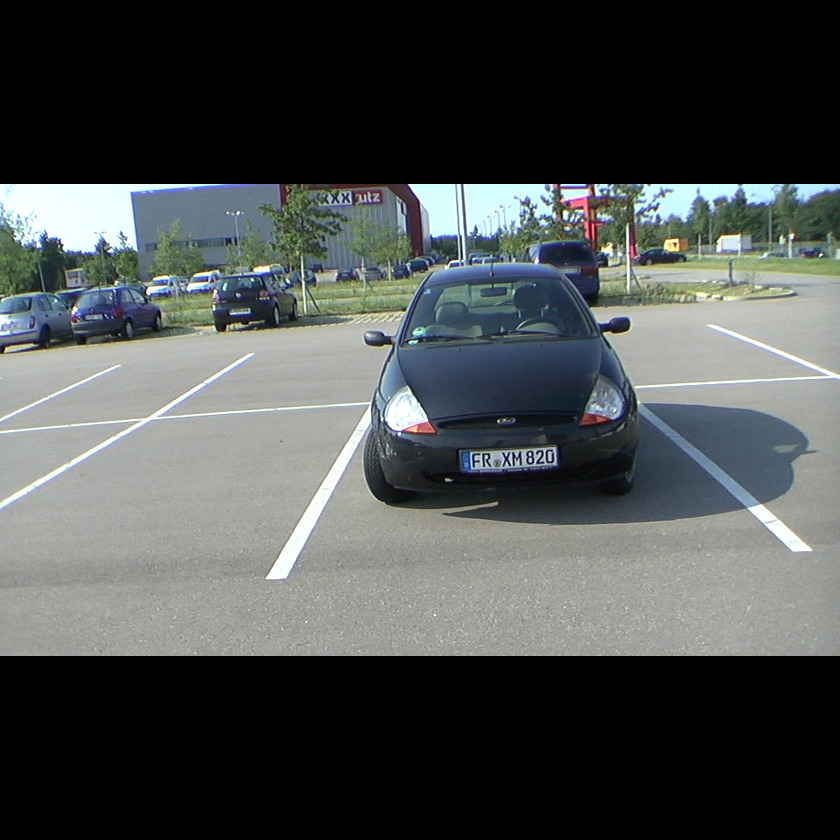}
         & \includegraphics[width=.2\linewidth, height=.2\linewidth]{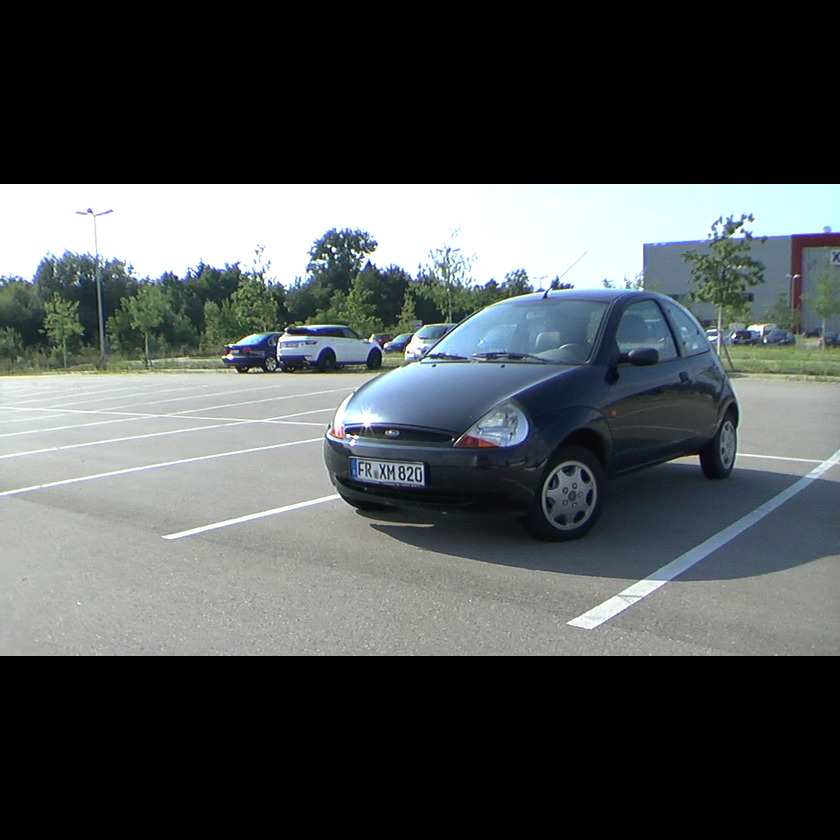}\\
        \rotatebox{90}{\parbox[t]{3cm}{\hspace*{\fill}\Large{DINOv2}\hspace*{\fill}}}\hspace*{5pt}
         & \includegraphics[width=.2\linewidth, height=.2\linewidth]{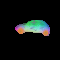}
         & \includegraphics[width=.2\linewidth, height=.2\linewidth]{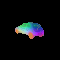}
         & \includegraphics[width=.2\linewidth, height=.2\linewidth]{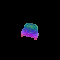}
         & \includegraphics[width=.2\linewidth, height=.2\linewidth]{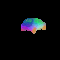}
         & \includegraphics[width=.2\linewidth, height=.2\linewidth]{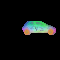}
         & \includegraphics[width=.2\linewidth, height=.2\linewidth]{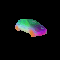}
         & \includegraphics[width=.2\linewidth, height=.2\linewidth]{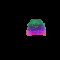}
         & \includegraphics[width=.2\linewidth, height=.2\linewidth]{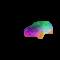}\\
        \rotatebox{90}{\parbox[t]{3cm}{\hspace*{\fill}\Large{SD}\hspace*{\fill}}}\hspace*{5pt}
         & \includegraphics[width=.2\linewidth, height=.2\linewidth]{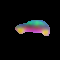}
         & \includegraphics[width=.2\linewidth, height=.2\linewidth]{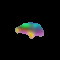}
         & \includegraphics[width=.2\linewidth, height=.2\linewidth]{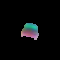}
         & \includegraphics[width=.2\linewidth, height=.2\linewidth]{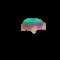}
         & \includegraphics[width=.2\linewidth, height=.2\linewidth]{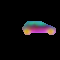}
         & \includegraphics[width=.2\linewidth, height=.2\linewidth]{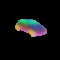}
         & \includegraphics[width=.2\linewidth, height=.2\linewidth]{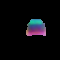}
         & \includegraphics[width=.2\linewidth, height=.2\linewidth]{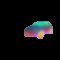}\\
        \rotatebox{90}{\parbox[t]{3cm}{\hspace*{\fill}\Large{DINOv2+SD}\hspace*{\fill}}}\hspace*{5pt}
         & \includegraphics[width=.2\linewidth, height=.2\linewidth]{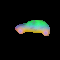}
         & \includegraphics[width=.2\linewidth, height=.2\linewidth]{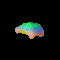}
         & \includegraphics[width=.2\linewidth, height=.2\linewidth]{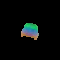}
         & \includegraphics[width=.2\linewidth, height=.2\linewidth]{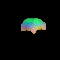}
         & \includegraphics[width=.2\linewidth, height=.2\linewidth]{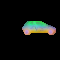}
         & \includegraphics[width=.2\linewidth, height=.2\linewidth]{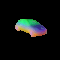}
         & \includegraphics[width=.2\linewidth, height=.2\linewidth]{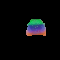}
         & \includegraphics[width=.2\linewidth, height=.2\linewidth]{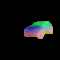}\\
        \rotatebox{90}{\parbox[t]{3cm}{\hspace*{\fill}\Large{Sphere}\hspace*{\fill}}}\hspace*{5pt}
         & \includegraphics[width=.2\linewidth, height=.2\linewidth]{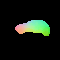}
         & \includegraphics[width=.2\linewidth, height=.2\linewidth]{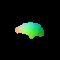}
         & \includegraphics[width=.2\linewidth, height=.2\linewidth]{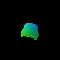}
         & \includegraphics[width=.2\linewidth, height=.2\linewidth]{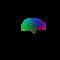}
         & \includegraphics[width=.2\linewidth, height=.2\linewidth]{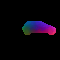}
         & \includegraphics[width=.2\linewidth, height=.2\linewidth]{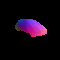}
         & \includegraphics[width=.2\linewidth, height=.2\linewidth]{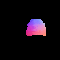}
         & \includegraphics[width=.2\linewidth, height=.2\linewidth]{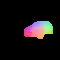}\\
        \rotatebox{90}{\parbox[t]{3cm}{\hspace*{\fill}\Large{Image}\hspace*{\fill}}}\hspace*{5pt}
         & \includegraphics[width=.2\linewidth, height=.2\linewidth]{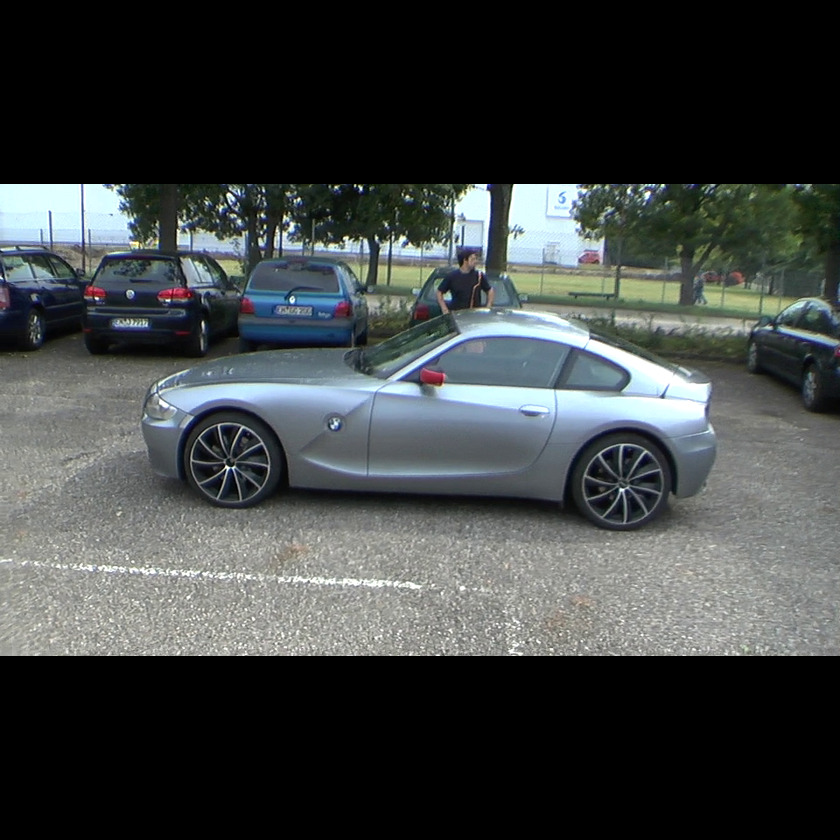}
         & \includegraphics[width=.2\linewidth, height=.2\linewidth]{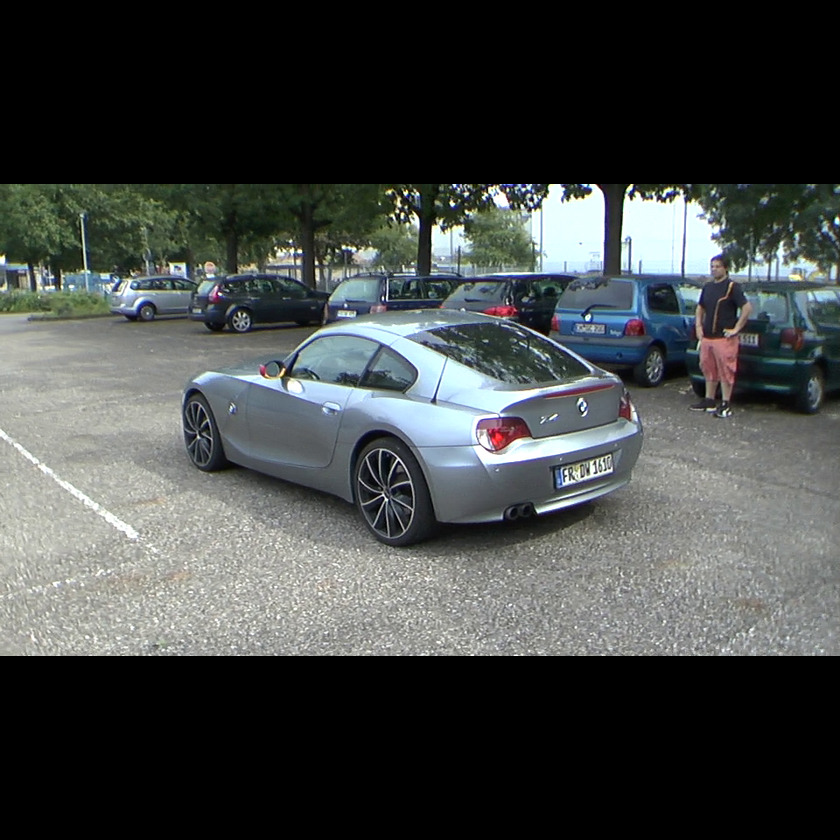}
         & \includegraphics[width=.2\linewidth, height=.2\linewidth]{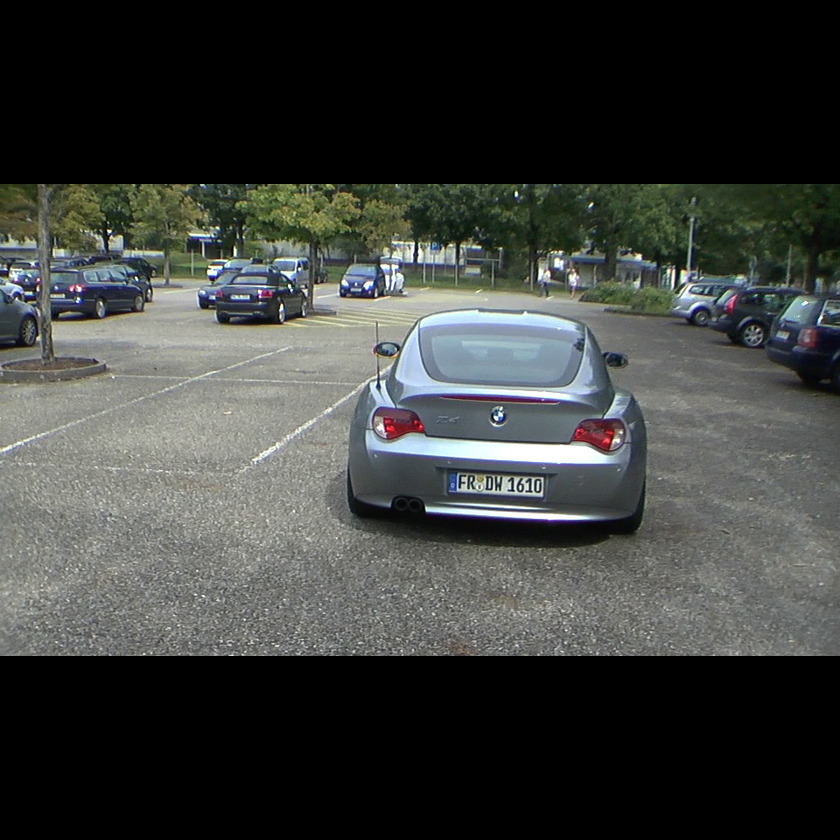}
         & \includegraphics[width=.2\linewidth, height=.2\linewidth]{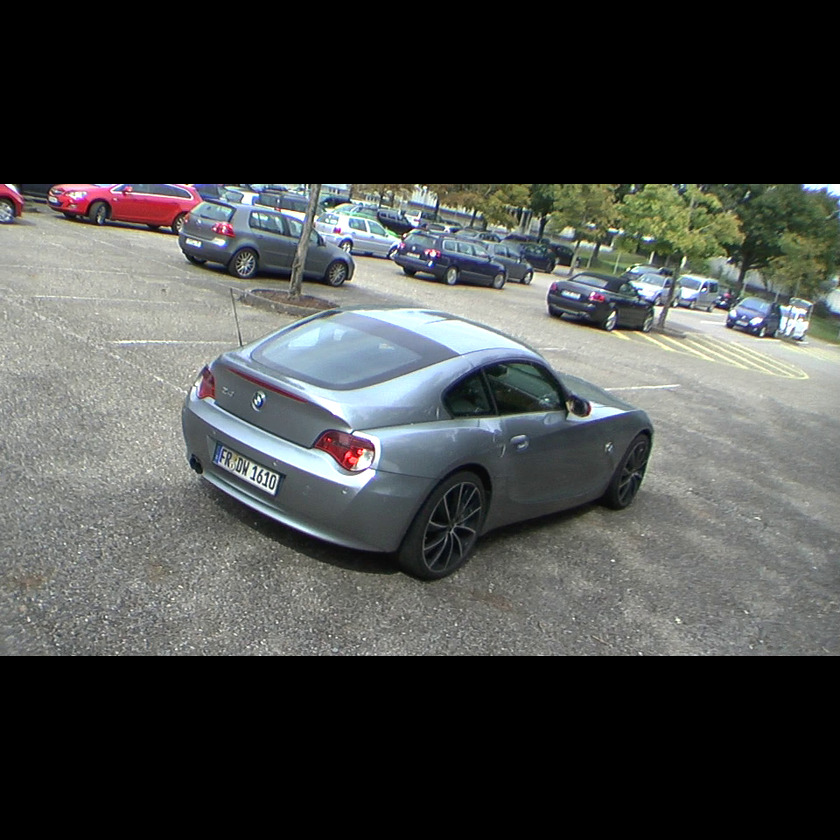}
         & \includegraphics[width=.2\linewidth, height=.2\linewidth]{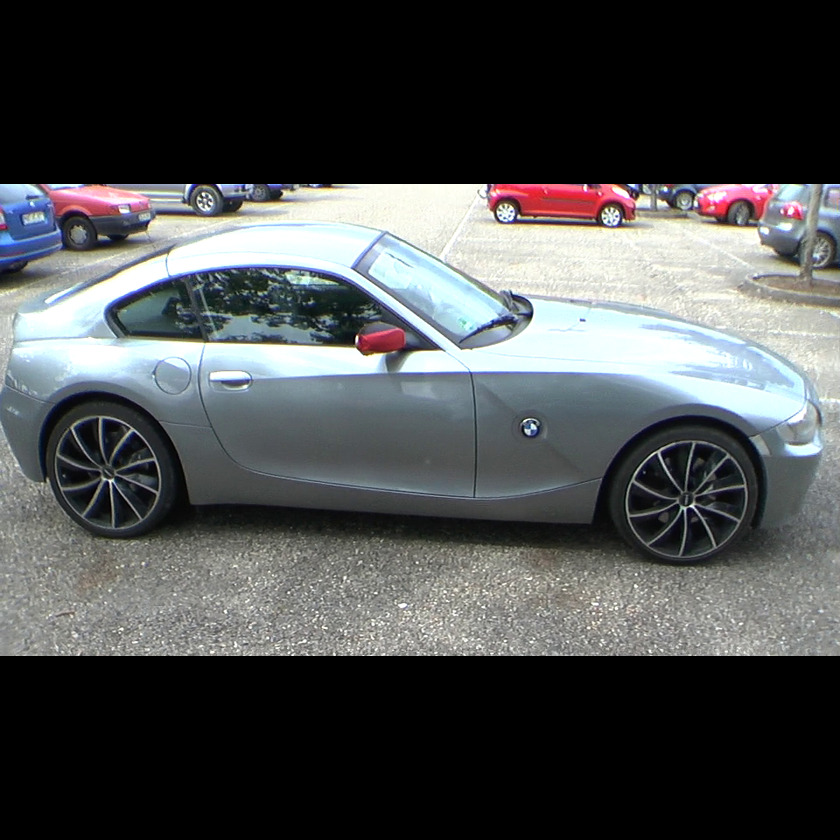}
         & \includegraphics[width=.2\linewidth, height=.2\linewidth]{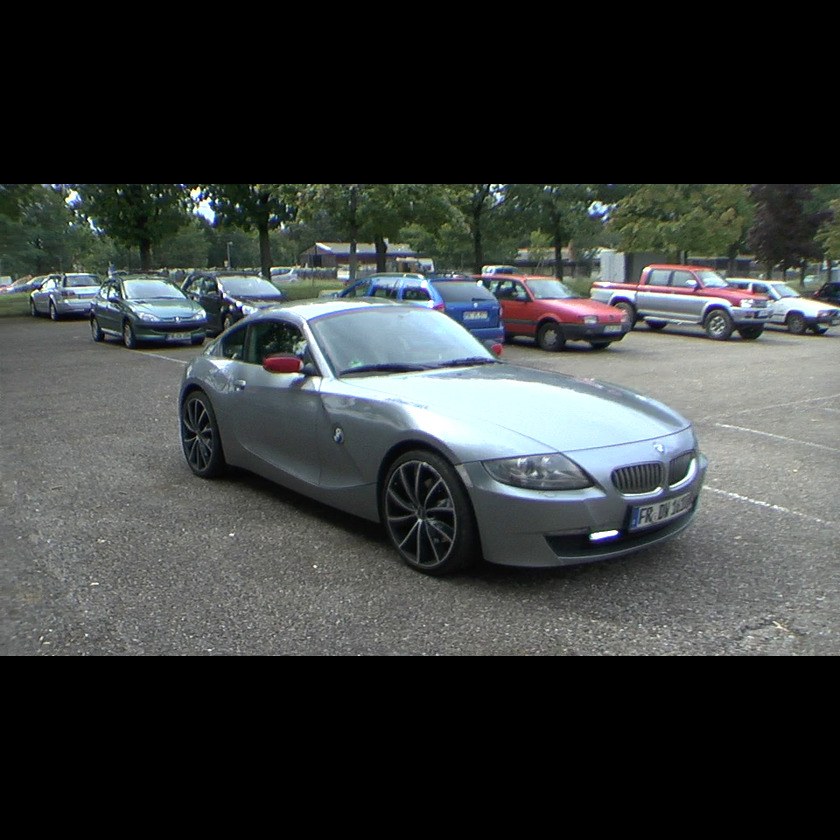}
         & \includegraphics[width=.2\linewidth, height=.2\linewidth]{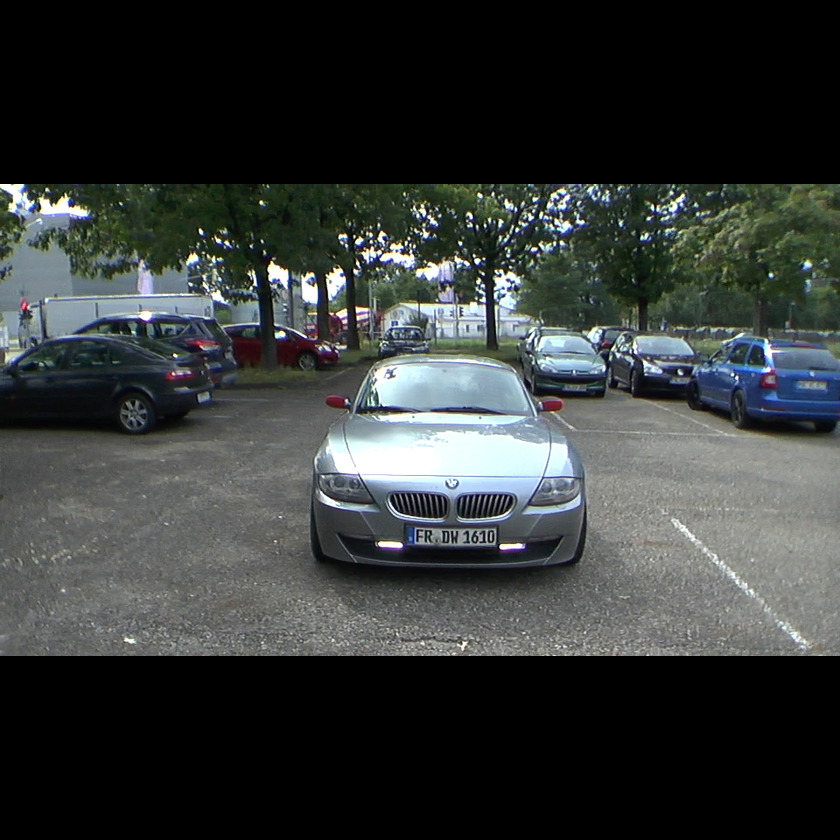}
         & \includegraphics[width=.2\linewidth, height=.2\linewidth]{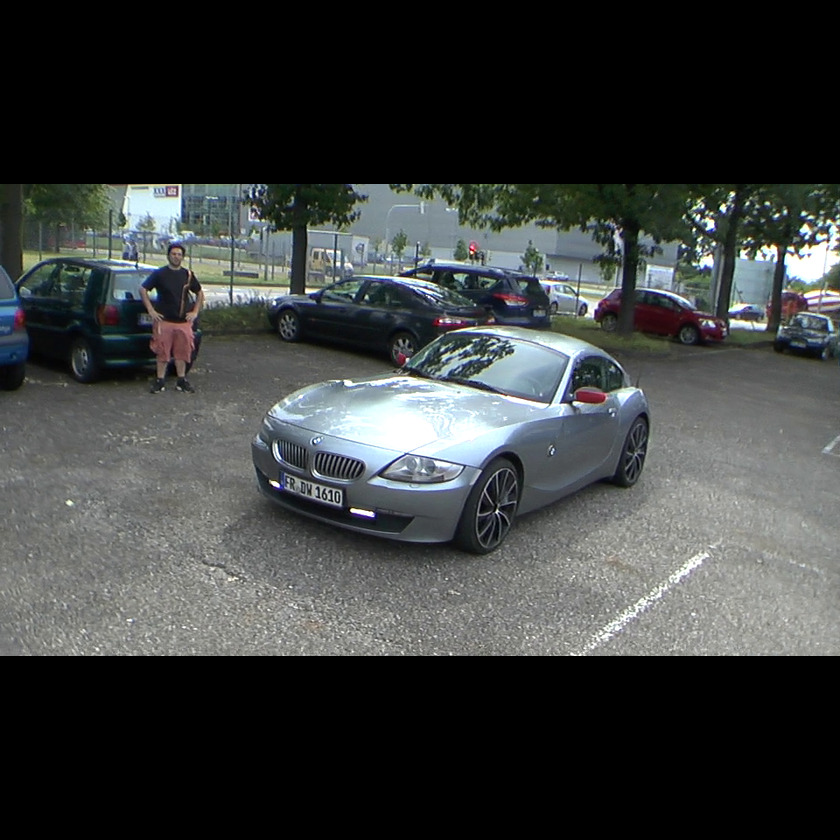}\\
        \rotatebox{90}{\parbox[t]{3cm}{\hspace*{\fill}\Large{DINOv2}\hspace*{\fill}}}\hspace*{5pt}
         & \includegraphics[width=.2\linewidth, height=.2\linewidth]{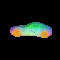}
         & \includegraphics[width=.2\linewidth, height=.2\linewidth]{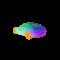}
         & \includegraphics[width=.2\linewidth, height=.2\linewidth]{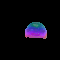}
         & \includegraphics[width=.2\linewidth, height=.2\linewidth]{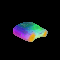}
         & \includegraphics[width=.2\linewidth, height=.2\linewidth]{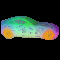}
         & \includegraphics[width=.2\linewidth, height=.2\linewidth]{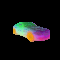}
         & \includegraphics[width=.2\linewidth, height=.2\linewidth]{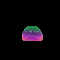}
         & \includegraphics[width=.2\linewidth, height=.2\linewidth]{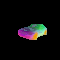}\\
        \rotatebox{90}{\parbox[t]{3cm}{\hspace*{\fill}\Large{SD}\hspace*{\fill}}}\hspace*{5pt}
         & \includegraphics[width=.2\linewidth, height=.2\linewidth]{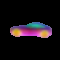}
         & \includegraphics[width=.2\linewidth, height=.2\linewidth]{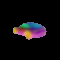}
         & \includegraphics[width=.2\linewidth, height=.2\linewidth]{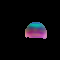}
         & \includegraphics[width=.2\linewidth, height=.2\linewidth]{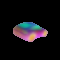}
         & \includegraphics[width=.2\linewidth, height=.2\linewidth]{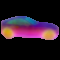}
         & \includegraphics[width=.2\linewidth, height=.2\linewidth]{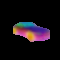}
         & \includegraphics[width=.2\linewidth, height=.2\linewidth]{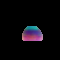}
         & \includegraphics[width=.2\linewidth, height=.2\linewidth]{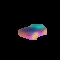}\\
        \rotatebox{90}{\parbox[t]{3cm}{\hspace*{\fill}\Large{DINOv2+SD}\hspace*{\fill}}}\hspace*{5pt}
         & \includegraphics[width=.2\linewidth, height=.2\linewidth]{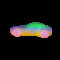}
         & \includegraphics[width=.2\linewidth, height=.2\linewidth]{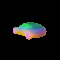}
         & \includegraphics[width=.2\linewidth, height=.2\linewidth]{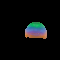}
         & \includegraphics[width=.2\linewidth, height=.2\linewidth]{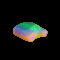}
         & \includegraphics[width=.2\linewidth, height=.2\linewidth]{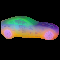}
         & \includegraphics[width=.2\linewidth, height=.2\linewidth]{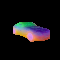}
         & \includegraphics[width=.2\linewidth, height=.2\linewidth]{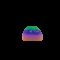}
         & \includegraphics[width=.2\linewidth, height=.2\linewidth]{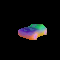}\\
        \rotatebox{90}{\parbox[t]{3cm}{\hspace*{\fill}\Large{Sphere}\hspace*{\fill}}}\hspace*{5pt}
         & \includegraphics[width=.2\linewidth, height=.2\linewidth]{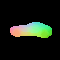}
         & \includegraphics[width=.2\linewidth, height=.2\linewidth]{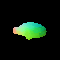}
         & \includegraphics[width=.2\linewidth, height=.2\linewidth]{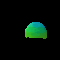}
         & \includegraphics[width=.2\linewidth, height=.2\linewidth]{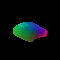}
         & \includegraphics[width=.2\linewidth, height=.2\linewidth]{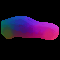}
         & \includegraphics[width=.2\linewidth, height=.2\linewidth]{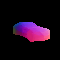}
         & \includegraphics[width=.2\linewidth, height=.2\linewidth]{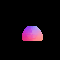}
         & \includegraphics[width=.2\linewidth, height=.2\linewidth]{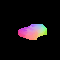}\\
    \end{tabular}
    }
    \vspace{-5pt}
    \caption{Here we illustrate the multi-view consistency of our approach at test time on two different car sequences from the Freiburg cars dataset~\cite{sedaghat2015unsupervised}. 
    For each sequence, we show input images from different view points, DINOv2, Stable Diffusion (SD), and DINOv2+SD PCA feature maps, and our predicted spherical maps. 
    While other models capture semantic parts, in contrast to us, they fail to correctly disambiguate the two different sides of each car resulting in the same features for the left and right sides. They also fail to produce distinct features for individual car wheels. Note, these large viewpoint changes are typically not assessed in the Spair-71k~\cite{min2019spair} benchmark. Please see the supplementary video for 360$^\circ$ videos.
    }
    \vspace{-10pt}
    \label{fig:Fcar}
\end{figure*}

Finally, our \href{groups.inf.ed.ac.uk/vico/research/sphericalmaps}{project website} contains supplementary videos where we compare to different methods using held-out image sequences. 
While the results demonstrate predictions on images from short video sequences, the models do not use any temporal information, and in our case of our approach we are training on the held-out SPair-71k training set.  
It is very apparent in the video that the baseline methods confuse the different sides of the cars and horses, in addition to generating the same features for the different wheels of the car. 
This is evident by the fact that these distinct parts have the same color which is obtained by performing PCA to reduce the features to three dimensions. 
In contrast, our spherical-based approach attempts to map each point on the surface of the object to unique features. 
Note, that we show images from the same car sequences as in \cref{fig:Fcar} where PCA is computed over images from those sequences, but in the video PCA is compute over five sequences.

\begin{figure*}[t]
    \setlength{\tabcolsep}{3pt}
    \resizebox{\textwidth}{!}{
    \begin{tabular}{ccccccccccccccc}
        \rotatebox{90}{\parbox[t]{3cm}{\hspace*{\fill}\Large{Image}\hspace*{\fill}}}\hspace*{5pt}
         & \includegraphics[width=.2\linewidth, height=.2\linewidth]{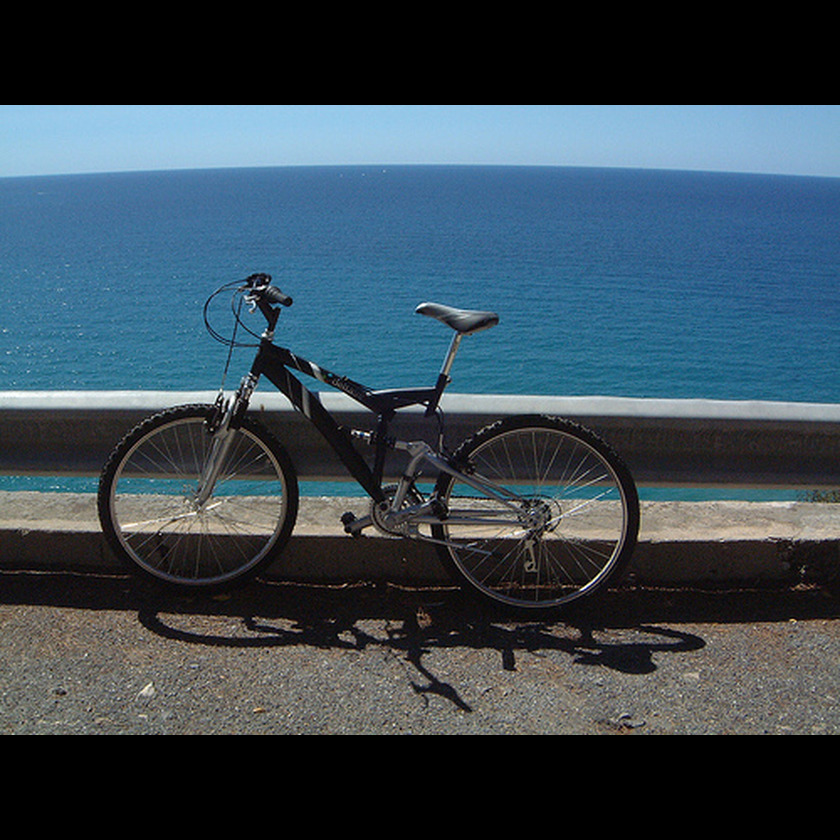}
         & \includegraphics[width=.2\linewidth, height=.2\linewidth]{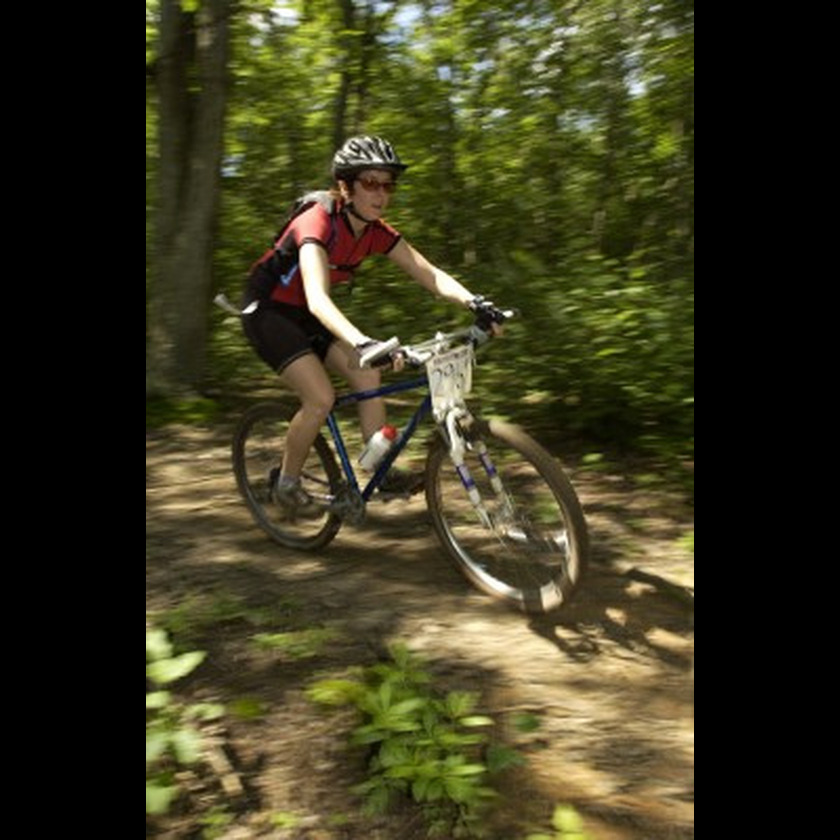}
         & \includegraphics[width=.2\linewidth, height=.2\linewidth]{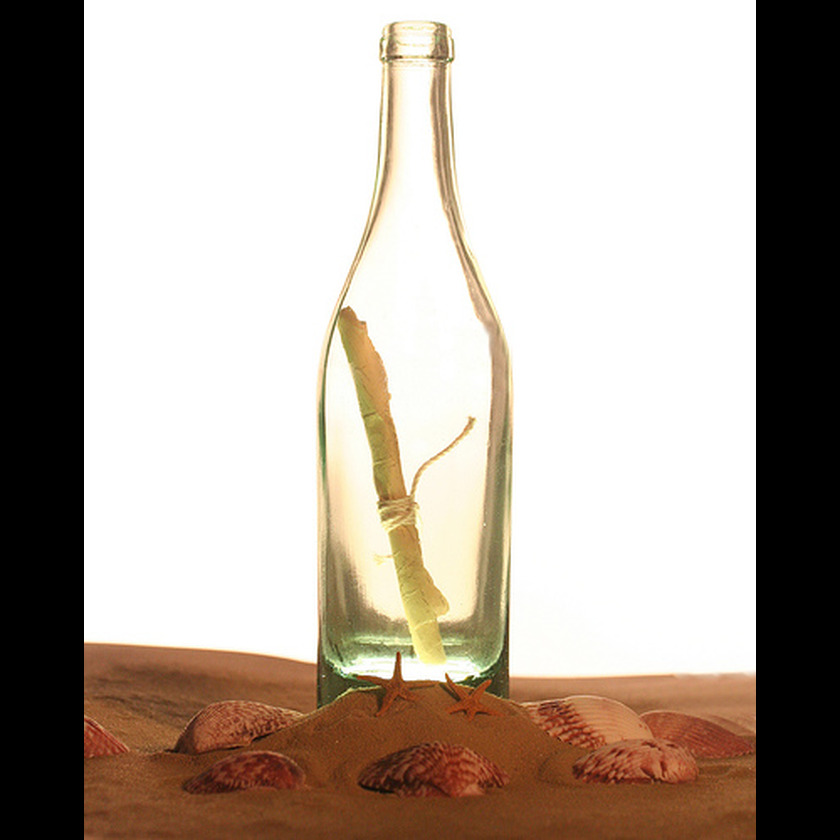}
         & \includegraphics[width=.2\linewidth, height=.2\linewidth]{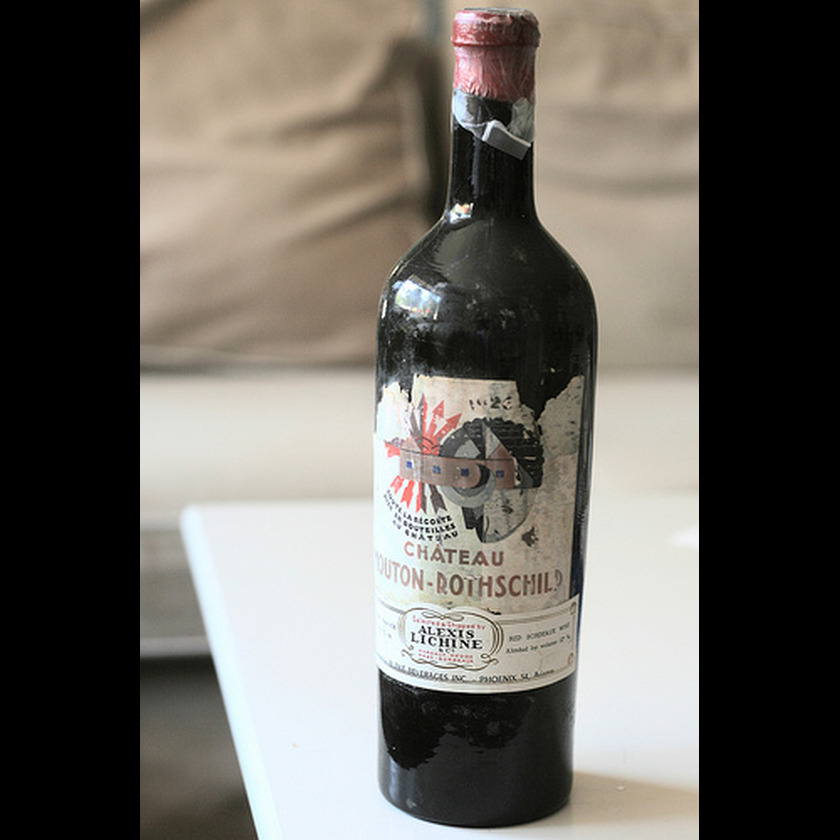}
         & \includegraphics[width=.2\linewidth, height=.2\linewidth]{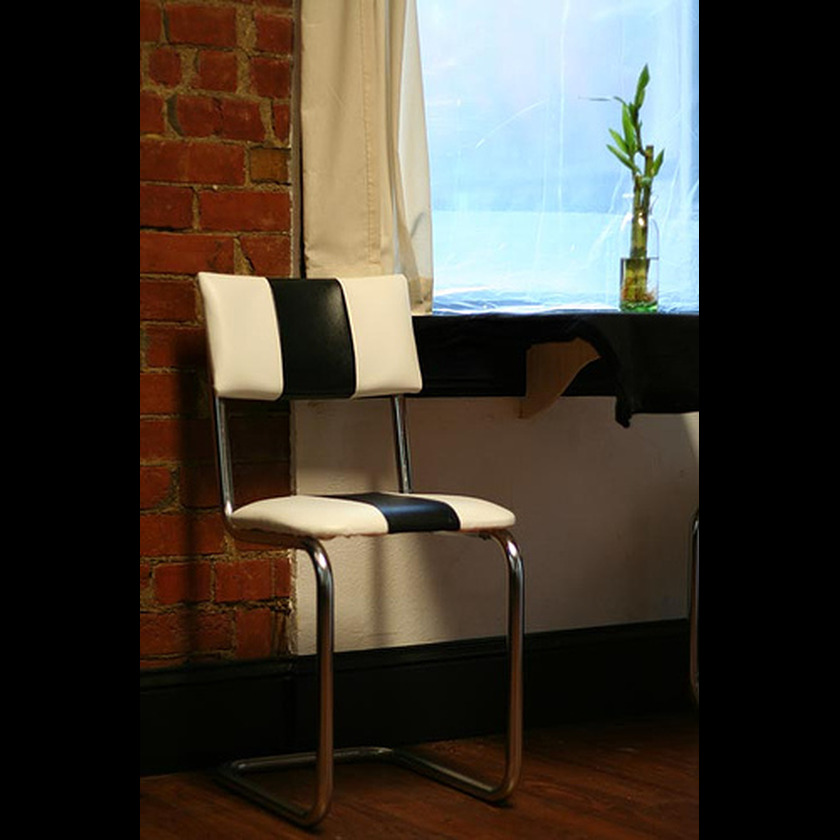}
         & \includegraphics[width=.2\linewidth, height=.2\linewidth]{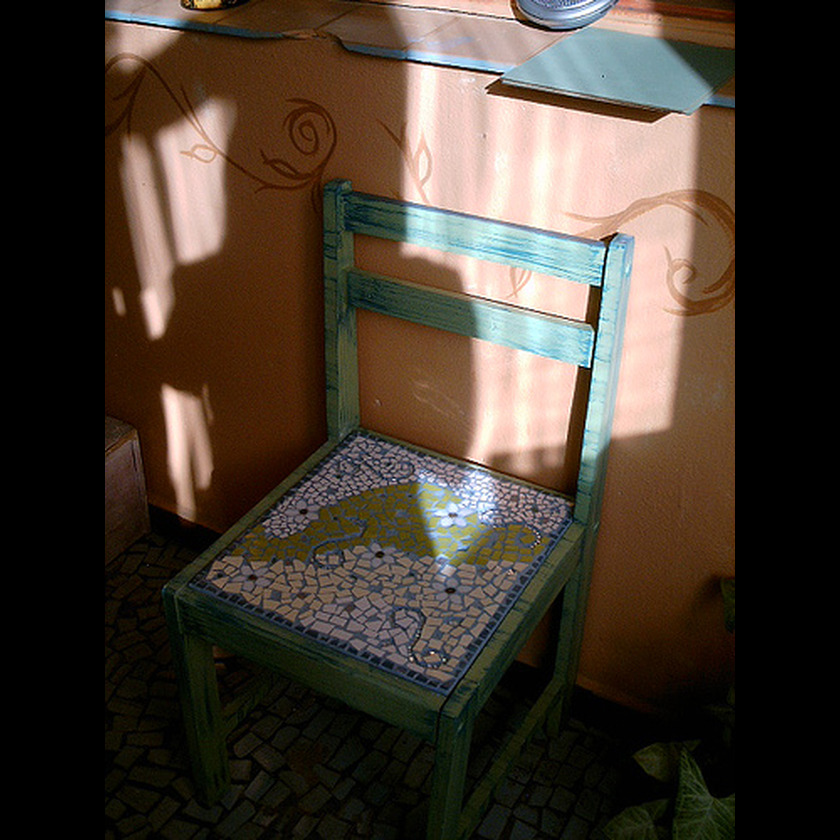}
         \\
        \rotatebox{90}{\parbox[t]{3cm}{\hspace*{\fill}\Large{DINOv2}\hspace*{\fill}}}\hspace*{5pt}
         & \includegraphics[width=.2\linewidth, height=.2\linewidth]{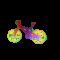}
         & \includegraphics[width=.2\linewidth, height=.2\linewidth]{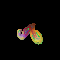}
         & \includegraphics[width=.2\linewidth, height=.2\linewidth]{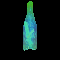}
         & \includegraphics[width=.2\linewidth, height=.2\linewidth]{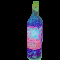}
         & \includegraphics[width=.2\linewidth, height=.2\linewidth]{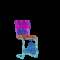}
         & \includegraphics[width=.2\linewidth, height=.2\linewidth]{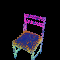}
         \\
        \rotatebox{90}{\parbox[t]{3cm}{\hspace*{\fill}\Large{SD}\hspace*{\fill}}}\hspace*{5pt}
         & \includegraphics[width=.2\linewidth, height=.2\linewidth]{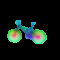}
         & \includegraphics[width=.2\linewidth, height=.2\linewidth]{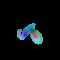}
         & \includegraphics[width=.2\linewidth, height=.2\linewidth]{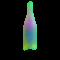}
         & \includegraphics[width=.2\linewidth, height=.2\linewidth]{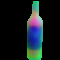}
         & \includegraphics[width=.2\linewidth, height=.2\linewidth]{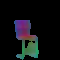}
         & \includegraphics[width=.2\linewidth, height=.2\linewidth]{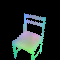}
         \\
        \rotatebox{90}{\parbox[t]{3cm}{\hspace*{\fill}\Large{DINOv2+SD}\hspace*{\fill}}}\hspace*{5pt}
         & \includegraphics[width=.2\linewidth, height=.2\linewidth]{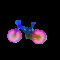}
         & \includegraphics[width=.2\linewidth, height=.2\linewidth]{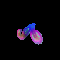}
         & \includegraphics[width=.2\linewidth, height=.2\linewidth]{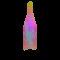}
         & \includegraphics[width=.2\linewidth, height=.2\linewidth]{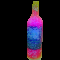}
         & \includegraphics[width=.2\linewidth, height=.2\linewidth]{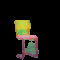}
         & \includegraphics[width=.2\linewidth, height=.2\linewidth]{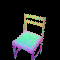}
         \\
        \rotatebox{90}{\parbox[t]{3cm}{\hspace*{\fill}\Large{Sphere}\hspace*{\fill}}}\hspace*{5pt}
         & \includegraphics[width=.2\linewidth, height=.2\linewidth]{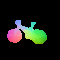}
         & \includegraphics[width=.2\linewidth, height=.2\linewidth]{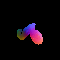}
         & \includegraphics[width=.2\linewidth, height=.2\linewidth]{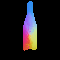}
         & \includegraphics[width=.2\linewidth, height=.2\linewidth]{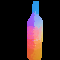}
         & \includegraphics[width=.2\linewidth, height=.2\linewidth]{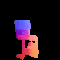}
         & \includegraphics[width=.2\linewidth, height=.2\linewidth]{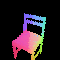}
         \\
    \end{tabular}
    }
    \vspace{-7pt}
    \caption{Example dense correspondence maps for categories from the SPair-71k~\cite{min2019spair} dataset. 
    }
    \label{fig:supp_Spair_maps}
    \vspace{-12pt}
\end{figure*}

\end{document}